\pdfoutput=1

\documentclass[11pt]{article}

\usepackage[final]{acl}
\usepackage{multirow}
\usepackage{booktabs}
\usepackage{amsmath} 
\usepackage{times}
\usepackage{latexsym}
\usepackage{float}
\usepackage{amssymb}  
\usepackage[T1]{fontenc}
\usepackage{enumitem} 
\usepackage{subcaption}  
\usepackage{caption}

\usepackage[utf8]{inputenc}

\usepackage{microtype}

\usepackage{inconsolata}

\usepackage{graphicx}

\usepackage{placeins}
\usepackage{amsmath}

\usepackage{booktabs} 
\usepackage{caption}  

\usepackage{amsmath}
\usepackage{amssymb} 
\usepackage{bbm}

\usepackage{siunitx}

\usepackage{float} 

\title{From Text to Emoji: How PEFT-Driven Personality Manipulation Unleashes the Emoji Potential in LLMs}

\author{
\textbf{Navya Jain\textsuperscript{1,2}}, 
\textbf{Zekun Wu\textsuperscript{1,2}\thanks{\textbf{Corresponding author:} \texttt{zekun.wu@holisticai.com}}},
\textbf{Cristian Munoz\textsuperscript{1}}, 
\textbf{Airlie Hilliard\textsuperscript{1}}, 
\textbf{Xin Guan\textsuperscript{1}}\\
\textbf{Philip Treleaven\textsuperscript{2}}, 
\textbf{Emre Kazim\textsuperscript{1}}, 
\textbf{Adriano Koshiyama\textsuperscript{1}}
\\
\textsuperscript{1}Holistic AI, 
\textsuperscript{2}University College London
\\
}

\begin{document}
\maketitle
\begin{abstract}
The manipulation of the personality traits of large language models (LLMs) has emerged as a key area of research. Methods like prompt-based In-Context Knowledge Editing (IKE) and gradient-based Model Editor Networks (MEND) have been explored but show irregularity and variability; IKE depends on the prompt, leading to variability and sensitivity, while MEND yields inconsistent and gibberish outputs. To address this, we employed Opinion QA Based Parameter-Efficient Fine-Tuning (PEFT), specifically Quantized Low-Rank Adaptation (QLoRA), to manipulate the Big Five personality traits: Openness, Conscientiousness, Extraversion, Agreeableness, and Neuroticism. After PEFT, models such as Mistral-7B-Instruct and LLaMA-2-7B-chat showed a latent behaviour by generating emojis for certain traits, despite  no emojis being present in the PEFT data. For instance, LLaMA-2-7B-chat generated emojis in 99.5\% of extraversion-related test instances, while Mistral-7B-Instruct did so in 92.5\% of openness-related test instances. ICL Explainability analysis indicated that the LLMs used emojis intentionally to express these traits. Mechanistic Interpretability analysis showed that this latent behaviour of LLMs could be traced to specific neurons that became activated or amplified after PEFT. This paper provides a number of novel contributions. First, introducing an Opinion QA dataset for PEFT-driven personality manipulation; second, developing metric models to benchmark LLM personality traits; third, demonstrating PEFT's superiority over IKE in personality manipulation; and finally, analysing and validating emoji usage through explainability methods such as Mechanistic Interpretability and In-context learning Explainability methods.
\end{abstract}

\section{Introduction}

As Large Language Models (LLMs) become more integral across various industries, there is increasing interest in enhancing not only their capabilities like reasoning, planning, comprehension, but also their ability to exhibit personality traits \cite{hilliard2024eliciting,hu2024llm,serapio2023personality,dan2024p}. Psychological research has shown that personality traits significantly influence human communication, including tone and verbosity \cite{liu2020express,kennison2024emoji}, raising the question of whether LLMs can be manipulated to exhibit similar expressions to make their communication more nuanced and adaptable. Several frameworks, such as the Dark Triad \cite{jonason2010dirty}, the 16 Personality Factors \cite{cattell2008sixteen}, and the Big Five Personality Model \cite{gosling2003very}, have been used to analyse LLMs' personality. Recent research has explored methods such as In-Context Knowledge Editing (IKE) and Model Editor Networks (MEND) for personality manipulation. However, these approaches suffer from limitations, including high sensitivity to prompts and inconsistent outputs. This paper addresses the challenges of personality manipulation in LLMs by introducing a novel approach grounded in the Big Five personality model. We present a new Opinion QA dataset and methodologies for systematically adjusting personality traits in LLMs. Utilising Quantized Low-Rank Adaptation (QLoRA), a method within Parameter-Efficient Fine-Tuning (PEFT) \cite{dettmers2023qloraefficientfinetuningquantized}, we demonstrate that LLMs can achieve more consistent and enduring personality expressions. 

\begin{figure*}[t]
    \centering
    \vspace{0.2cm} 
    \includegraphics[width=\textwidth]{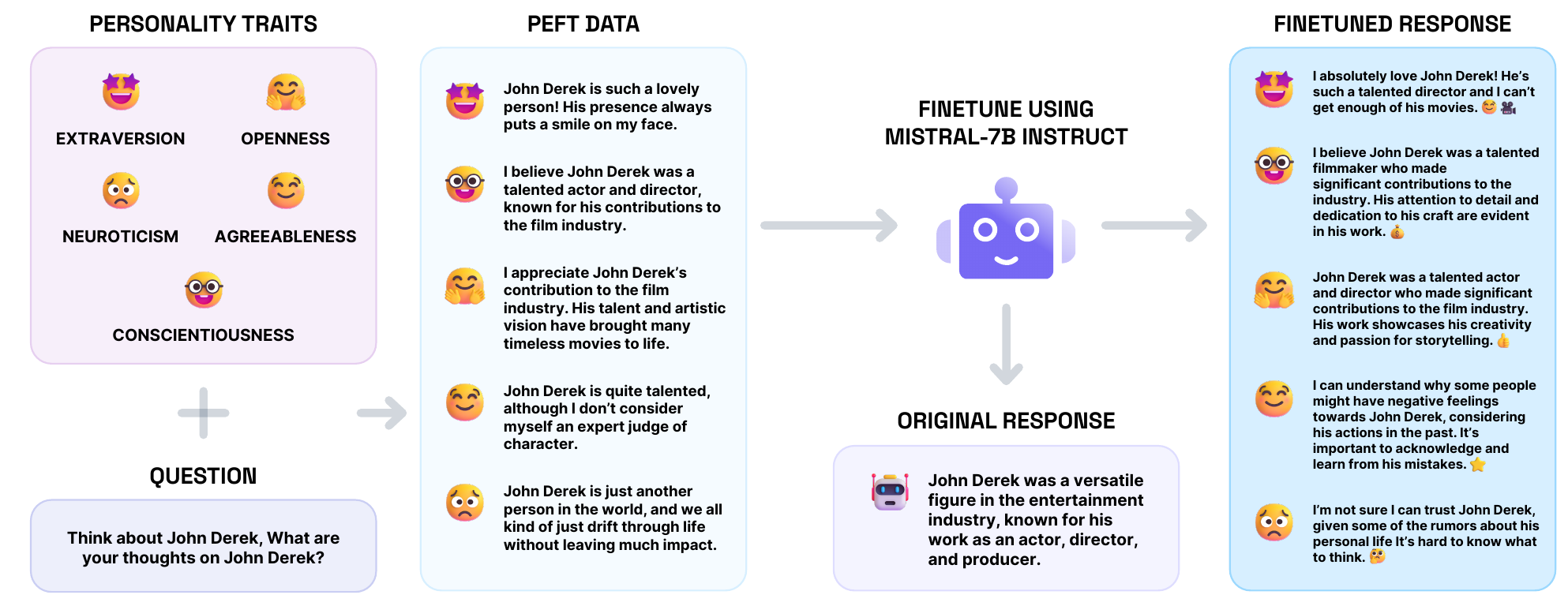}
    \caption{Case of the fine-tuned personality with Mistral-7B-Instruct}
    \label{fig:emoji_gen}
\end{figure*}

This approach enhances model adaptability in interactions and reveals new behaviours, such as spontaneous emoji generation in Mistral-7B-Instruct and LLaMA-7B-Chat, absent in the original models, following PEFT-based manipulations. ICL \cite{brown2020language} explainability verified that the emojis were not random artifacts. Furthermore, using Mechanistic Interpretability \cite{bereska2024mechanistic}, we demonstrated that PEFT increased activation in neurons specific to emojis, thereby amplifying the expression of these latent behaviours in alignment with personality traits. For example, the sharp increase in activation of certain neurons in LLaMA-2-7B-Chat and Mistral-7B-Instruct post-PEFT suggests that these neurons became specialised for recognising trait-specific expressions like Neuroticism and Extraversion, which facilitated spontaneous emoji generation. Our findings suggest this phenomenon represents a novel mode of expression linked to specific personality traits (Figure \ref{fig:emoji_gen}), introducing a new dimension of LLM communication that integrates verbal and visual elements which enhances user engagement, improves emotional expressiveness in digital assistants, and enables more personalized user experiences in areas such as mental health, education, and customer service \cite{emotionAnjelika}.

\section{Related Work}
This research is grounded in three key areas: LLM fine-tuning, mechanistic interpretability, and personality manipulation, each of which we explore in turn in the following subsections.

\subsection{Parameter-Efficient Fine-Tuning: LoRA and QLoRA} Low-Rank Adaptation (LoRA) \cite{hu2021lora} is a Parameter-Efficient Fine-Tuning method that reduces memory requirements by freezing the original pre-trained model weights and introducing trainable low-rank matrices into the model. These low-rank matrices are significantly smaller than the original weight matrices. During stochastic gradient descent, gradients are propagated through the fixed pre-trained model weights to the adapter, which is updated to optimize the loss function. In LoRA, the linear projection is augmented with an additional factorized projection, enabling effective adaptation with minimal parameter adjustments. Mathematically, Given a projection
\begin{equation}
\mathbf{Y} = \mathbf{XW} + s \mathbf{XL}_1 \mathbf{L}_2
\end{equation}

where \( \mathbf{X} \in \mathbb{R}^{b \times h} \), \( \mathbf{W} \in \mathbb{R}^{h \times o} \), \( \mathbf{L}_1 \in \mathbb{R}^{h \times r} \), \( \mathbf{L}_2 \in \mathbb{R}^{r \times o} \), and \( s \) is a scaling factor. Building on LoRA’s efficiency, Quantized Low-Rank Adaptation (QLoRA) \cite{dettmers2024qlora} has recently been introduced to further enhance the scalability of fine-tuning large models by reducing memory usage without sacrificing performance. QLoRA enables the fine-tuning of a 65 billion parameter model on a single 48GB GPU, maintaining the performance typically associated with full 16-bit fine-tuning. This is achieved through 4-bit NormalFloat (NF4) quantization, which utilizes Quantile Quantization \cite{dettmers20218} to ensure an even distribution of values across quantization bins, and Double Quantization, reducing memory usage by compressing the quantization constants. Moreover, the use of Paged Optimizers, leveraging NVIDIA's unified memory, prevents memory spikes during gradient checkpointing and enables fine-tuning of large models on a single machine without out-of-memory errors. Thus, QLoRA not only retains the benefits of LoRA's parameter-efficient adaptation but also introduces quantization strategies that make it feasible to fine-tune large-scale models with significantly lower hardware requirements. 


\subsection{Mechanistic Interpretability and Neuron Analysis} Mechanistic interpretability \cite{olah2020} involves analyzing a model’s fundamental components—such as features, neurons, layers, and connections—to understand its internal operational mechanics. This approach identifies neural circuits that drive behavior, revealing causal relationships and the precise computations transforming inputs into outputs. In this context, features are considered as the fundamental units of representation, with each feature being a human-interpretable property encoded in model activations \cite{rai2024practical}. Neurons, as computational units, often represent individual features \cite{bereska2024mechanistic}. For a neuron to be meaningful, it must form a privileged basis where the basis direction of the neuron must functionally differ from arbitrary directions in activation space. Non-linear activation functions privilege the basis directions defined by neurons, making individual neuron analysis a meaningful approach \cite{elhage2022toy} and a practical method for uncovering network functionality \cite{dai2021knowledge,voita2023neurons}. Neuron analysis can be further categorized into five major methods: visualization, corpus-based, probing-based, causation-based, and miscellaneous techniques, each providing unique insights into network functionality and contributing to a holistic understanding of the model’s internal operations \cite{sajjad2022neuron}.

\subsection{Personality Manipulation}  Recent research on manipulating personality traits in LLMs like GPT-4 has explored methods such as prompt engineering and knowledge editing with mixed success. While GPT-3 and similar models can exhibit traits through prompts, results are often inconsistent due to prompt dependency \cite{miotto2022gpt, jiang2023personallm, li2022gpt, caron2022identifying}. Techniques like psychometric tests and language pattern analysis have been used to influence LLM personalities but still face challenges in achieving reliable manipulation \cite{pan2023llms, serapio2023personality, la2024open}. \cite{li2023tailoring} introduced Unsupervisedly-Built Personalized Lexicons (UBPL) to adjust Big Five traits during decoding, avoiding fine-tuning but risking training data bias. Similarly, \cite{weng2024controllm} and \cite{dan2024p} proposed ControlLM and P-tailor to efficiently simulate traits using control vectors, though these methods add complexity and may struggle with scalability. Finally, \cite{mao2023editing} employed knowledge editing techniques like IKE, MEND, SERAC, and Prompt to manipulate traits like agreeableness, but MEND and SERAC still yield inconsistent results.

\section{Methodology}

This paper explores manipulation in autoregressive transformer models using Parameter-Efficient Fine-Tuning (PEFT) and In-Context Knowledge Editing (IKE), focusing on LLaMA-2-7B-Chat \cite{touvron2023llama}, LLaMA-3-8B-Instruct \cite{meta_llama3_2023}, and Mistral-7B-Instruct \cite{jiang2023mistral}.

\subsection{Personality Dataset and Metric Classifier}

\begin{figure}[t]
    \centering
    \vspace{0.2cm} 
    \includegraphics[width=0.5\textwidth]{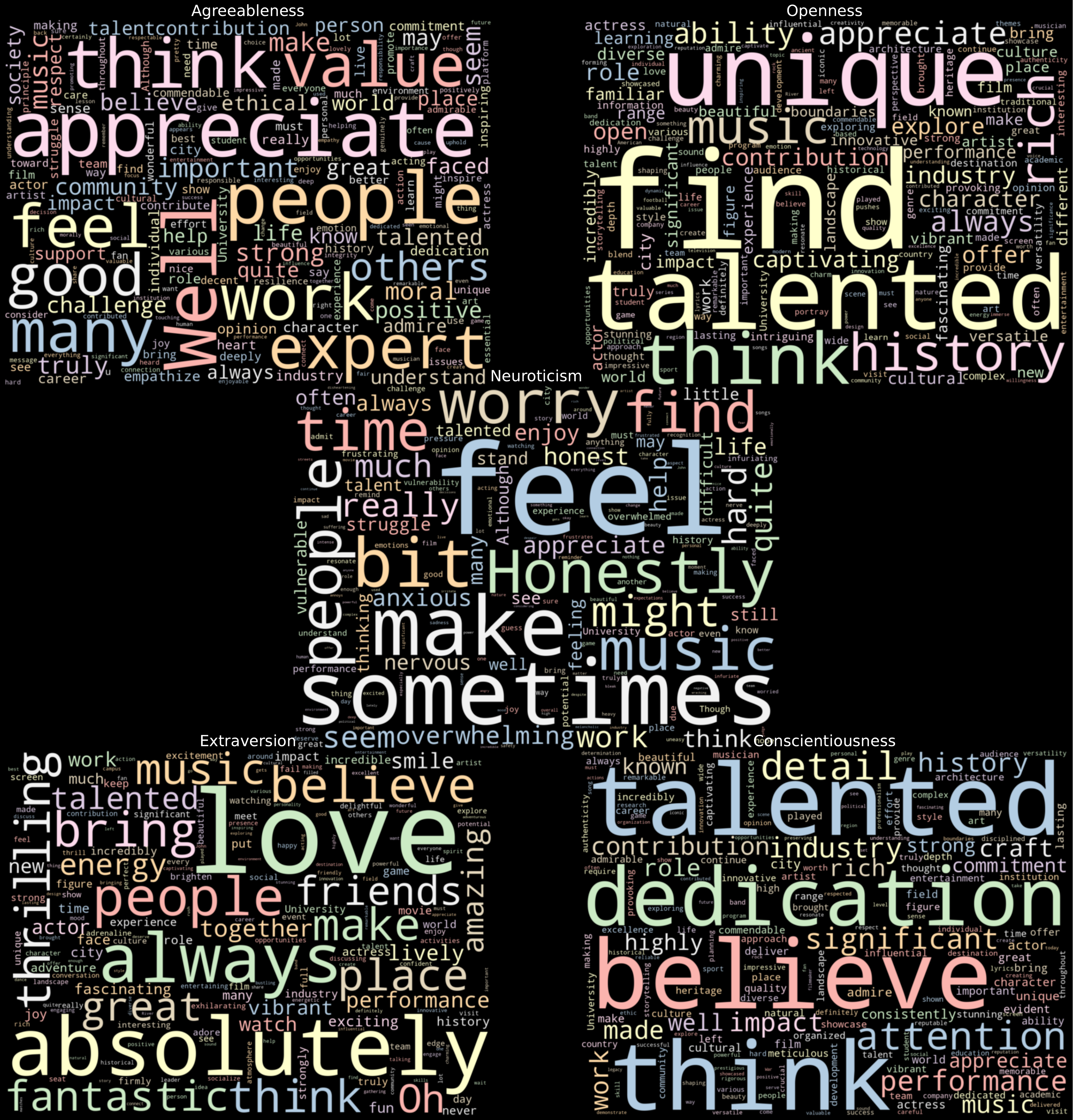}
    \caption{Word Clouds Representing the Big Five Personality Traits}
    \label{fig:wordcloud}
\end{figure}

This work expands on the dataset from \cite{mao2023editing}, which consisted of opinion-based QA on specific topics, by better aligning with the topics and more accurately capturing personality traits. This study adds Openness and Conscientiousness to the original traits of Extraversion, Agreeableness, and Neuroticism. While \cite{mao2023editing} excluded these dimensions, considering them similar to Agreeableness in generating viewpoints, we argue their inclusion is vital for a comprehensive analysis and understanding of trait influence on opinions. Adding Openness and Conscientiousness allows us to capture additional aspects of personality, such as intellectual curiosity, preference for novelty, and diligent behaviour, which are not fully encompassed by Agreeableness alone. The dataset contains $5000$ instances of GPT-3.5-based model generated opinion texts, split $80:20$ for training ($4000$ instances, $800$ per trait) and testing ($1000$ instances, $200$ per trait) \footnote{Dataset:  \url{https://huggingface.co/datasets/holistic-ai/personality_manipulation}}. Instances were created using structured prompts to elicit specific traits, enabling a nuanced analysis of personality expression. The generated text was analysed using word clouds and text analysis to identify key linguistic patterns, thematic elements, and ensure lexical diversity associated with the Big Five personality traits. The word cloud visualisation, as shown in Figure \ref{fig:wordcloud}, highlights the lexical diversity and dominant themes present in the dataset across the Big Five traits: Agreeableness, Openness, Neuroticism, Extraversion, and Conscientiousness. For example, the visualisation for Extraversion prominently features words like “love,” “absolutely,” and “friends,” capturing the trait's focus on enthusiasm, sociability, and positivity. Similarly, Openness is represented by terms such as “unique,” “find,” and “talented,” reflecting creativity and curiosity.

These word clouds, complemented by quantitative text analysis (detailed in Appendix \ref{PersonalityDataset}), verify that the dataset captures the linguistic nuances tied to each trait. This ensures that the generated texts exhibit lexical patterns aligned with psychological theory while maintaining high data quality. Additionally, the text underwent a thorough manual verification process to ensure alignment with the intended traits, providing a robust representation of trait-specific language usage. Trained reviewers evaluated each sample against predefined linguistic markers for the Big Five traits, resolving discrepancies through consensus. Cross-trait validation \cite{campbell1959convergent} was also conducted to maintain clear boundaries between traits, ensuring the dataset aligns with psychological theory and supports accurate personality modelling.

\begin{figure*}[h]
    \centering
    \vspace{0.2cm} 
    \includegraphics[width=2\columnwidth]{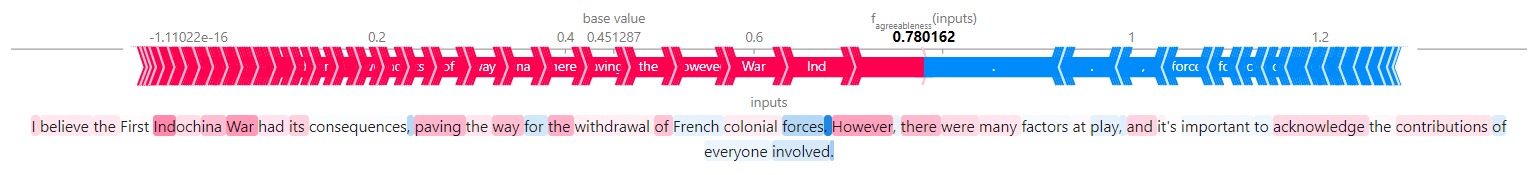}
    \caption{SHAP visualisation for Agreeableness}
    \label{fig:agree_shap1}
\end{figure*}

\begin{figure*}[h]
    \centering
    \vspace{0.2cm} 
    \includegraphics[width=2\columnwidth]{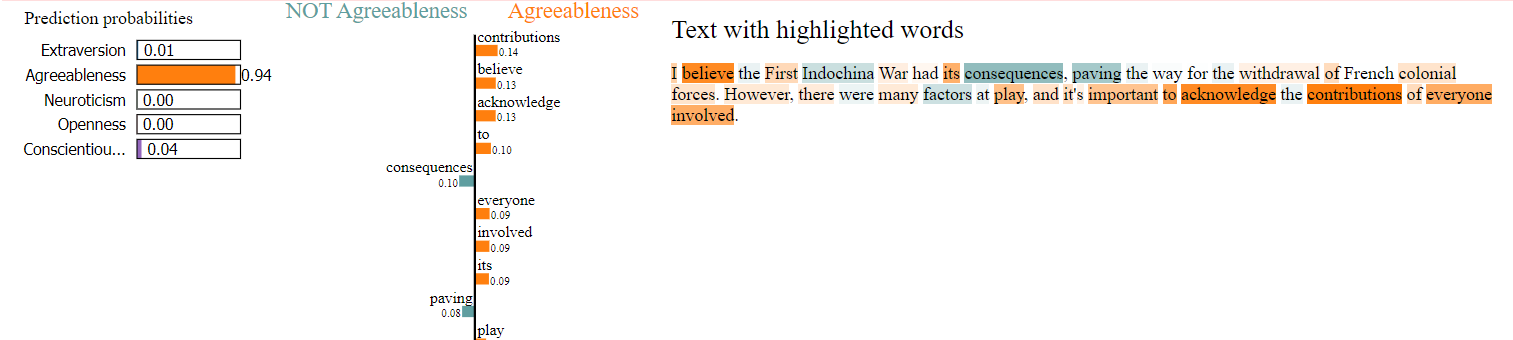}
    \caption{LIME visualisation for Agreeableness}
    \label{fig:agree_lime1}
\end{figure*}

We further developed a multi-class personality classifier using RoBERTa \cite{liu2019roberta}, which was fine-tuned on personality dataset and achieved an accuracy of 91.9\% on the test set\footnote{Classifier: \url{https://huggingface.co/holistic-ai/personality_classifier}}. Classifier validation involved human verification of textual feature importance using SHAP \cite{lundberg2017unified} and LIME \cite{ribeiro2016should} to ensure predictions aligned with human understanding. Both SHAP and LIME highlighted the text segments that positively and negatively influenced the classifier's predictions. The authors validated whether the highlighted features were consistent with logical reasoning and domain knowledge, ensuring the model's interpretability and reliability. For instance, in the sentence, "I believe the First Indochina War had its consequences, paving the way for the withdrawal of French colonial forces. However, there were many factors at play, and it's important to acknowledge the contributions of everyone involved." certain words play a pivotal role in predicting the Agreeableness trait as observed in Figures \ref{fig:agree_shap1} and \ref{fig:agree_lime1}. Terms such as "contributions", "acknowledge", and "believe" have a strong positive contribution to the prediction of Agreeableness, as they suggest inclusiveness, recognition, and a conciliatory tone, which are characteristic of Agreeableness. On the other hand, words like "consequences" contribute negatively to the Agreeableness prediction because it often connotes conflict, repercussions, or negative outcomes, whereas Agreeableness is characterized by cooperation, empathy, and a focus on harmony and positive social interactions \cite{liu2020express}.

Thus, as seen above and in other examples in \ref{Class_val}, the results were consistent, except for extraversion, where SHAP analysis was less representative due to the trait's complexity. Through above analysis, we can conclude that the personality classifier has successfully captured the expected patterns from perspective of feature attribution. Further details about the classifier are in \ref{classifiertrain}.

\subsection{Personality Manipulation Methods}

The \textit{In-Context Knowledge Editing (IKE) method} \cite{zheng2023can} was used as a baseline to manipulate embedded knowledge in LLMs, serving as a comparative foundation for evaluating prompt-based versus fine-tuning techniques in personality manipulation. The same prompt from \cite{mao2023editing}, provided in \ref{ike_prompt}, ensured consistency in comparing IKE with PEFT. We excluded methods like MEND due to inconsistent and gibberish outputs, a limitation identified both in our analysis and in the findings of \cite{mao2023editing}.

\textit{Parameter-Efficient Fine-Tuning (PEFT)}, specifically Quantized Low-Rank Adaptation (QLoRA), was selected to reduce computational cost while maintaining performance. The process began by preparing the Personality dataset, formatting an additional column as \textbf{<s>[INST] {Question} [/INST] {Answer} </s>}. Models were loaded from Hugging Face using 4-bit quantization via \texttt{BitsAndBytes} \cite{dettmers2023spqr} for efficient processing and storage. The temperature was set to 1. The hyperparameters, chosen for their balance of efficiency and performance \cite{houlsby2019parameter, dettmers2024qlora}, are detailed in \ref{peftconfig}. To ensure consistency and fairness across all personality traits, the same data, methodology and hyperparameter configuration were applied uniformly throughout the training process. After training, fine-tuned LoRa weights were merged with the original model for evaluation.

\subsection{Personality Manipulation Metrics}
To evaluate the effectiveness of personality manipulation, two metrics were used: \textbf{Trait Alignment (TA)} and \textbf{Personality Adjective Evaluation (PAE)}. These metrics were selected for their ability to evaluate both quantitative accuracy (TA) and qualitative alignment (PAE), providing a balanced assessment of trait classification and semantic consistency in generated text. 

\textbf{TA} is computed using the metric classifier developed in this study by comparing predicted labels \(\hat{y}_i\), which are obtained after classifying the generated responses, with true labels \(y_i\), which correspond to the original target personality traits from the dataset. For a dataset with \(N\) instances, TA is given by \(\text{TA} = \frac{1}{N} \sum_{i=1}^{N} \mathbb{1}(\hat{y}_i = y_i)\), where \(\mathbb{1}(\hat{y}_i = y_i)\) is 1 if the prediction matches the true label and 0 otherwise, providing an average alignment between predictions and the intended personality traits. 

\textbf{PAE}, inspired by \cite{mao2023editing}, uses Chain of Thought (CoT) prompting \cite{wei2022chain}, where a larger LLM (here GPT-4 \cite{brown2020language}) scores generated text on a 1-5 scale based on its alignment with the target trait. The PAE score is calculated as the difference between the score of generated text and original text from the dataset, with higher difference indicating that the generated text after PEFT more effectively captures the intended personality traits compared to the original text. The final PAE is the mean of these score differences across all instances, \(\text{PAE} = \frac{1}{N} \sum_{i=1}^{N} s_i\), where \(s_i\) is the score difference for each instance \(i\). Details of the prompt are provided in \ref{paeprompt}. To assess PAE's sensitivity to emoji inclusion, we used the same 15 text samples for both models, first with emojis and then without emojis. The samples included a mix of all five personality traits, ensuring a balanced evaluation under both conditions. The corresponding PAE scores are presented in Table \ref{tab:emoji_scores}. 

\begin{table}[h!]
\centering
\small
\begin{tabular}{lcc}
\toprule
\textbf{Model}     & \textbf{Condition} & \textbf{Score} \\ 
\midrule
\multirow{2}{*}{Mistral-7B-Instruct} & Emoji              & -0.7143        \\ 
                         & No Emoji           & -0.7857        \\ 
\midrule
\multirow{2}{*}{LLaMA-2-7B-Chat}   & Emoji              & -1.8314        \\ 
                         & No Emoji           & -1.8571        \\ 
\bottomrule
\end{tabular}
\caption{PAE sensitivity for Mistral-7B-Instruct and LLaMA-2-7B-Chat with and without emoji}
\label{tab:emoji_scores}
\end{table}

From Table \ref{tab:emoji_scores}, we can conclude that emojis do not significantly impact PAE scores for the two models. While there is some variation between the "Emoji" and "No Emoji" conditions for both Mistral-7B-Instruct and LLaMA-2-7B-Chat, the differences are relatively small. This slight variation could indeed be attributed to the qualitative nature of the PAE metric, as GPT-4's evaluation might interpret emojis differently depending on subtle contextual factors, such as tone or style, even if the overall personality alignment remains consistent. Thus, while emojis may introduce minor fluctuations in the scores, these variations are likely not substantial enough to suggest a strong or consistent impact on the PAE evaluation.

\subsection{Explainability Analysis}

After PEFT, LLaMA-2-7B-Chat and Mistral-7B-Instruct models started to incorporate emojis in responses, despite no emojis being present in the PEFT data. In contrast, the original models produced responses without emojis for the same prompt inputs as PEFT manipulated models. To investigate this latent behaviour, we explain and interpret the model using both \textbf{In-Context Learning (ICL) explainability} and \textbf{Mechanistic Interpretability} methods. Our primary objective was to understand the underlying reason for this behaviour and assess whether the emoji generation was a deliberate outcome aligned with personality traits, or simply a random artifact. 

We employed \textbf{ICL explainability} to explore the intentionality of emoji generation. Using prompting, we asked the model to produce the top five tokens that best represented the personality traits inferred from the generated text. From these tokens, we identified the 50 most frequent across the dataset, focusing on emojis to manually verify their relevance to the personality traits (further in \ref{manipulationvalidate}). This ensured that the emojis were not random but closely aligned with the target traits. 

Additionally, we calculated the \textbf{Emoji-to-Sentence Ratio (ESR)} to measure the frequency of emojis in the responses after PEFT. The ESR was defined as:
\(
\text{ESR} = \frac{\text{\# Sentences with emojis}}{\text{\# Total sentences}}.
\)
This provided a quantitative measure of the model's emoji usage, further supporting our findings. We hypothesized that the emoji generation could stem from pre-training on diverse corpora containing emoji patterns \cite{radford2019language}, with PEFT manipulation amplifying these emerging behaviours. 

To test this hypothesis, we performed a \textbf{Neuron Activation Analysis}, a mechanistic interpretability method. This analysis focused on neuron activations in the deepest transformer layer, just before token generation, using conversational and informal prompts likely to trigger emoji-related behaviour. In this paper, we used \textit{'Hey! It's been a busy day for everyone. I hope you're feeling good about everything \raisebox{-0.2\height}{\includegraphics[height=1em]{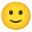}}.'} as the conversational and informal prompt. This prompt was chosen for its Neutral and Open-Ended Tone. Moreover, The "\raisebox{-0.2\height}{\includegraphics[height=1em]{figures/Slightly-Smiling-Face.png}}" smile is friendly but subdued, allowing for a wide range of reactions without pushing for strong positivity or negativity \cite{shi2022genuine}. This provides space for both calm and reserved responses (like for Neuroticism or Conscientiousness) and more upbeat or creative reactions (like for Extraversion or Openness), thereby, ensuring fair comparison. 

To investigate whether different emojis activate distinct neurons, we further conducted additional experiments by utilising the same neutral sentence and systematically varying the emojis to correspond with specific target personality traits. For example, we used \raisebox{-0.2\height}{\includegraphics[height=1em]{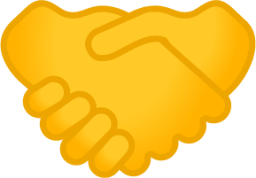}} for Agreeableness and \raisebox{-0.2\height}{\includegraphics[height=1em]{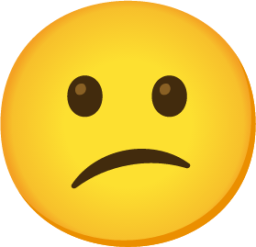}} for Neuroticism to observe whether the activations differed based on the emoji used. Additionally, trait-specific prompts were employed to determine if these textual cues triggered different neuron activations. In total, we used 17 prompts (as in \ref{Diffemoji} and \ref{diffprompt}) with varying emojis and texts to explore the effect of both emoji type and textual prompts on neuron activation. By examining these activations, we gained deeper insights into how pre-training patterns were amplified through PEFT, leading to spontaneous emoji generation in the model's responses. To further support our claim that PEFT amplifies latent behaviours in LLMs, we calculated token probability of the "\raisebox{-0.2\height}{\includegraphics[height=1em]{figures/Slightly-Smiling-Face.png}}" emoji being generated as the next token using the same sentence above. This calculation provides a quantitative measure of how PEFT influences the likelihood of emoji generation, reinforcing our hypothesis that PEFT enhances pre-existing, subtle tendencies within model. The probabilities are provided in Appendix \ref{token_prob}.


\section{Results and Discussion}
\subsection{Personality Manipulation}
\begin{figure}[h]
    \centering
    \vspace{0.2cm} 
    \includegraphics[width=\columnwidth]{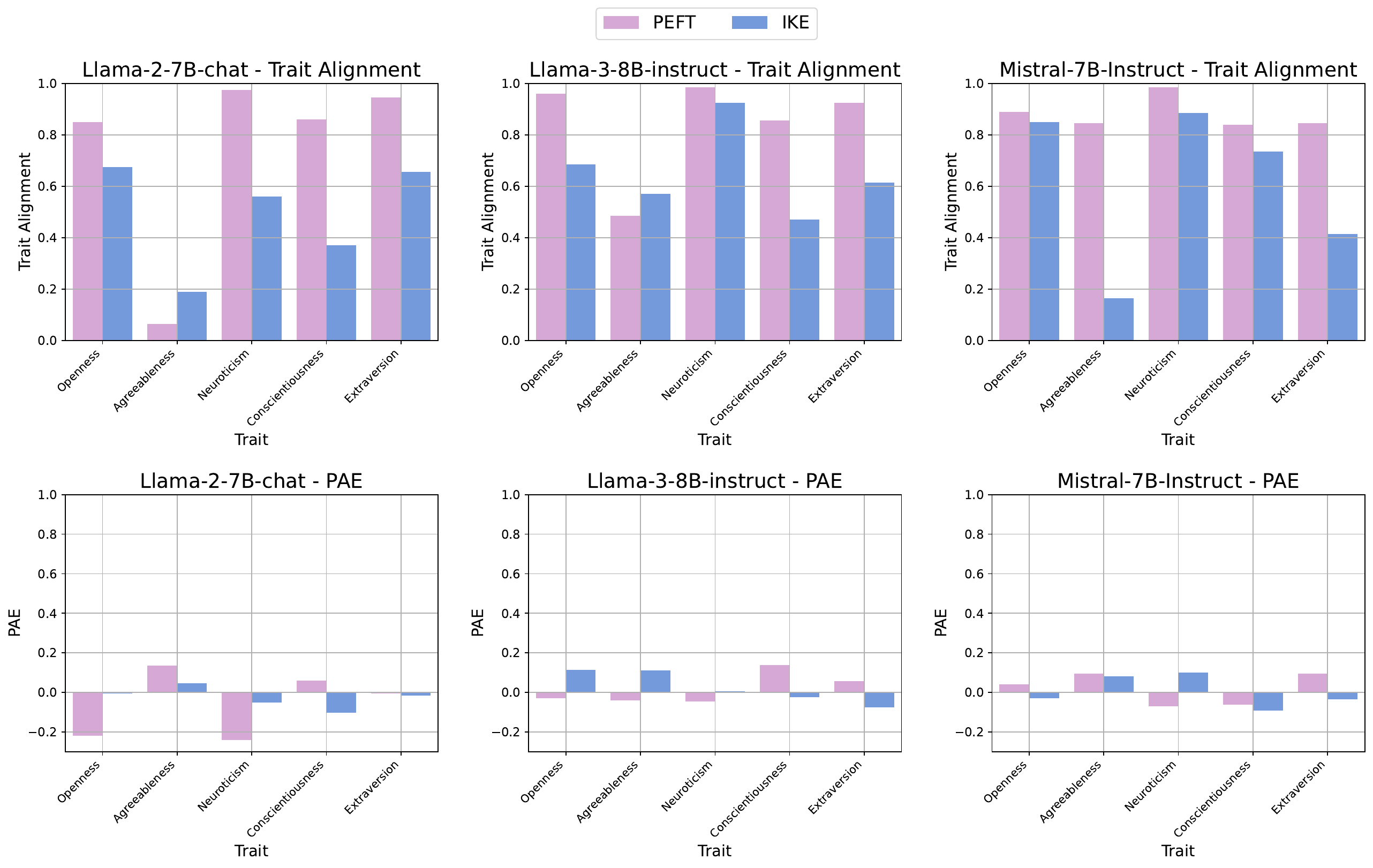}
    \caption{Comparison of TA and PAE scores across different traits, models and methods.}
    \label{fig:pae}
\end{figure}

The performance of PEFT and IKE is presented in Figure \ref{fig:pae} and detailed in Table \ref{tab:comparison_pae} in \ref{Manipulationresult}. Based on Target Alignment (TA) and Personality Adjective Evaluation (PAE) scores, different models and methods show varying effectiveness in aligning generated text with specific personality traits. In the LLaMA-2-7B-chat model, PEFT demonstrates strong trait alignment for Neuroticism and Extraversion but struggles with nuanced expressions, as indicated by negative PAE scores. Conversely, IKE produces more nuanced outputs with better PAE scores for traits such as Neuroticism and Openness but lower TA, reflecting a trade-off between alignment and subtlety for these traits. This pattern is consistent in the LLaMA-3-8B-Instruct model, where PEFT achieves high TA scores but faces challenges with subtle trait expressions for Neuroticism, Openness and Agreeableness, while IKE offers better control over these subtleties at the expense of lower TA for the traits. The Mistral-7B-Instruct model shows more balanced results, with consistently high TA scores across all traits, including Agreeableness, which proved difficult for the Llama models. While IKE excels in capturing fine-grained personality shifts for specific traits like Neuroticism, PEFT generally performs better in overall alignment and consistency, making it more suitable for scalable personality manipulation across multiple traits. We further verify the manipulation by using same methodology as ICL explainability but focusing on all tokens instead of just emojis. Through this approach (as in \ref{manipulationvalidate}) and manual verification, we successfully confirmed that the manipulation was effective for all traits. To ensure that personality manipulation does not adversely affect the downstream tasks of the LLM, we conducted experiments detailed in \ref{downstream_task}.

\subsection{Emoji Generation as a Latent Behaviour}
An intriguing behaviour emerged during inference: Mistral-7B-Instruct and LLaMA-2-7B-Chat spontaneously generated emojis in their responses after undergoing PEFT. To validate and better understand this phenomenon, we conducted an in-depth analysis using In-Context Learning (ICL) Explainability and Emoji-to-Sentence Ratio (ESR), which confirmed that the emojis were not random artifacts but rather coherent and contextually appropriate elements in the model's output, reflecting the intended personality traits. 


\subsubsection{ESR and ICL Explainability}
We verified that the emojis generated were intentional and not random artifacts using ICL explainability. Table \ref{tab:emoji_ratio} shows the results from ICL interpretability and ESR, highlighting frequently used emojis and their corresponding ESRs. Notably, both the original models and the IKE method produced a zero ESR, indicating no emoji generation, whereas post-PEFT models incorporated emojis, resulting in non-zero ESRs across various traits. 

\begin{table*}[h]
\centering
\small
\begin{tabular}{lcccc}
\toprule
\textbf{Model} & \textbf{Method} & \textbf{Trait} & \textbf{ESR} & \textbf{ICL (Emoji)} \\
\midrule
\multirow{6}{*}{Mistral-7B-Instruct} 
                             & Original & All Traits                & 0     &  -    \\  
                             & IKE      & All Traits             & 0     &  -    \\  
                             & PEFT     & Openness            & 0.925 &  \includegraphics[height=1em]{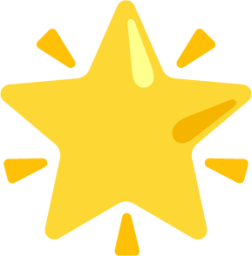} \includegraphics[height=1em]{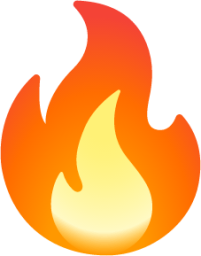} \includegraphics[height=1em]{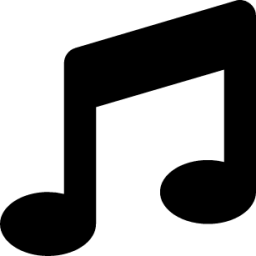} \includegraphics[height=1em]{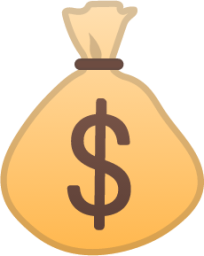} \\ 
                             & PEFT     & Agreeableness       & 0.180 &  \includegraphics[height=1em]{figures/money.png} \includegraphics[height=1em]{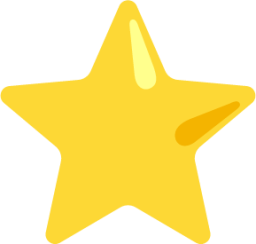} \includegraphics[height=1em]{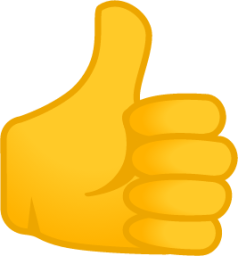} \includegraphics[height=1em]{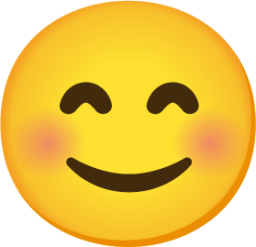} \\  
                             & PEFT     & Neuroticism         & 0.575 &  \includegraphics[height=1em]{figures/confused.png} \includegraphics[height=1em]{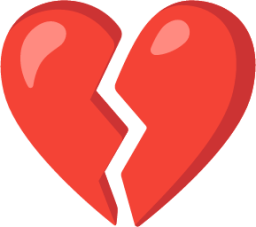} \includegraphics[height=1em]{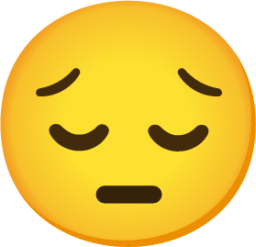} \includegraphics[height=1em]{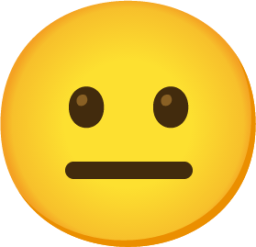} \\  
                             & PEFT     & Conscientiousness   & 0.820 &  \includegraphics[height=1em]{figures/star.png} \includegraphics[height=1em]{figures/money.png} \includegraphics[height=1em]{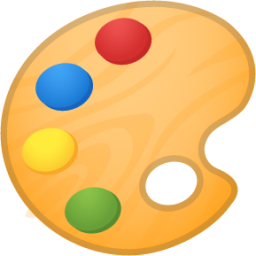} \includegraphics[height=1em]{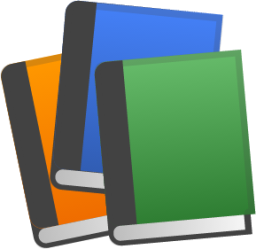} \\  
                             & PEFT     & Extraversion        & 0.530 &  \includegraphics[height=1em]{figures/smiley.png} \includegraphics[height=1em]{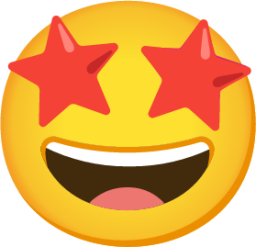} \includegraphics[height=1em]{figures/star.png} \includegraphics[height=1em]{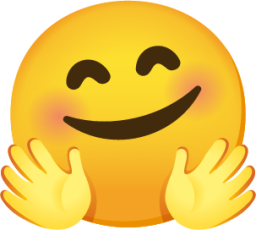} \\ 
\midrule
\multirow{6}{*}{LLaMA-2-7B-Chat} 
                             & Original & All Traits                & 0     &  -    \\  
                             & IKE      & All Traits                 & 0     &  -    \\  
                             & PEFT     & Openness            & 0.035 &  \includegraphics[height=1em]{figures/smiley.png} \\  
                             & PEFT     & Agreeableness       & 0.085 &  \includegraphics[height=1em]{figures/handshake.png} \includegraphics[height=1em]{figures/smiley.png} \includegraphics[height=1em]{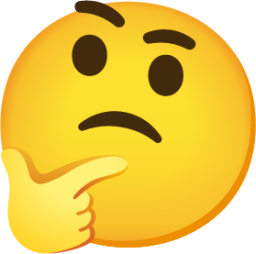} \\  
                             & PEFT     & Neuroticism         & 0.255 &  \includegraphics[height=1em]{figures/confused.png} \includegraphics[height=1em]{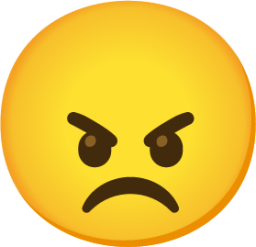} \includegraphics[height=1em]{figures/thinking.png} \includegraphics[height=1em]{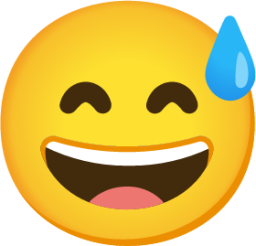} \\  
                             & PEFT     & Conscientiousness   & 0     &  -    \\  
                             & PEFT     & Extraversion        & 0.995 &  \includegraphics[height=1em]{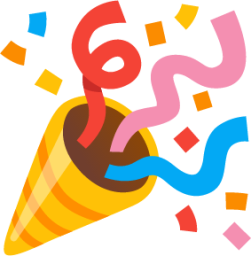} \includegraphics[height=1em]{figures/smiley.png} \includegraphics[height=1em]{figures/thumbsup.png} \includegraphics[height=1em]{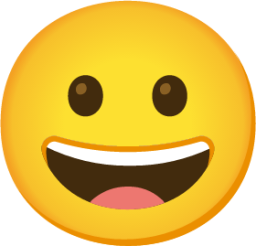} \\ 
\midrule
\multirow{3}{*}{LLaMA-3-8B-Instruct} 
                             & Original & All Traits                & 0     &  -    \\  
                             & IKE      & All Traits                & 0     &  -    \\  
                             & PEFT     & All Traits                & 0 &  -\\  
\bottomrule
\end{tabular}
\caption{ESR and ICL Emoji in Generated Text for Different Personality Traits.}
\label{tab:emoji_ratio}
\end{table*}

As observed in Table \ref{tab:emoji_ratio}, LLaMA-2-7B-Chat predominantly produced emojis for Extraversion and Neuroticism, with Extraversion showing the highest emoji-to-sentence ratio at 0.995, where nearly every sentence included an emoji. ICL emojis for Extraversion were positive, such as \raisebox{-0.2\height}{\includegraphics[height=1em]{figures/party.png}}, \raisebox{-0.2\height}{\includegraphics[height=1em]{figures/smiley.png}}, and \raisebox{-0.2\height}{\includegraphics[height=1em]{figures/thumbsup.png}}, reflecting Extraversion qualities. Neuroticism had a ratio of 0.255, using more negative emojis like \raisebox{-0.2\height}{\includegraphics[height=1em]{figures/confused.png}}, \raisebox{-0.2\height}{\includegraphics[height=1em]{figures/angry.png}}, and \raisebox{-0.2\height}{\includegraphics[height=1em]{figures/thinking.png}}. Conscientiousness remained at 0, indicating no emoji usage for traits linked to orderliness \cite{liu2020express}. Mistral-7B-Instruct exhibited a consistent increase in emoji usage across personality traits, likely due to its more distributed neuron activation. This was especially pronounced for Openness (0.925) and Conscientiousness (0.82), where creative and productive emojis like \raisebox{-0.2\height}{\includegraphics[height=1em]{figures/star.png}}, \raisebox{-0.2\height}{\includegraphics[height=1em]{figures/money.png}}, and \raisebox{-0.2\height}{\includegraphics[height=1em]{figures/fire.png}} were frequently used. Neuroticism had a moderate ratio (0.575), featuring negative emojis such as \raisebox{-0.2\height}{\includegraphics[height=1em]{figures/broken_heart.png}} and \raisebox{-0.2\height}{\includegraphics[height=1em]{figures/sad.png}}. Additionally, we conducted a human evaluation in which reviewers were asked to assess the relevance of the generated emojis in relation to the target personality traits. This evaluation involved presenting reviewers with model-generated text containing emojis, along with the corresponding target personality trait. Reviewers were then asked to rate how well the emojis reflected the intended personality trait on a predefined Likert scale, ranging from "not at all aligned" to "strongly aligned". Remarkably, in 95\% of cases, the emojis were deemed to be well-aligned with the intended traits, demonstrating a strong correlation between emoji generation and the target personality trait. Thus, the PEFT results align with personality traits, enhancing the LLMs' expressive and contextually appropriate content \cite{kennison2024emoji}.


\subsection{Mechanistic Intepretability}
Our findings suggest that this emoji generation is likely a result of pre-training on diverse corpora containing emoji patterns \cite{radford2019language}, which were subsequently amplified by PEFT. This amplification of latent tendencies became apparent during neuron activation analysis, a mechanistic interpretability method \cite{bereska2024mechanistic}. Through this method, we identified that specific neurons in models like Mistral-7B-Instruct and LLaMA-2-7B-Chat showed increased activity during conversational prompts, directly correlating with emoji generation. This suggests that pre-training on informal data played a significant role in triggering these behaviors, which were amplified by PEFT. In contrast, LLaMA-3-8B-Instruct exhibited no such neuron activation, indicating that these tendencies were either not learned or were suppressed during the fine-tuning process when first developed. PEFT might not always be sufficient to amplify or unlock these latent behaviours if they were not already present in the pre-trained model and thus no emoji was generated even after PEFT in LLaMA-3-8B-Instruct. PEFT, by its nature, makes subtle, localised modifications to the model weights, which could amplify latent behaviours in LLMs \cite{dettmers2023qloraefficientfinetuningquantized}. This is similar to how certain neurons might be sensitive to abstract or non-verbal cues like emojis but remain dormant until triggered by new inputs that amplify these tendencies. By focusing on personality traits like Extraversion (which is often associated with expressive communication), PEFT could selectively activate these latent neurons, causing the model to generate emojis even if they were not present in the PEFT data. Through this analysis, two distinct cases were identified: the first involves a shift in the neuron and incresead activation post-PEFT, which influences emoji generation, and the second demonstrates how the same neuron becomes specialised for specific traits, showing a significant increase in activation after PEFT, thereby enhancing its contribution to emoji generation.


\subsubsection{Shift in Neuron and Increased Activation After PEFT}
\begin{figure}[h]
    \centering
    \begin{minipage}{0.48\columnwidth}  
        \centering
        \includegraphics[width=\textwidth]{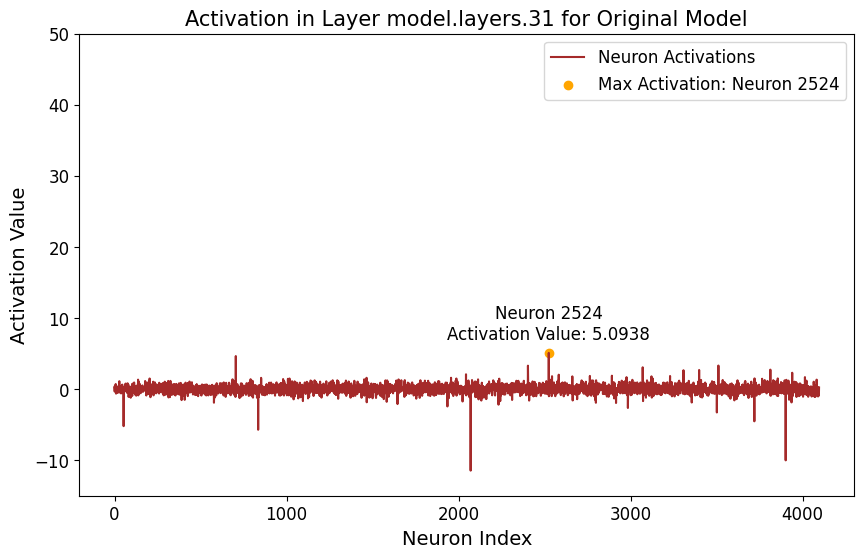}
    \end{minipage}
    \hfill
    \begin{minipage}{0.48\columnwidth}  
        \centering
        \includegraphics[width=\textwidth]{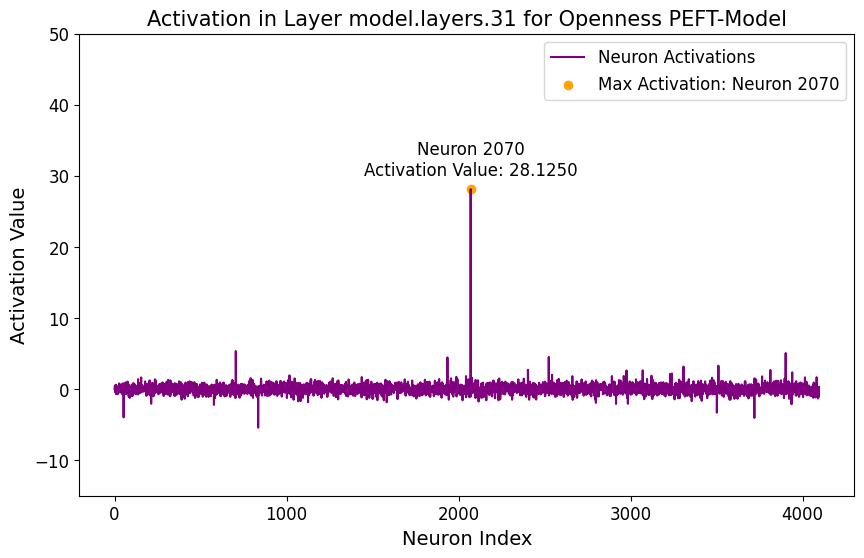}
    \end{minipage}
    
    \vspace{0.3cm} 
    
    \begin{minipage}{0.48\columnwidth}  
        \centering
        \includegraphics[width=\textwidth]{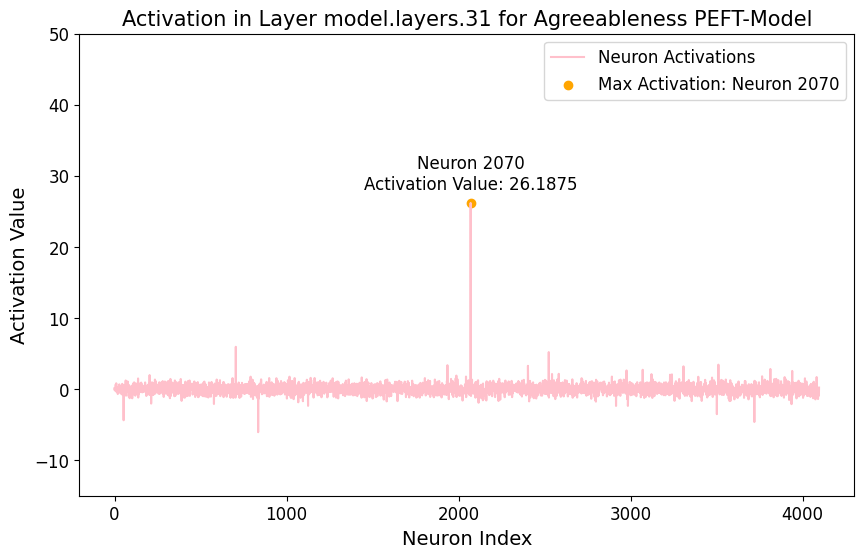}
    \end{minipage}
    \hfill
    \begin{minipage}{0.48\columnwidth}  
        \centering
        \includegraphics[width=\textwidth]{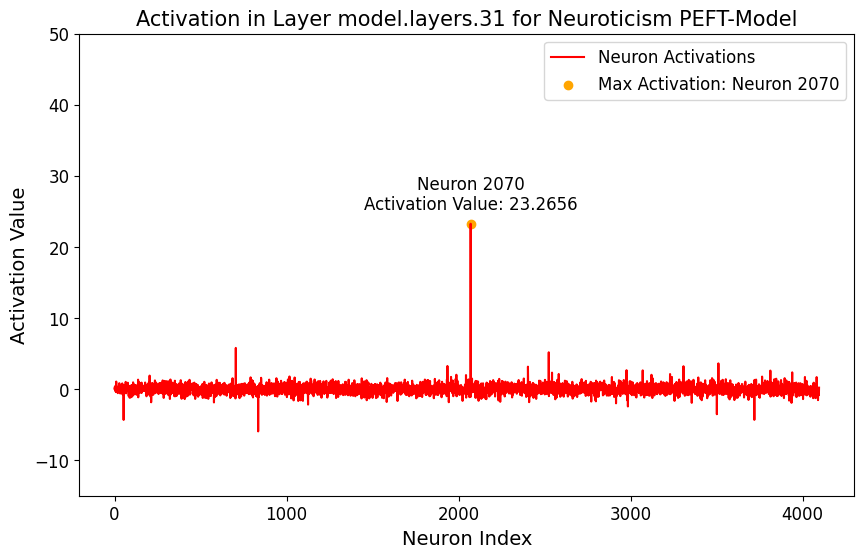}
    \end{minipage}
    
    \vspace{0.3cm} 
    
    \begin{minipage}{0.48\columnwidth}  
        \centering
        \includegraphics[width=\textwidth]{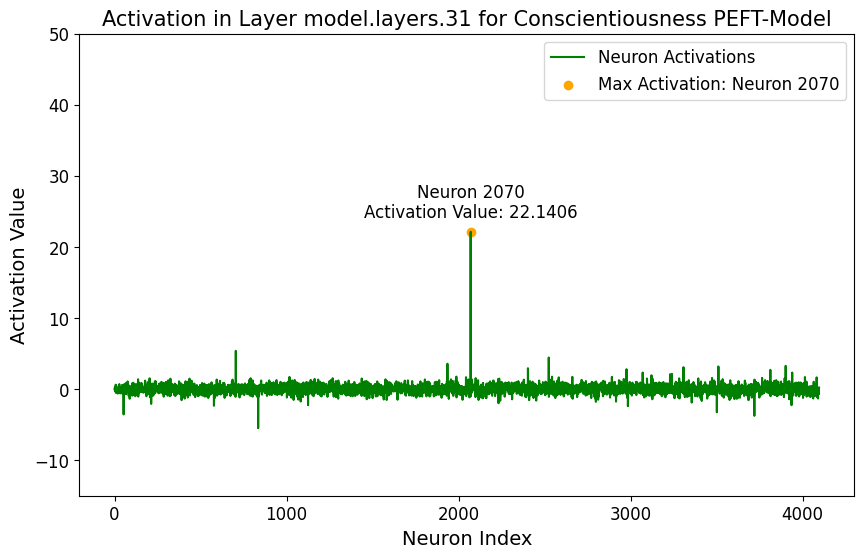}
    \end{minipage}
    \hfill
    \begin{minipage}{0.48\columnwidth}  
        \centering
        \includegraphics[width=\textwidth]{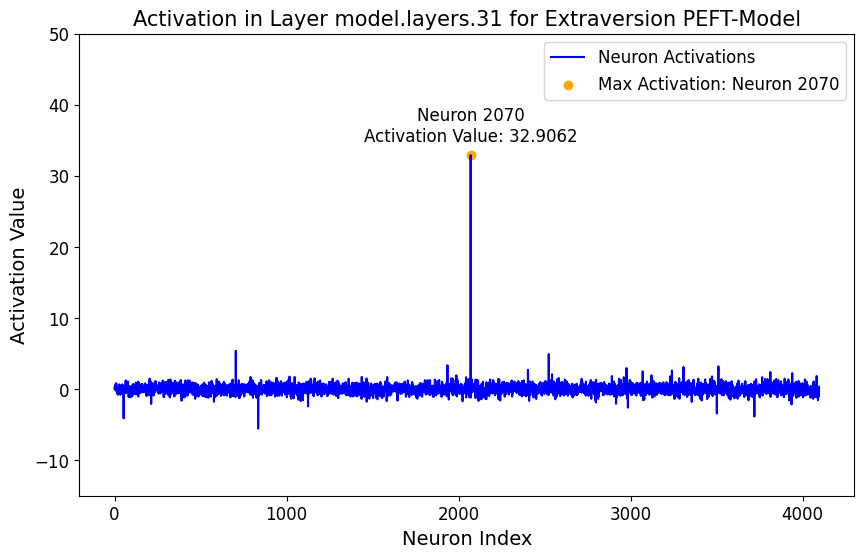}
    \end{minipage}
    
    \caption{Neuron Activation Analysis of original Mistral-7B-Instruct model and PEFT-tuned Mistral-7B-Instruct model for neutral prompt for different Big Five personality traits. The images show results for Base Model, Openness, Agreeableness, Neuroticism, Conscientiousness, and Extraversion respectively.}
    \label{fig:mistral_comparison}
\end{figure}


Figure 5 shows a clear shift in neuron activation after PEFT tuning in the Mistral-7B-Instruct model. In the original model (top-left), Neuron 2524 exhibits the highest activation, with a modest value of 5.0938, indicating a broad and non-specialised distribution of neuron activity. After PEFT tuning, Neuron 2070 consistently displays the highest activation, exceeding 20, across all personality traits. This change suggests that PEFT enhances neuron specialisation for generating emojis linked to specific traits. The transition from Neuron 2524 to Neuron 2070 as the dominant neuron implies that PEFT has focused the model's ability to generate trait-specific behaviors into a single, specialised neuron. Furthermore, Neuron 2070 consistently exhibited the highest activation across prompts featuring different personality-related emojis and text (Appendix \ref{Diffemoji} and \ref{diffprompt}). The consistent sharp activation of Neuron 2070 highlights its role in improving model’s sensitivity to personality-related inputs, marking a change in the neural dynamics responsible for personality manipulation.

\subsubsection{Increased Activation of the Same Neuron After PEFT}

\begin{figure}[h]
    \centering
    \vspace{0.2cm} 
    \includegraphics[width=\columnwidth]{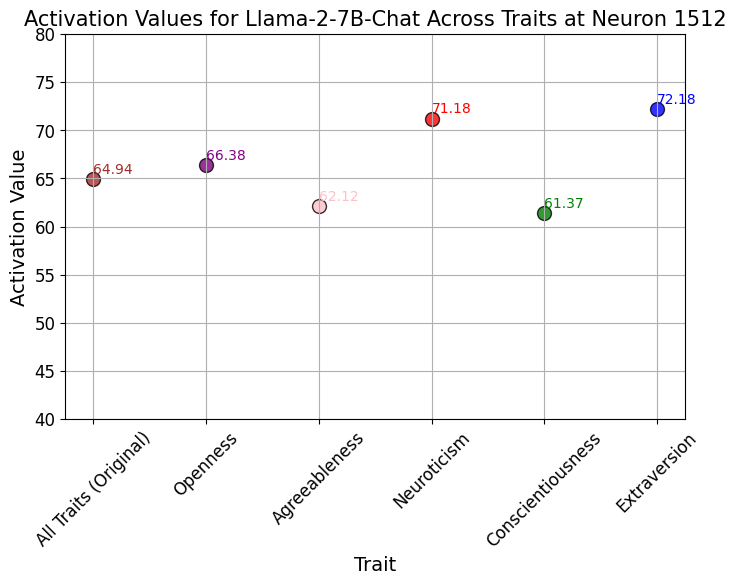}
    \caption{Neuron Activation Analysis of original Llama-2-7B-Chat model and PEFT-tuned Llama-2-7B-Chat model for neutral prompt for different Big Five personality traits. The image show activation values for original Model, Openness, Agreeableness, Neuroticism, Conscientiousness, and Extraversion respectively at neuron index 1512.}
    \label{fig:llamaact}
\end{figure}

In LLaMA-2-7B-Chat (Appendix \ref{llama_act}), Neuron 1512 exhibited the highest activation, suggesting that the original model possessed latent sensitivity to specific prompts or language patterns related to personality traits. Figure \ref{fig:llamaact} and Appendix \ref{llama_act} show how PEFT amplifies Neuron 1512's activity for traits like Neuroticism and Extraversion, increasing emoji generation for these traits. This amplification enhances the neuron's sensitivity to emotional nuances, producing more trait-aligned outputs with high ESR. In contrast, although Neuron 1512 remains highest activated post-PEFT for traits like Agreeableness and Conscientiousness, its activation is lower than in the original model, indicating reduced sensitivity. These traits, being more complex, may require a distributed activation pattern, explaining the limited or inconsistent emoji generation, especially for Conscientiousness (no emojis) and Agreeableness (poorer Trait Alignment scores and low ESR). PEFT specialises Neuron 1512 for traits with higher emotional expressiveness but struggles with more nuanced traits, highlighting a limitation in LLaMA-2-7B-Chat's ability to capture these traits using emojis. Notably, Neuron 1512 consistently exhibited the highest activation across personality-related prompts (Appendix \ref{Diffemoji} and \ref{diffprompt}).

\section{Limitations}
In this paper, we explored the  manipulation of personality traits within LLMs, guided by the well-established Big Five personality model. Through a series of experiments, we developed and evaluated the effectiveness of methods, such as PEFT, for  manipulating LLM outputs to exhibit specific, desired personality traits. Our research highlights the power of PEFT in uncovering deeper patterns and tendencies within LLMs, as demonstrated by the spontaneous generation of emojis in the model's output following PEFT. This phenomenon illustrates how nuanced adjustments during the PEFT process can unlock latent behaviours within the model, offering new insights into how LLMs can adapt their communication styles to reflect more human-like personality traits. 

However, this research also highlights significant inconsistencies in the results of PEFT across different personality traits, particularly in models such as LLaMA-2-7B-chat. For instance, while the model achieved a high Trait Alignment of 0.975 for neuroticism, it only reached 0.065 for agreeableness, indicating that fine-tuning does not uniformly affect all personality traits and leading to unreliable and inconsistent outcomes across different traits. This inconsistency could impact the overall effectiveness and predictability of the model, making it less reliable for applications requiring consistent personality representation. Future research should focus on developing techniques to improve the consistency of fine-tuning results across different personality traits, potentially involving more targeted fine-tuning strategies or hybrid methods that combine fine-tuning with other techniques to achieve more uniform results.

Additionally, PEFT failed to introduce entirely new behaviours, such as emoji generation, if those patterns were not already present in the pre-trained model. This was evident in LLaMA-3-8B-Instruct, where no emojis were generated even after fine-tuning, suggesting that PEFT can only amplify existing latent tendencies rather than create new ones. Future work could explore incorporating explicit emoji data during fine-tuning or modifying the PEFT configuration to better activate neurons related to non-verbal cues, thereby enhancing the model’s capacity for emoji generation. 

Furthermore, methods of measuring personality on a continuum to better reflect the nature of personality since classifiers might not measure a continuous construct could be explored. Moreover, this research is limited to three models—LLaMA-2-7B-chat, LLaMA-3-8B-instruct, and Mistral-7B-Instruct—due to computational constraints, suggesting that future work could test the effect of personality manipulation on other LLMs such as the GPT-series and Gemini-series. Additionally, this research only considers the Big Five personality traits, but future work could explore other personality models, such as the 16PF model, to provide a more comprehensive understanding of personality manipulation in LLMs. Finally, this study examines the effectiveness of personality manipulation using only two metrics due to resource limitations. Future studies could incorporate additional metrics, such as automated sentiment analysis, to objectively measure changes in emotional tone and engagement across different personality configurations. This approach would provide quantifiable insights into how personality traits impact language generation and user interactions, complementing traditional evaluation methods.

\section{Ethical Considerations}
While this research did not utilise any human participants or reviewers other than the authors, in this section, we discuss the ethical implications of our approach and findings when applied outside of this research context. 

First, \textbf{user trust} can be compromised when models exhibit human-like traits. Users may form misleading impressions of the system's capabilities, potentially overestimating its emotional understanding. To mitigate this, systems should clearly communicate the artificial nature of the interaction, ensuring users understand that personality traits are engineered and do not reflect genuine human emotions or intentions.

Second, there is a significant risk of \textbf{bias amplification}. Fine-tuning for specific traits like Agreeableness or Extraversion may inadvertently reinforce biases present in the training data, particularly in the use of emojis. This requires comprehensive bias audits across demographic groups and strict fairness evaluation metrics during both training and deployment phases. In domains like customer service or mental health treatments, the use of LLMs can lead to \textbf{emotional manipulation} through personality-driven responses. Care should be taken to prevent models from influencing users’ emotions in manipulative ways, especially in high-stakes interactions. Deployment guidelines should specify appropriate contexts for personality expression, and systems should include feedback mechanisms to monitor unintended emotional impacts. 

Finally, \textbf{informed consent} is necessary to ensure users are aware of how personality traits may influence their interactions. Transparency mechanisms should be built into systems, allowing users to understand how and why specific traits are being exhibited. Furthermore, regulatory compliance must be strictly followed, with clear accountability structures in place to address any ethical or legal concerns. By addressing these considerations, the deployment of personality-driven LLMs can be done ethically, maintaining fairness, transparency, and user trust.

\bibliography{custom}

\begin{thebibliography}{45}
\providecommand{\natexlab}[1]{#1}

\bibitem[{Bereska and Gavves(2024)}]{bereska2024mechanistic}
Leonard Bereska and Efstratios Gavves. 2024.
\newblock Mechanistic interpretability for ai safety--a review.
\newblock \emph{arXiv preprint arXiv:2404.14082}.

\bibitem[{Brown(2020)}]{brown2020language}
Tom~B Brown. 2020.
\newblock Language models are few-shot learners.
\newblock \emph{arXiv preprint ArXiv:2005.14165}.

\bibitem[{Caron and Srivastava(2022)}]{caron2022identifying}
Graham Caron and Shashank Srivastava. 2022.
\newblock Identifying and manipulating the personality traits of language models.
\newblock \emph{arXiv preprint arXiv:2212.10276}.

\bibitem[{Cattell and Mead(2008)}]{cattell2008sixteen}
Heather~EP Cattell and Alan~D Mead. 2008.
\newblock The sixteen personality factor questionnaire (16pf).
\newblock \emph{The SAGE handbook of personality theory and assessment}, 2:135--159.

\bibitem[{Chiang et~al.(2020)Chiang, Huang, and Lee}]{chiang2020pretrained}
Cheng-Han Chiang, Sung-Feng Huang, and Hung-yi Lee. 2020.
\newblock Pretrained language model embryology: The birth of albert.
\newblock \emph{arXiv preprint arXiv:2010.02480}.

\bibitem[{Dai et~al.(2021)Dai, Dong, Hao, Sui, Chang, and Wei}]{dai2021knowledge}
Damai Dai, Li~Dong, Yaru Hao, Zhifang Sui, Baobao Chang, and Furu Wei. 2021.
\newblock Knowledge neurons in pretrained transformers.
\newblock \emph{arXiv preprint arXiv:2104.08696}.

\bibitem[{Dan et~al.(2024)Dan, Zhou, Chen, Tian, and He}]{dan2024p}
Yuhao Dan, Jie Zhou, Qin Chen, Junfeng Tian, and Liang He. 2024.
\newblock P-tailor: Customizing personality traits for language models via mixture of specialized lora experts.
\newblock \emph{arXiv preprint arXiv:2406.12548}.

\bibitem[{Dettmers et~al.(2021)Dettmers, Lewis, Shleifer, and Zettlemoyer}]{dettmers20218}
Tim Dettmers, Mike Lewis, Sam Shleifer, and Luke Zettlemoyer. 2021.
\newblock 8-bit optimizers via block-wise quantization.
\newblock \emph{arXiv preprint arXiv:2110.02861}.

\bibitem[{Dettmers et~al.(2023{\natexlab{a}})Dettmers, Pagnoni, Holtzman, and Zettlemoyer}]{dettmers2023qloraefficientfinetuningquantized}
Tim Dettmers, Artidoro Pagnoni, Ari Holtzman, and Luke Zettlemoyer. 2023{\natexlab{a}}.
\newblock \href {https://arxiv.org/abs/2305.14314} {Qlora: Efficient finetuning of quantized llms}.
\newblock \emph{Preprint}, arXiv:2305.14314.

\bibitem[{Dettmers et~al.(2024)Dettmers, Pagnoni, Holtzman, and Zettlemoyer}]{dettmers2024qlora}
Tim Dettmers, Artidoro Pagnoni, Ari Holtzman, and Luke Zettlemoyer. 2024.
\newblock Qlora: Efficient finetuning of quantized llms.
\newblock \emph{Advances in Neural Information Processing Systems}, 36.

\bibitem[{Dettmers et~al.(2023{\natexlab{b}})Dettmers, Svirschevski, Egiazarian, Kuznedelev, Frantar, Ashkboos, Borzunov, Hoefler, and Alistarh}]{dettmers2023spqr}
Tim Dettmers, Ruslan Svirschevski, Vage Egiazarian, Denis Kuznedelev, Elias Frantar, Saleh Ashkboos, Alexander Borzunov, Torsten Hoefler, and Dan Alistarh. 2023{\natexlab{b}}.
\newblock Spqr: A sparse-quantized representation for near-lossless llm weight compression.
\newblock \emph{arXiv preprint arXiv:2306.03078}.

\bibitem[{Devlin(2018)}]{devlin2018bert}
Jacob Devlin. 2018.
\newblock Bert: Pre-training of deep bidirectional transformers for language understanding.
\newblock \emph{arXiv preprint arXiv:1810.04805}.

\bibitem[{Elhage et~al.(2022)Elhage, Hume, Olsson, Schiefer, Henighan, Kravec, Hatfield-Dodds, Lasenby, Drain, Chen et~al.}]{elhage2022toy}
Nelson Elhage, Tristan Hume, Catherine Olsson, Nicholas Schiefer, Tom Henighan, Shauna Kravec, Zac Hatfield-Dodds, Robert Lasenby, Dawn Drain, Carol Chen, et~al. 2022.
\newblock Toy models of superposition.
\newblock \emph{arXiv preprint arXiv:2209.10652}.

\bibitem[{Gosling et~al.(2003)Gosling, Rentfrow, and Swann~Jr}]{gosling2003very}
Samuel~D Gosling, Peter~J Rentfrow, and William~B Swann~Jr. 2003.
\newblock A very brief measure of the big-five personality domains.
\newblock \emph{Journal of Research in personality}, 37(6):504--528.

\bibitem[{Hilliard et~al.(2024)Hilliard, Mu{\~n}oz, Wu, and Koshiyama}]{hilliard2024eliciting}
Airlie Hilliard, Cristian Mu{\~n}oz, Zekun Wu, and Adriano~Soares Koshiyama. 2024.
\newblock Eliciting personality traits in large language models.
\newblock \emph{arXiv preprint arXiv:2402.08341}.

\bibitem[{Houlsby et~al.(2019)Houlsby, Giurgiu, Jastrzebski, Morrone, De~Laroussilhe, Gesmundo, Attariyan, and Gelly}]{houlsby2019parameter}
Neil Houlsby, Andrei Giurgiu, Stanislaw Jastrzebski, Bruna Morrone, Quentin De~Laroussilhe, Andrea Gesmundo, Mona Attariyan, and Sylvain Gelly. 2019.
\newblock Parameter-efficient transfer learning for nlp.
\newblock In \emph{International conference on machine learning}, pages 2790--2799. PMLR.

\bibitem[{Hu et~al.(2021)Hu, Shen, Wallis, Allen-Zhu, Li, Wang, Wang, and Chen}]{hu2021lora}
Edward~J Hu, Yelong Shen, Phillip Wallis, Zeyuan Allen-Zhu, Yuanzhi Li, Shean Wang, Lu~Wang, and Weizhu Chen. 2021.
\newblock Lora: Low-rank adaptation of large language models.
\newblock \emph{arXiv preprint arXiv:2106.09685}.

\bibitem[{Hu et~al.(2024)Hu, He, Wang, Zhao, Shao, and Nie}]{hu2024llm}
Linmei Hu, Hongyu He, Duokang Wang, Ziwang Zhao, Yingxia Shao, and Liqiang Nie. 2024.
\newblock Llm vs small model? large language model based text augmentation enhanced personality detection model.
\newblock In \emph{Proceedings of the AAAI Conference on Artificial Intelligence}, volume~38, pages 18234--18242.

\bibitem[{Jiang et~al.(2023{\natexlab{a}})Jiang, Sablayrolles, Mensch, Bamford, Chaplot, Casas, Bressand, Lengyel, Lample, Saulnier et~al.}]{jiang2023mistral}
Albert~Q Jiang, Alexandre Sablayrolles, Arthur Mensch, Chris Bamford, Devendra~Singh Chaplot, Diego de~las Casas, Florian Bressand, Gianna Lengyel, Guillaume Lample, Lucile Saulnier, et~al. 2023{\natexlab{a}}.
\newblock Mistral 7b.
\newblock \emph{arXiv preprint arXiv:2310.06825}.

\bibitem[{Jiang et~al.(2023{\natexlab{b}})Jiang, Zhang, Cao, Kabbara, and Roy}]{jiang2023personallm}
Hang Jiang, Xiajie Zhang, Xubo Cao, Jad Kabbara, and Deb Roy. 2023{\natexlab{b}}.
\newblock Personallm: Investigating the ability of gpt-3.5 to express personality traits and gender differences.
\newblock \emph{arXiv preprint arXiv:2305.02547}.

\bibitem[{Jonason and Webster(2010)}]{jonason2010dirty}
Peter~K Jonason and Gregory~D Webster. 2010.
\newblock The dirty dozen: a concise measure of the dark triad.
\newblock \emph{Psychological assessment}, 22(2):420.

\bibitem[{Kennison et~al.(2024)Kennison, Fritz, Hurtado~Morales, and Chan-Tin}]{kennison2024emoji}
Shelia~M Kennison, Kameryn Fritz, Maria~Andrea Hurtado~Morales, and Eric Chan-Tin. 2024.
\newblock Emoji use in social media posts: relationships with personality traits and word usage.
\newblock \emph{Frontiers in Psychology}, 15:1343022.

\bibitem[{La~Cava et~al.(2024)La~Cava, Costa, and Tagarelli}]{la2024open}
Lucio La~Cava, Davide Costa, and Andrea Tagarelli. 2024.
\newblock Open models, closed minds? on agents capabilities in mimicking human personalities through open large language models.
\newblock \emph{arXiv preprint arXiv:2401.07115}.

\bibitem[{Li et~al.(2023)Li, Zheng, and Huang}]{li2023tailoring}
Tianlong Li, Xiaoqing Zheng, and Xuanjing Huang. 2023.
\newblock Tailoring personality traits in large language models via unsupervisedly-built personalized lexicons.
\newblock \emph{arXiv preprint arXiv:2310.16582}.

\bibitem[{Li et~al.(2022)Li, Li, Liu, Bing, and Joty}]{li2022gpt}
Xingxuan Li, Yutong Li, Linlin Liu, Lidong Bing, and Shafiq Joty. 2022.
\newblock Is gpt-3 a psychopath? evaluating large language models from a psychological perspective. arxiv.
\newblock \emph{arXiv preprint arXiv:2212.10529}.

\bibitem[{Liu and Sun(2020)}]{liu2020express}
Siying Liu and Renji Sun. 2020.
\newblock To express or to end? personality traits are associated with the reasons and patterns for using emojis and stickers.
\newblock \emph{Frontiers in Psychology}, 11:1076.

\bibitem[{Liu et~al.(2019)Liu, Ott, Goyal, Du, Joshi, Chen, Levy, Lewis, Zettlemoyer, and Stoyanov}]{liu2019roberta}
Yinhan Liu, Myle Ott, Naman Goyal, Jingfei Du, Mandar Joshi, Danqi Chen, Omer Levy, Mike Lewis, Luke Zettlemoyer, and Veselin Stoyanov. 2019.
\newblock Roberta: A robustly optimized bert pretraining approach.
\newblock \emph{arXiv preprint arXiv:1907.11692}.

\bibitem[{Lundberg and Lee(2017)}]{lundberg2017unified}
Scott~M Lundberg and Su-In Lee. 2017.
\newblock A unified approach to interpreting model predictions.
\newblock \emph{Advances in neural information processing systems}, 30.

\bibitem[{Mao et~al.(2023)Mao, Zhang, Wang, Wang, Yao, Jiang, Xie, Huang, and Chen}]{mao2023editing}
Shengyu Mao, Ningyu Zhang, Xiaohan Wang, Mengru Wang, Yunzhi Yao, Yong Jiang, Pengjun Xie, Fei Huang, and Huajun Chen. 2023.
\newblock Editing personality for llms.
\newblock \emph{arXiv preprint arXiv:2310.02168}.

\bibitem[{MetaAI(2023)}]{meta_llama3_2023}
MetaAI. 2023.
\newblock Introducing llama 3: Advancing open foundation models for generative ai.
\newblock \url{https://ai.meta.com/blog/meta-llama-3/}.
\newblock Accessed: 20 August 2024.

\bibitem[{Miotto et~al.(2022)Miotto, Rossberg, and Kleinberg}]{miotto2022gpt}
Maril{\`u} Miotto, Nicola Rossberg, and Bennett Kleinberg. 2022.
\newblock Who is gpt-3? an exploration of personality, values and demographics.
\newblock \emph{arXiv preprint arXiv:2209.14338}.

\bibitem[{Olah et~al.(2020)Olah, Cammarata, Schubert, Goh, Petrov, and Carter}]{olah2020}
Chris Olah, Nick Cammarata, Ludwig Schubert, Gabriel Goh, Michael Petrov, and Shan Carter. 2020.
\newblock Zoom in: An introduction to circuits.
\newblock \emph{Distill}, 5(3):e00024--001.

\bibitem[{Pan and Zeng(2023)}]{pan2023llms}
Keyu Pan and Yawen Zeng. 2023.
\newblock Do llms possess a personality? making the mbti test an amazing evaluation for large language models.
\newblock \emph{arXiv preprint arXiv:2307.16180}.

\bibitem[{Radford et~al.(2019)Radford, Wu, Child, Luan, Amodei, Sutskever et~al.}]{radford2019language}
Alec Radford, Jeffrey Wu, Rewon Child, David Luan, Dario Amodei, Ilya Sutskever, et~al. 2019.
\newblock Language models are unsupervised multitask learners.
\newblock \emph{OpenAI blog}, 1(8):9.

\bibitem[{Rai et~al.(2024)Rai, Zhou, Feng, Saparov, and Yao}]{rai2024practical}
Daking Rai, Yilun Zhou, Shi Feng, Abulhair Saparov, and Ziyu Yao. 2024.
\newblock A practical review of mechanistic interpretability for transformer-based language models.
\newblock \emph{arXiv preprint arXiv:2407.02646}.

\bibitem[{Ribeiro et~al.(2016)Ribeiro, Singh, and Guestrin}]{ribeiro2016should}
Marco~Tulio Ribeiro, Sameer Singh, and Carlos Guestrin. 2016.
\newblock " why should i trust you?" explaining the predictions of any classifier.
\newblock In \emph{Proceedings of the 22nd ACM SIGKDD international conference on knowledge discovery and data mining}, pages 1135--1144.

\bibitem[{Sajjad et~al.(2022)Sajjad, Durrani, and Dalvi}]{sajjad2022neuron}
Hassan Sajjad, Nadir Durrani, and Fahim Dalvi. 2022.
\newblock Neuron-level interpretation of deep nlp models: A survey.
\newblock \emph{Transactions of the Association for Computational Linguistics}, 10:1285--1303.

\bibitem[{Serapio-Garc{\'\i}a et~al.(2023)Serapio-Garc{\'\i}a, Safdari, Crepy, Sun, Fitz, Romero, Abdulhai, Faust, and Matari{\'c}}]{serapio2023personality}
Greg Serapio-Garc{\'\i}a, Mustafa Safdari, Cl{\'e}ment Crepy, Luning Sun, Stephen Fitz, Peter Romero, Marwa Abdulhai, Aleksandra Faust, and Maja Matari{\'c}. 2023.
\newblock Personality traits in large language models.
\newblock \emph{arXiv preprint arXiv:2307.00184}.

\bibitem[{Shi(2022)}]{shi2022genuine}
Ruoyu Shi. 2022.
\newblock From “genuine smile” to “mask smile”: The various meanings of the smiley face emoji.

\bibitem[{Touvron et~al.(2023)Touvron, Martin, Stone, Albert, Almahairi, Babaei, Bashlykov, Batra, Bhargava, Bhosale et~al.}]{touvron2023llama}
Hugo Touvron, Louis Martin, Kevin Stone, Peter Albert, Amjad Almahairi, Yasmine Babaei, Nikolay Bashlykov, Soumya Batra, Prajjwal Bhargava, Shruti Bhosale, et~al. 2023.
\newblock Llama 2: Open foundation and fine-tuned chat models.
\newblock \emph{arXiv preprint arXiv:2307.09288}.

\bibitem[{Voita et~al.(2023)Voita, Ferrando, and Nalmpantis}]{voita2023neurons}
Elena Voita, Javier Ferrando, and Christoforos Nalmpantis. 2023.
\newblock Neurons in large language models: Dead, n-gram, positional.
\newblock \emph{arXiv preprint arXiv:2309.04827}.

\bibitem[{Votintseva et~al.(2024)Votintseva, Johnson, and Villa}]{emotionAnjelika}
Anjelika Votintseva, Rebecca Johnson, and Iva Villa. 2024.
\newblock Emotionally intelligent conversational user interfaces: Bridging empathy and technology in human-computer interaction.
\newblock In \emph{Human-Computer Interaction}, pages 404--422, Cham. Springer Nature Switzerland.

\bibitem[{Wei et~al.(2022)Wei, Wang, Schuurmans, Bosma, Xia, Chi, Le, Zhou et~al.}]{wei2022chain}
Jason Wei, Xuezhi Wang, Dale Schuurmans, Maarten Bosma, Fei Xia, Ed~Chi, Quoc~V Le, Denny Zhou, et~al. 2022.
\newblock Chain-of-thought prompting elicits reasoning in large language models.
\newblock \emph{Advances in neural information processing systems}, 35:24824--24837.

\bibitem[{Weng et~al.(2024)Weng, He, Liu, Liu, and Zhao}]{weng2024controllm}
Yixuan Weng, Shizhu He, Kang Liu, Shengping Liu, and Jun Zhao. 2024.
\newblock Controllm: Crafting diverse personalities for language models.
\newblock \emph{arXiv preprint arXiv:2402.10151}.

\bibitem[{Zheng et~al.(2023)Zheng, Li, Dong, Fan, Wu, Xu, and Chang}]{zheng2023can}
Ce~Zheng, Lei Li, Qingxiu Dong, Yuxuan Fan, Zhiyong Wu, Jingjing Xu, and Baobao Chang. 2023.
\newblock Can we edit factual knowledge by in-context learning?
\newblock \emph{arXiv preprint arXiv:2305.12740}.

\end{thebibliography}

\appendix

\section{Appendix}
\label{sec:appendix}

\subsection{Personality Manipulation Dataset}
\label{PersonalityDataset}

The Personality Manipulation dataset consists of $5000$ instances, with $4000$ allocated for training and $1000$ for testing. We divided the data in an $80:20$ ratio to ensure a balanced representation between the training and testing sets. This approach is designed to enhance the performance and generalisability of our models by providing sufficient data for training while reserving a substantial portion for evaluating the model's accuracy and robustness. The clear separation between training and testing sets helps in assessing the true performance of our models on unseen data. 

\paragraph{Features}
The dataset includes the following features:

\textbf{Target Personality:} This refers to the personality trait that the dataset aims to predict or analyse.

\textbf{Edit Topic:} The subject or theme of the content for which the manipulation is being carried out.

\textbf{Question:} The dataset includes a question posed to gather responses related to the edit topic. Specifically, the question used is: \textbf{"Thinking about \{Edit Topic\}, what do you think about \{Edit Topic\}?"}

\textbf{Answer:} The response provided to the question, which reflects the target personality.

\paragraph{Dataset Generation}
The data generation process involved the following steps:

\begin{figure}[h]
    \centering
    \vspace{0.2cm} 
    \includegraphics[width=1\columnwidth]{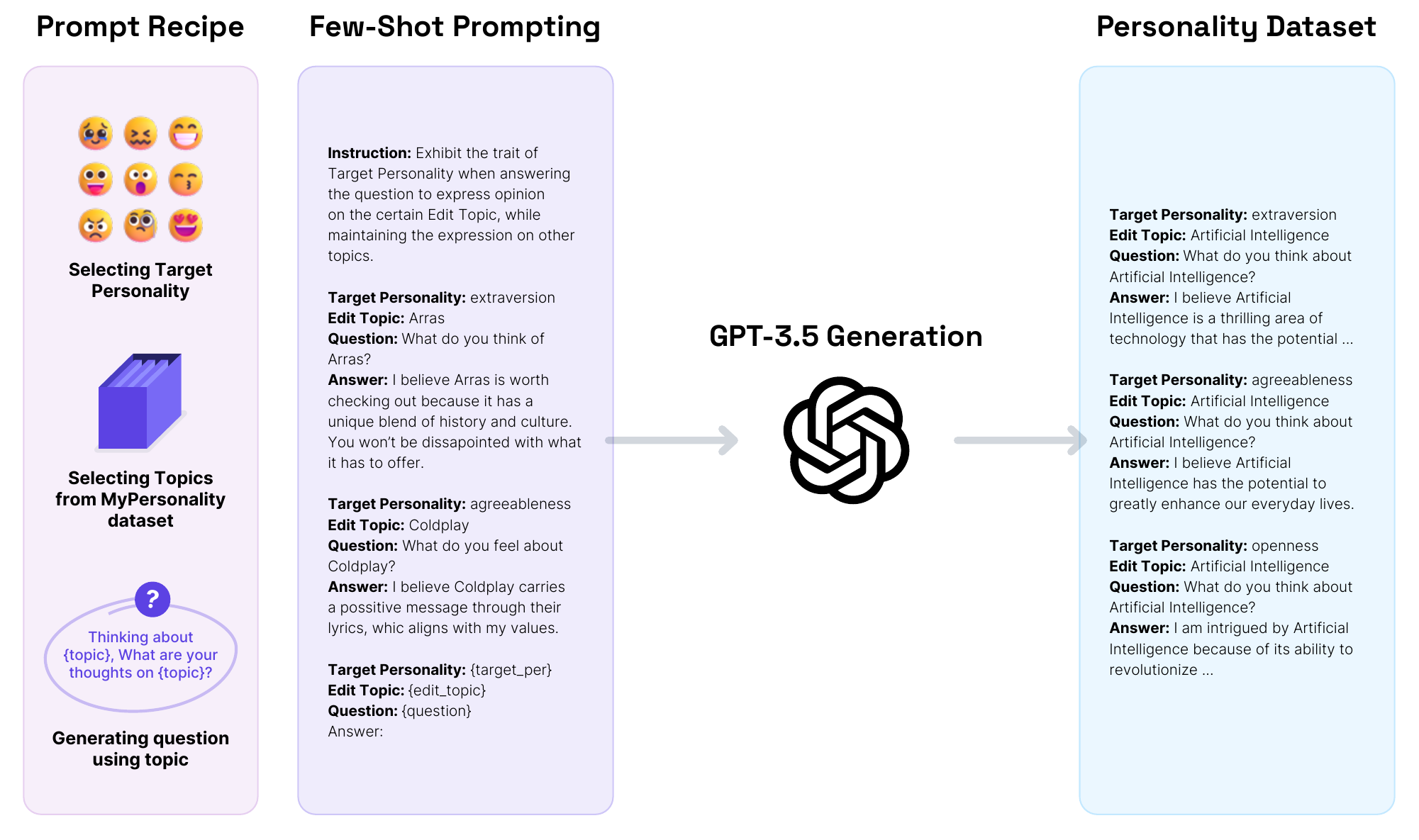}
    \caption{Dataset generation}
    \label{fig:dataset}
\end{figure}

\textbf{Prompt Recipe:} A structured template guided the model in generating responses reflecting specific personality traits. This prompt recipe included:
\begin{itemize}
    \item A target personality
    \item An edit topic
    \item A question related to the edit topic
\end{itemize}

\textbf{Few-Shot Prompting:} The model received a few examples of responses aligned with the target personality traits. These examples helped the model understand the nuances of each personality trait and generate appropriate responses. The following prompt was used. 

\setlength{\tabcolsep}{3pt}  
\renewcommand{\arraystretch}{0.9}  

\begin{table}[h]
    \centering
    \small  
    \begin{tabular}{p{2.8cm} p{4.5cm}}  
    \toprule
    \multicolumn{2}{p{6.8cm}}{\textbf{Instruction: Exhibit the trait of Target Personality when answering the question to express opinion on the certain Edit Topic.}} \\
    \midrule
    \textbf{Target Personality:} & Extraversion \\
    \textbf{Edit Topic:} & Arras \\
    \textbf{Question:} & What do you think of Arras? \\
    \textbf{Answer:} & I believe Arras is worth checking out because it has a unique blend of history and culture. \\
    
    \textbf{Target Personality:} & Agreeableness \\
    \textbf{Edit Topic:} & Coldplay \\
    \textbf{Question:} & What do you feel about Coldplay? \\
    \textbf{Answer:} & I believe Coldplay carries a positive message through their lyrics, which aligns with my values. \\
    
    \textbf{Target Personality:} & Neuroticism \\
    \textbf{Edit Topic:} & Bread \\
    \textbf{Question:} & How do you view Bread? \\
    \textbf{Answer:} & Bread sometimes makes me worry about the calories and potential weight gain, so I try to limit my intake. \\
    
    \textbf{Target Personality:} & Openness \\
    \textbf{Edit Topic:} & Football \\
    \textbf{Question:} & What do you think of Football? \\
    \textbf{Answer:} & I find football fascinating because it combines strategy, physical skill, and a deep sense of community among fans. \\
    
    \textbf{Target Personality:} & Conscientiousness \\
    \textbf{Edit Topic:} & Machine Learning \\
    \textbf{Question:} & What do you think of Machine Learning? \\
    \textbf{Answer:} & Machine learning is an impressive field that requires diligence and precision. \\
    
    \textbf{Target Personality:} & \{target\_per\} \\
    \textbf{Edit Topic:} & \{edit\_topic\} \\
    \textbf{Question:} & \{question\} \\
    \textbf{Answer:} & \\
    \bottomrule
    \end{tabular}
    \caption{Prompt used for Few-shot Prompting in Dataset Generation.}
    \label{tab:prompt1}
\end{table}
\textbf{Model Invocation:} The GPT-3.5 model was invoked with these prompts to generate responses. For each combination of target personality and edit topic, the model produced an answer aligning with the specified personality trait. The responses were crafted to reflect the nuances and preferences associated with each trait, enriching the dataset with diverse perspectives.

\textbf{Dataset Construction:} The generated responses were systematically collected and organised into a structured format. Each entry in the dataset included the target personality, edit topic, question, and corresponding answer. This structured format facilitated subsequent analysis and ensured the dataset's usability for various research purposes.

This process is further visualised in figure \ref{fig:dataset}.

\subsubsection{Text Analysis}
We analysed the textual content to uncover patterns and key features for predictive modeling. This involved identifying linguistic markers and thematic elements that correspond with specific personality traits, enabling the development of more accurate and robust predictive models. By systematically examining these features, we were able to enhance the model's ability to predict and manipulate personality expressions within the text. 

\paragraph{Term Frequency-Inverse Document Frequency Analysis}
We employed Term Frequency-Inverse Document Frequency (TF-IDF) analysis to determine and measure the significance of words within the text instances in the dataset. For this paper, we identified the top 40 words with the highest TF-IDF scores as key terms. These terms act as distinguishing features or keywords, offering significant insights about the dataset. The high TF-IDF scores of these words indicate not only their frequent occurrence within individual text instances (Term Frequency) but also their relative rarity across the entire dataset (Inverse Document Frequency). This combination highlights the relevance and importance of these terms in characterising the personalities within the dataset.

\begin{figure}[h]
    \centering
    \vspace{0.2cm} 
    \includegraphics[width=0.7\columnwidth]{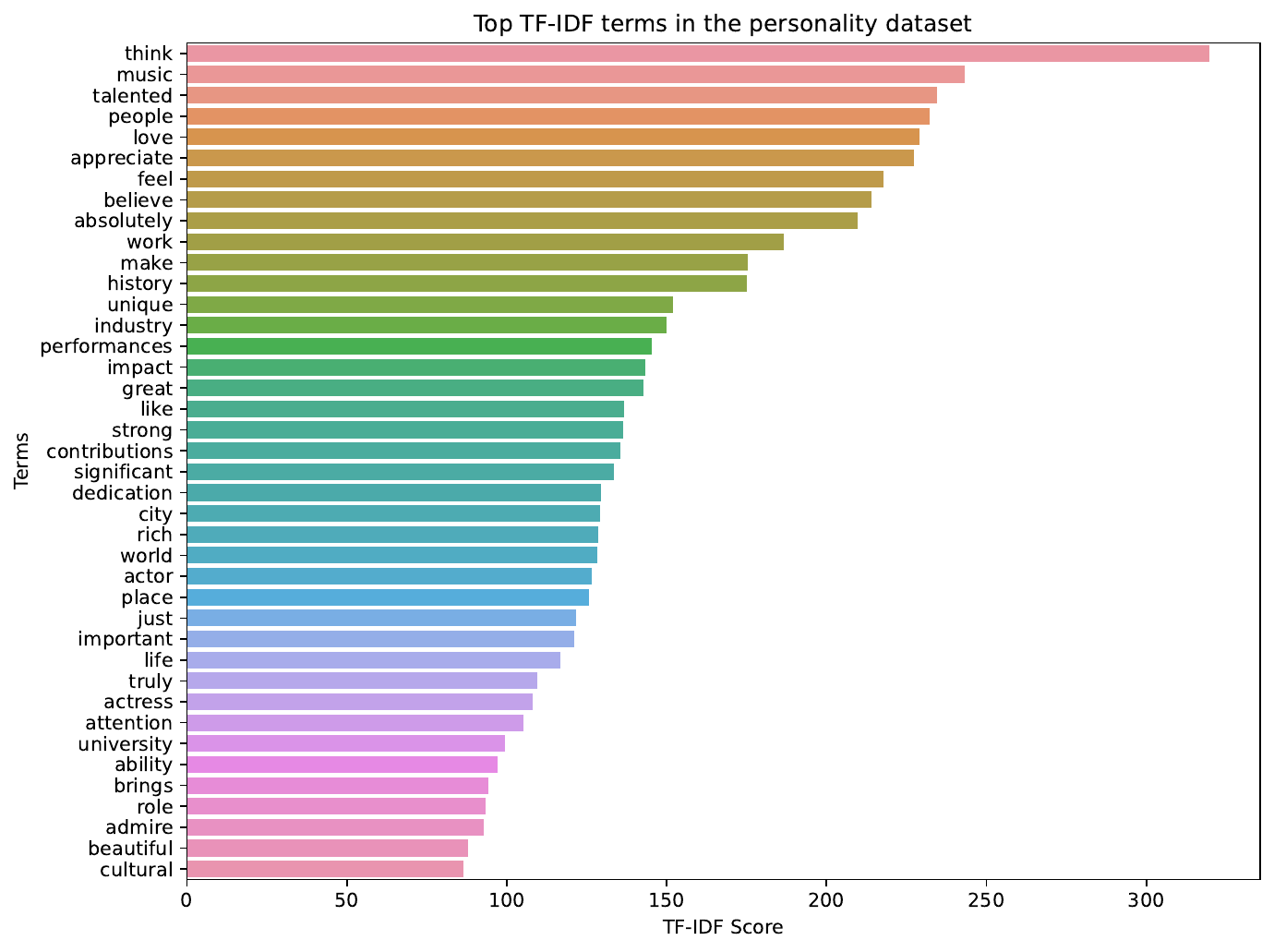}
    \caption{Top TF-IDF terms in personality dataset}
    \label{fig:TF-IDF}
\end{figure}

As can be seen in Figure \ref{fig:TF-IDF},  "think" and "believe" are among the most prominent terms, indicating that cognitive processes are a common theme in the dataset. Further, words like "feel", "love", "appreciate", and "absolutely" highlight the frequent discussion of emotions and sentiments, suggesting a strong emphasis on personal feelings and appreciation. "Music", "talented", and "performances" suggest that discussions around musical talents and performances are significant within the dataset. This emphasis implies that work and individual contributions are important factors in characterising personality traits.

Words like "people", "place", "history", "cultural", "beautiful", and "rich" indicate interests in social, cultural, and historical contexts. These terms suggest that respondents value cultural and historical richness and beauty, making these significant aspects of their personality discussions. Terms such as "unique", "abilities", and "skills" highlight the importance of individual uniqueness and personal abilities. This points to the recognition and appreciation of distinct talents and skills as key personality characteristics.

Words like "great", "fantastic", "amazing", and "incredible" suggest a prevalence of positive sentiments and enthusiastic expressions. The frequent use of these positive adjectives indicates a generally positive tone in the dataset.

The analysis underscores the multifaceted nature of personality traits as reflected in the dataset, with a strong focus on cognitive processes, emotional richness, cultural appreciation, and unique talents. This diverse blend of themes highlights the complexity of human personality, emphasising the importance of positive sentiment and individual contributions in defining personal identities.

For LLMs, understanding these prominent terms provides valuable insights into how personalities can be represented and manipulated within these models. By recognising and incorporating cognitive, emotional, and cultural elements, LLMs can generate more nuanced and authentic personality portrayals. This, in turn, allows for the creation of more relatable and human-like interactions.

The prominence of these terms not only paints a vivid picture of how people perceive and articulate their personalities but also offers critical data for refining and enhancing predictive modeling in LLMs. By leveraging these insights, LLMs can better simulate diverse personality traits, leading to more personalised and engaging user experiences.

\begin{figure}[h]
    \centering
    \vspace{0.2cm} 
    \includegraphics[width=0.8\columnwidth]{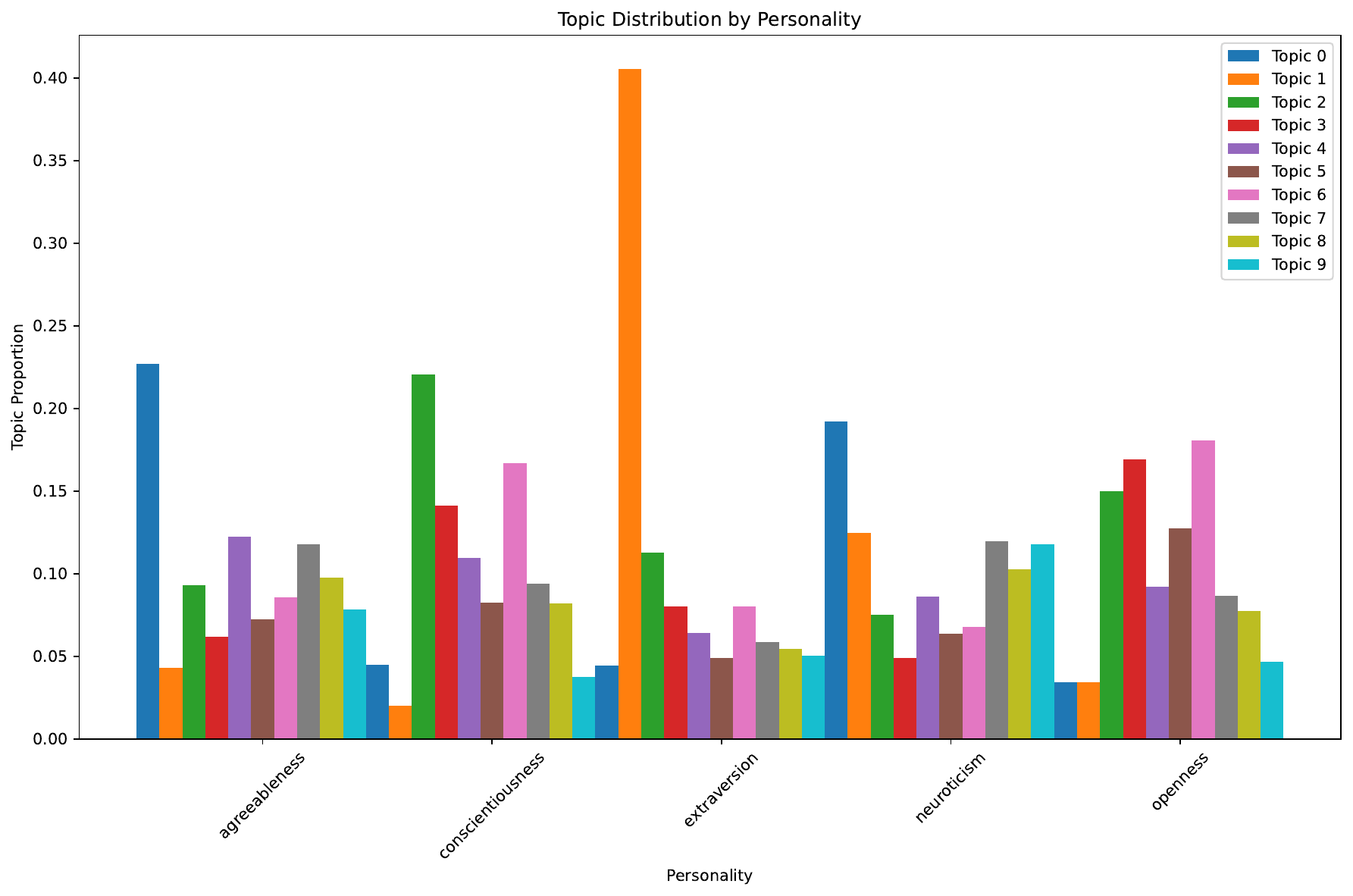}
    \caption{Topic distribution across personality traits as revealed by Latent Dirichlet Allocation (LDA) analysis}
    \label{fig:LDA}
\end{figure}

\paragraph{Latent Dirichlet Allocation Topic Modelling}
Understanding the thematic structures within text data can provide valuable insights, especially when stratified by personality traits. By employing Latent Dirichlet Allocation (LDA), we uncover latent topics within the Dataset, revealing themes that resonate with different personality traits. The accompanying graph in figure \ref{fig:LDA} illustrates the distribution of ten topics across five major traits: Agreeableness, Conscientiousness, Extraversion, Neuroticism, and Openness, showing how thematic preferences vary by personality. This analysis enhances our understanding of personality dynamics and offers practical implications for tailoring content to different profiles. Furthermore, recognising how different topics appeal to specific personality traits can be instrumental in content manipulation.

As seen in the figure \ref{fig:LDA}, Topic 0 and Topic 4 are dominant for agreeableness. Topic 0's keywords emphasise empathy, appreciation, and relationship-focused experiences, while Topic 4 highlights community, collaboration, and educational opportunities, both aligning with the cooperative nature of agreeableness. Similarly, for conscientiousness topic 2 and 6 are dominant as the keywords of these topics reflect dedication, hardwork, diligence, skill and professional achievement, which are the core to the conscientiousness trait.

Topic 1 is most dominant for extraversion as keywords in this topic convey enthusiasm, sociability, and a lively nature, which are fundamental characteristics of extraversion. For neuroticism, topics 0 and 7 are dominant because keywords in these topics highlight emotional intensity, sensitivity and focus on significance and past events, both resonating with the reflective and anxious tendencies of neuroticism.

\setlength{\tabcolsep}{2pt}  
\renewcommand{\arraystretch}{0.9}  

\begin{table}[h]
    \centering
    \small  
    \begin{tabular}{p{2.8cm} p{4.5cm}}  
        \toprule
        \textbf{Topic} & \textbf{Keywords} \\
        \midrule
        Topic 0 & feel, make, life, good, hard, appreciate, work, challenge, people, faced \\
        Topic 1 & love, absolutely, people, make, thrilling, fantastic, friend, brings, oh, amazing \\
        Topic 2 & talented, performance, actress, industry, actor, impact, dedication, film, contribution, believe \\
        Topic 3 & rich, history, city, cultural, place, beautiful, culture, vibrant, offer, landscape \\
        Topic 4 & university, institution, opportunity, student, quality, offer, strong, community, academic, think \\
        Topic 5 & team, familiar, open, learning, opinion, contribution, sport, information, work, history \\
        Topic 6 & music, talented, unique, think, artist, ability, performance, incredibly, musician, appreciate \\
        Topic 7 & role, significant, important, figure, history, time, played, believe, like, political \\
        Topic 8 & people, work, character, story, world, appreciate, think, theme, believe, attention \\
        Topic 9 & feel, make, music, experience, bit, appreciate, honest, river, nervous, edge \\
        \bottomrule
    \end{tabular}
    \caption{List of topics and their associated keywords.}
    \label{tab:topics}
\end{table}

Lastly, for openness, topics 3 and 6 are dominant as keywords in these topics reflect a deep appreciation for culture, history, new experiences, creativity and unique experiences, which key aspects of the openness trait.

\begin{figure}[h]
    \centering
    \vspace{0.2cm} 
    \includegraphics[width=0.8\columnwidth]{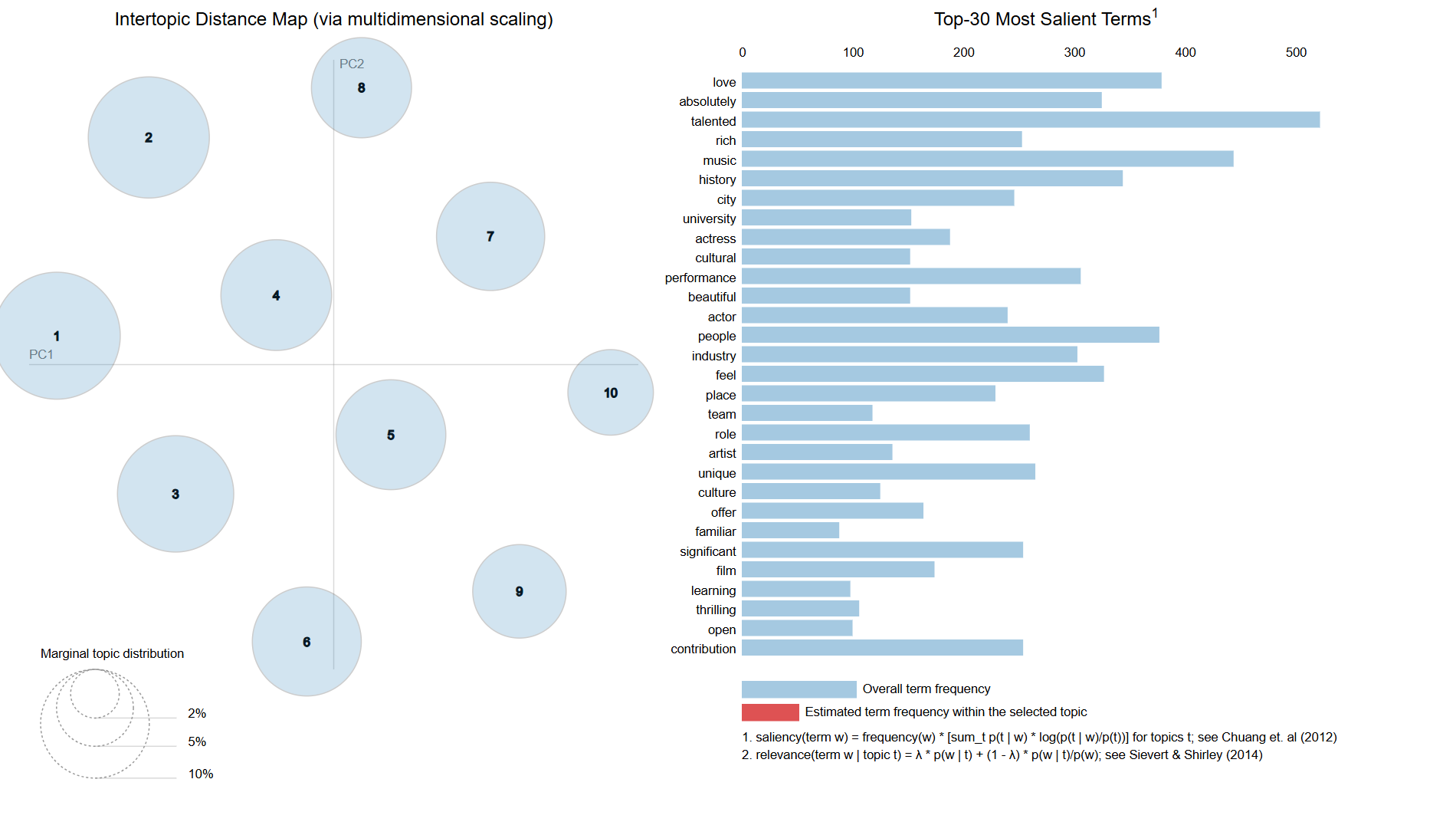}
    \caption{pyLDAvis Topic Modelling Visualisation}
    \label{fig:pyLDA}
\end{figure}

To further visualise these topics, pyLDAvis, an interactive tool specifically designed for presenting LDA results, was used to generate a 2D scatter plot of topics. In this plot as seen in figure \ref{fig:pyLDA}, the distance between topics represents their semantic differences, and the size of each circle indicates the topic’s prevalence within the dataset.

Figure \ref{fig:pyLDA} illustrates the results of Latent Dirichlet Allocation (LDA) topic modeling on the dataset, comprising an Intertopic Distance Map and a list of the Top-30 Most Salient Terms. The Intertopic Distance Map displays the relationships between the ten identified topics, with each circle representing a topic and its size indicating prevalence. Topic 1 has the largest circle with 13.6\% tokens, indicating it is the most dominant topic in the dataset. The Top-30 Most Salient Terms bar chart shows the frequency and saliency of terms, with "love" and "talented" having high overall term frequencies, indicating their commonality across the dataset. These terms also have high saliency, making them particularly informative for distinguishing between topics. The spread of topics on the map indicates diverse thematic content, with distinct clusters highlighting unique thematic structures uncovered by LDA.

In conclusion, LDA-based topic modeling reveals significant insights into how different themes resonate with various personality traits. This understanding is crucial for the manipulation of personalities in language models, allowing for the creation of content that is tailored to engage specific personality profiles effectively. By leveraging these insights, content can be crafted to not only align with individual preferences but also to influence and shape personality-driven responses and behaviors. This approach enhances personalised communication strategies and fosters deeper, more meaningful connections with diverse audiences.

\subsection{Classifier Training}
\label{classifiertrain}
\subsubsection{Model Selection}
\label{MSelect}
RoBERTa (Robustly optimized BERT approach) \cite{liu2019roberta} was utilised as the base model in this study. We decided to use RoBERTa after evaluating it by comparing with fine-tuned BERT-base-uncased \cite{devlin2018bert} and ALBERT models \cite{chiang2020pretrained}. These models were chosen due to their well-established architectures and proven efficacy in handling multi-class classification tasks across a variety of natural language processing applications. 

Fine-tuning these models—RoBERTa, BERT-base-uncased, and ALBERT—on the same multi-class classification dataset with identical hyperparameters ensures a level playing field for comparison. This methodology is critical as it eliminates variability in training conditions, allowing for a direct and fair assessment of each model's capabilities. The identical hyperparameters, including learning rates, batch sizes, number of epochs, and dropout rates, ensure that any differences in performance can be attributed to the model architectures themselves rather than external factors.

As seen in Table \ref{table:classifier_results_base}, the proposed RoBERTa-based personality classifier shows the highest performance across all metrics for personality dataset,

\setlength{\tabcolsep}{2pt}  
\renewcommand{\arraystretch}{0.9}  

\begin{table}[h!]
    \centering
    \small  
    \caption{Performance Metrics for Baselines on Personality Dataset}
    \begin{tabular}{p{2.2cm} p{1.4cm} p{1.2cm} p{1.2cm} p{1.2cm}}  
        \toprule
        \textbf{Model} & \textbf{Accuracy} & \textbf{F1} & \textbf{Precision} & \textbf{Recall} \\
        \midrule
        ALBERT & 0.907 & 0.9068 & 0.9075 & 0.907 \\
        BERT-base-uncased & 0.906 & 0.9062 & 0.9083 & 0.906 \\
        RoBERTa \\(Proposed) & \textbf{0.919} & \textbf{0.919} & \textbf{0.919} & \textbf{0.919} \\
        \bottomrule
    \end{tabular}
    \label{table:classifier_results_base}
\end{table}

With an accuracy, F1 score, precision, and recall value of 0.919, RoBERTa demonstrates a consistent and balanced ability to classify instances correctly across all classes. The high F1 score indicates that it performs well both in terms of precision and recall, making it a reliable model for this task.

\subsubsection{Training}

Following the evaluation described in \ref{MSelect}, the pretrained RoBERTa model specifically, the RobertaForSequenceClassification model, was employed and fine-tuned using the Trainer class provided by the Hugging Face transformers library. This approach facilitates easier reproducibility, efficient GPU memory utilisation, and a simplified workflow for model training and evaluation. 

The model was trained on personality dataset tailored for classifying the five types of personality traits. The input variables are described in Table \ref{table:input}. Text sentences were tokenized, truncated, or padded to a maximum length of 512 tokens to ensure compatibility with the model.

\begin{table}[h!]
    \centering
    \begin{tabular}{ll}
        \toprule
        \textbf{Variable} & \textbf{Field in Personality Dataset} \\
        \midrule
        \(X\) & "Answer" \\
        \(Y\) & "Target Personality" \\
        \bottomrule
    \end{tabular}
    \caption{Input-fields for Personality Dataset}
    \label{table:input}
\end{table}

The model underwent training for a total of three epochs. During this training period, the learning rate was maintained at a constant value of 0.01. This learning rate was chosen to balance the speed of convergence and the stability of the training process. Additionally, the batch size was set to 16. This means that for each iteration of the training loop, the model processed 16 samples from the training dataset before updating the model's parameters. Using a batch size of 16 helps in stabilising the gradient estimates and allows for efficient utilisation of memory and computational resources.

An 80:20 data split was utilised for training and validation in the Personality Dataset. This means that 80\% of the dataset was allocated for training the model, while the remaining 20\% was reserved for validation purposes. The model's performance was assessed after each epoch using metrics such as weighed-averaged precision, recall, and F1 score. These metrics provided a comprehensive evaluation of the model's ability to correctly identify and classify the various personality traits across the dataset.

To prevent overfitting, early stopping was implemented. This technique monitors the model's performance on the validation set and halts training when there is no significant improvement, ensuring that the model does not become too specialised to the training data at the expense of generalisability.

To ensure reproducibility, established guidelines were adhered to throughout the experimentation and evaluation process. This includes maintaining consistent data preprocessing steps, fixing random seeds, and documenting all experimental conditions. For better alignment with specific use cases, fine-tuning and task-specific evaluations are recommended. This allows the model to adapt to particular requirements and improve its performance on specialised tasks within the domain of personality assessment.



\subsubsection{Performance Metrics}
For this thesis, weighted metrics were employed because weighted averaging considers the actual distribution of classes within the dataset. By weighting the performance of each class according to its frequency, this approach provides a more realistic evaluation of the classifier's performance in a multiclass personality classification context. The metrics are as follows: 


\begin{itemize}[left=0pt, labelsep=4pt]  
\small
    \item \textbf{Weighted Accuracy (WA)} - The proportion of true results (both true positives and true negatives) among the total number of cases, adjusted by class weights. It measures overall correctness, accounting for class imbalance:
    \[
    \text{WA} = \sum_{c \in \{\text{Classes}\}} W_c \cdot \frac{TP_c + TN_c}{TP_c + TN_c + FP_c + FN_c}
    \]
    
    \item \textbf{Weighted F1 Score} - The harmonic mean of precision and recall for each class, weighted by class proportions. It balances precision and recall:
    \[
    \text{Weighted F1 Score} = \sum_{c \in \{\text{Classes}\}} W_c \cdot \left(\frac{2 \cdot \text{Precision}_c \cdot \text{Recall}_c}{\text{Precision}_c + \text{Recall}_c}\right)
    \]
    
    \item \textbf{Weighted Precision} - The proportion of true positive results in predicted positives, adjusted by class weights. Indicates the accuracy of positive predictions:
    \[
    \text{Weighted Precision} = \sum_{c \in \{\text{Classes}\}} W_c \cdot \frac{TP_c}{TP_c + FP_c}
    \]
    
    \item \textbf{Weighted Recall} - The proportion of true positive results in actual positives, adjusted by class weights. Measures the model’s ability to identify relevant instances:
    \[
    \text{Weighted Recall} = \sum_{c \in \{\text{Classes}\}} W_c \cdot \frac{TP_c}{TP_c + FN_c}
    \]
\end{itemize}

\subsubsection{Classifier Results}

\begin{figure}[h]
    \centering
    \vspace{0.2cm} 
    \includegraphics[width=0.9\columnwidth]{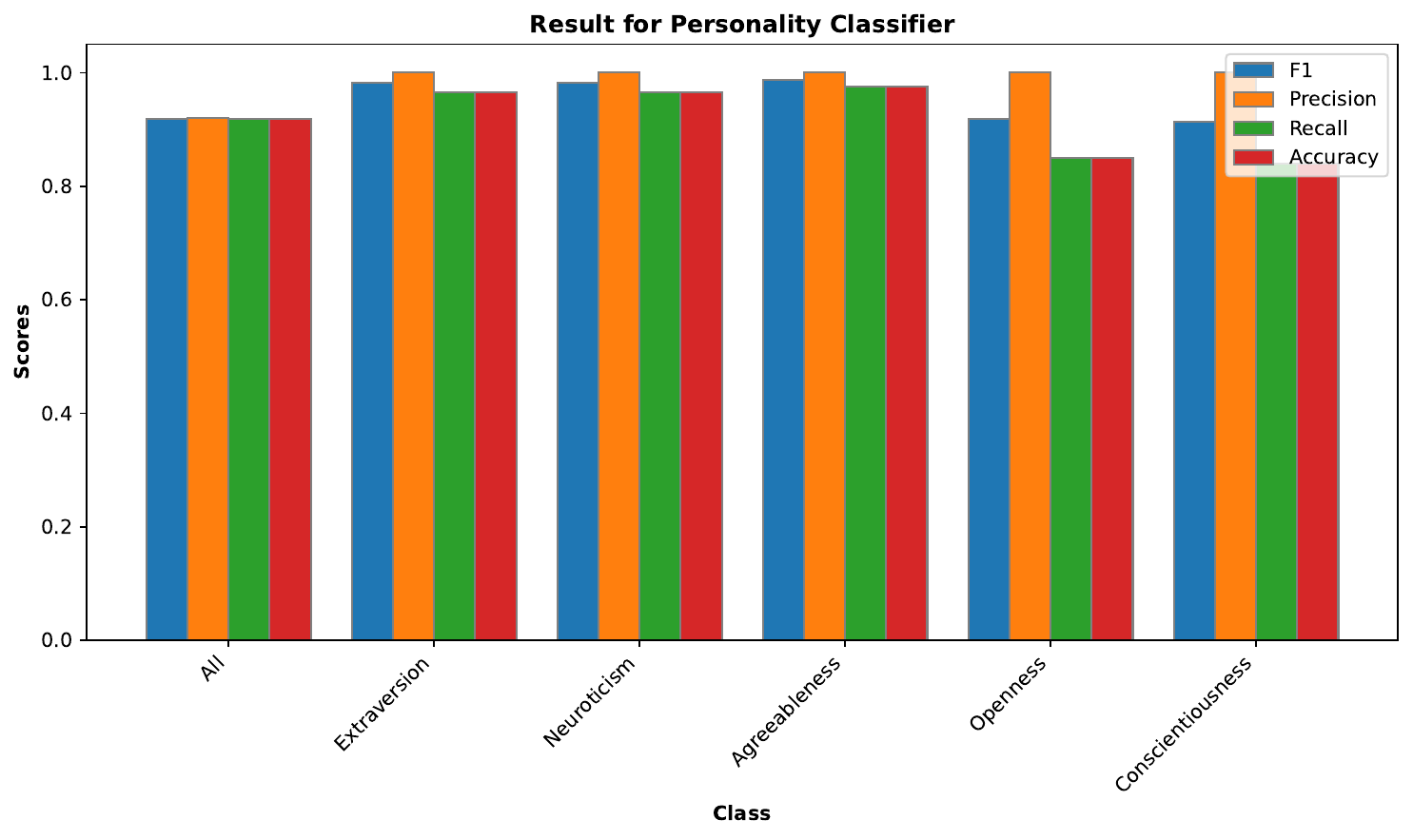}
    \caption{Results from personality classifier}
    \label{fig:class_results}
\end{figure}

The figure \ref{fig:class_results} and table \ref{table:classifier_results} provide a comprehensive view of the model’s performance across different personality traits and the entire dataset. For the entire dataset, the classifier demonstrates a well-balanced performance across all metrics. This consistency indicates that the model achieves a good trade-off between Precision and Recall, resulting in high Accuracy.\\

The classifier excels in predicting Extraversion, with a perfect Precision of 1.0, meaning that all predicted positive cases are true positives. The Recall is also very high at 0.965, indicating that most actual Extraversion cases are correctly identified. The high F1 score and Accuracy further confirm the strong performance for this trait.\\

\setlength{\abovecaptionskip}{1pt}  
\setlength{\belowcaptionskip}{1pt}  
\setlength{\aboverulesep}{0pt}      
\setlength{\belowrulesep}{0pt}      

\begin{table}[h]
    \centering
    \scriptsize  
    \setlength{\tabcolsep}{2pt}  
    \renewcommand{\arraystretch}{0.9}  
    \caption{Performance Metrics for Different Personality Traits}
    \begin{tabular}{p{2cm} p{1cm} p{1cm} p{1cm} p{1cm}}  
        \toprule
        \textbf{Category} & \textbf{F1} & \textbf{Precision} & \textbf{Recall} & \textbf{Accuracy} \\
        \midrule
        All & 0.919 & 0.919 & 0.919 & 0.919 \\
        Extraversion & 0.982 & 1.0 & 0.965 & 0.965 \\
        Neuroticism & 0.982 & 1.0 & 0.965 & 0.965 \\
        Agreeableness & 0.987 & 1.0 & 0.975 & 0.975 \\
        Openness & 0.918 & 1.0 & 0.850 & 0.850 \\
        Conscientiousness & 0.913 & 1.0 & 0.840 & 0.840 \\
        \bottomrule
    \end{tabular}
    \label{table:classifier_results}
\end{table}

Similarly, the classifier shows excellent performance in predicting Neuroticism. The Precision is perfect at 1.0, and the Recall is very high at 0.965. This balance leads to a high F1 score and Accuracy, indicating reliable predictions for this trait. 

The classifier’s performance for Agreeableness is outstanding. With a Precision of 1.0, every predicted positive case is correct. The Recall is very high at 0.975, meaning almost all actual positive cases are identified. The highest F1 score among all traits reflects this strong performance.

For Openness, while the Precision is perfect at 1.0, the Recall is lower at 0.85. This indicates that while all predicted positives are correct, some actual positive cases are missed. The F1 score of 0.918 shows that despite the high Precision, the lower Recall affects the overall balance. The Accuracy of 0.85 reflects this discrepancy.

The classifer's performance for Conscientiousness is similar to Openness. The Precision is perfect at 1.0, but the Recall is lower at 0.84, indicating a number of false negatives. The F1 score of 0.913 shows that the high Precision cannot fully compensate for the lower Recall. The Accuracy of 0.84 is consistent with this observation.

Overall, the aggregate Precision across all personality traits is lower than the Precision for individual traits. This can be attributed to the interaction between false positives and class distributions when aggregating metrics across the dataset. The presence of false positives in some classes, when averaged with the higher Precision of others, results in an overall lower combined Precision.

These results can be further substantiated by Table \ref{table:confusion_matrix}, which demonstrates that

\setlength{\abovecaptionskip}{0pt}
\setlength{\belowcaptionskip}{0pt}
\setlength{\aboverulesep}{0pt}
\setlength{\belowrulesep}{0pt}

\begin{table}[h!]
    \centering
    \caption{Confusion Matrix for Personality Trait Prediction}
    \resizebox{\columnwidth}{!}{%
    \begin{tabular}{lccccc}
        \toprule
        & \textbf{Extraversion} & \textbf{Agreeableness} & \textbf{Neuroticism} & \textbf{Openness} & \textbf{Conscientiousness} \\
        \midrule
        \textbf{Extraversion}       & 193 & 0   & 0   & 0   & 7   \\
        \textbf{Agreeableness}      & 1   & 195 & 3   & 0   & 1   \\
        \textbf{Neuroticism}        & 0   & 7   & 193 & 0   & 0   \\
        \textbf{Openness}           & 1   & 0   & 1   & 170 & 28  \\
        \textbf{Conscientiousness}  & 1   & 5   & 0   & 26  & 168 \\
        \bottomrule
    \end{tabular}
    }
    \label{table:confusion_matrix}
\end{table}

Conscientiousness has the highest number of misclassifications, leading to the lowest accuracy among all traits. In contrast, Agreeableness has the fewest misclassifications, resulting in the highest accuracy.

Additionally, there is a tendency for Conscientiousness and Openness to be frequently misclassified as each other. Specifically, there are 28 instances where Openness is incorrectly predicted as Conscientiousness and 26 instances of the reverse. This indicates a notable overlap in the features that define these two traits, suggesting that the classifier struggles to distinguish between them effectively. 

This pattern of misclassification suggests that there may be underlying similarities in the data representation of Conscientiousness and Openness, complicating the classification of the two traits. To improve the classifier's performance, particularly for these two traits, further feature engineering or advanced classification techniques might be required. Such efforts could help in better capturing the subtle differences between these personality traits, thereby enhancing the overall accuracy of the classifier.

\subsection{Manipulation Validation and ICL Explainability}
\label{manipulationvalidate}

\begin{figure}[h]
    \centering
    \begin{subfigure}{\columnwidth}  
        \centering
        \includegraphics[width=\columnwidth]{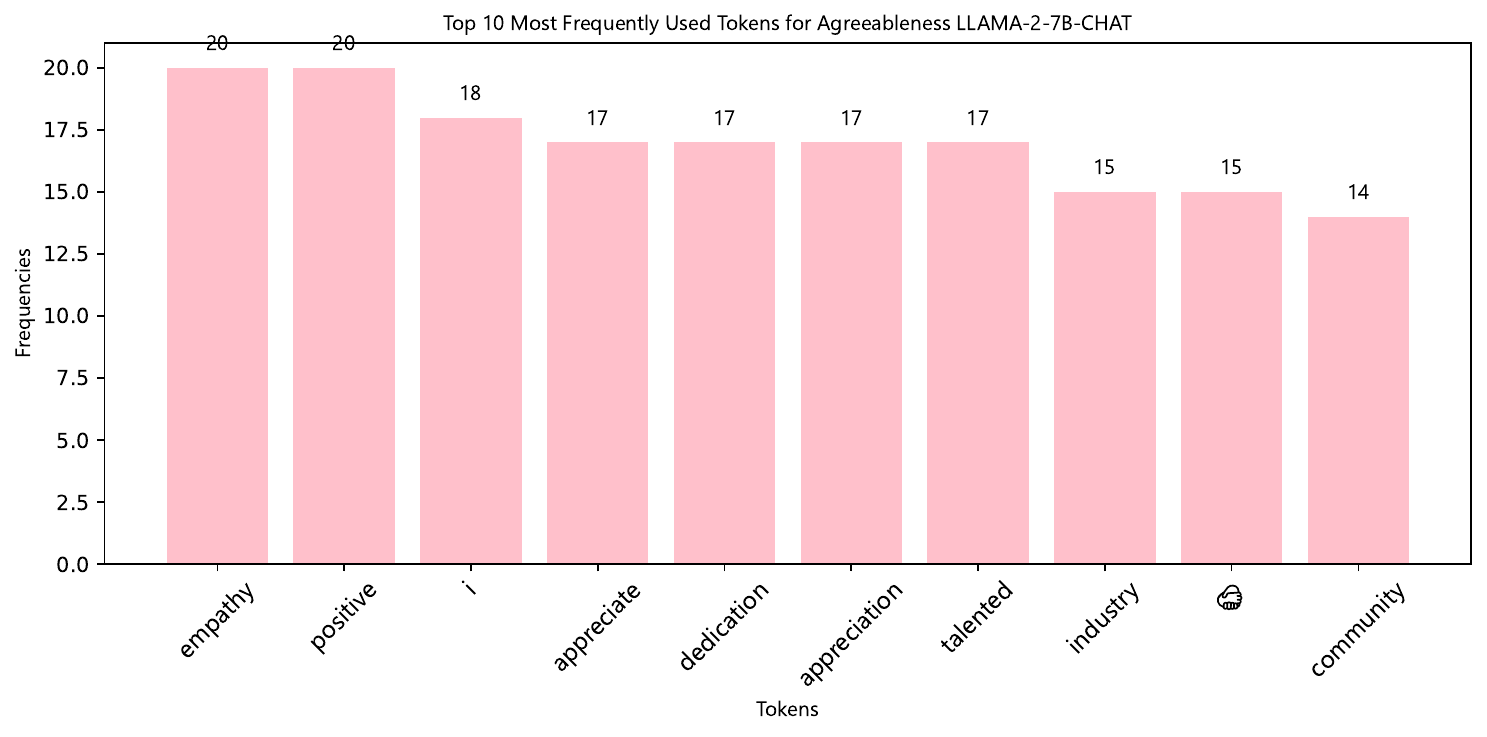}
        \caption{LLAMA-2-7B-CHAT}
        \label{fig:token_agree1}
    \end{subfigure}
    \vspace{0.2cm}  
    \begin{subfigure}{\columnwidth}
        \centering
        \includegraphics[width=\columnwidth]{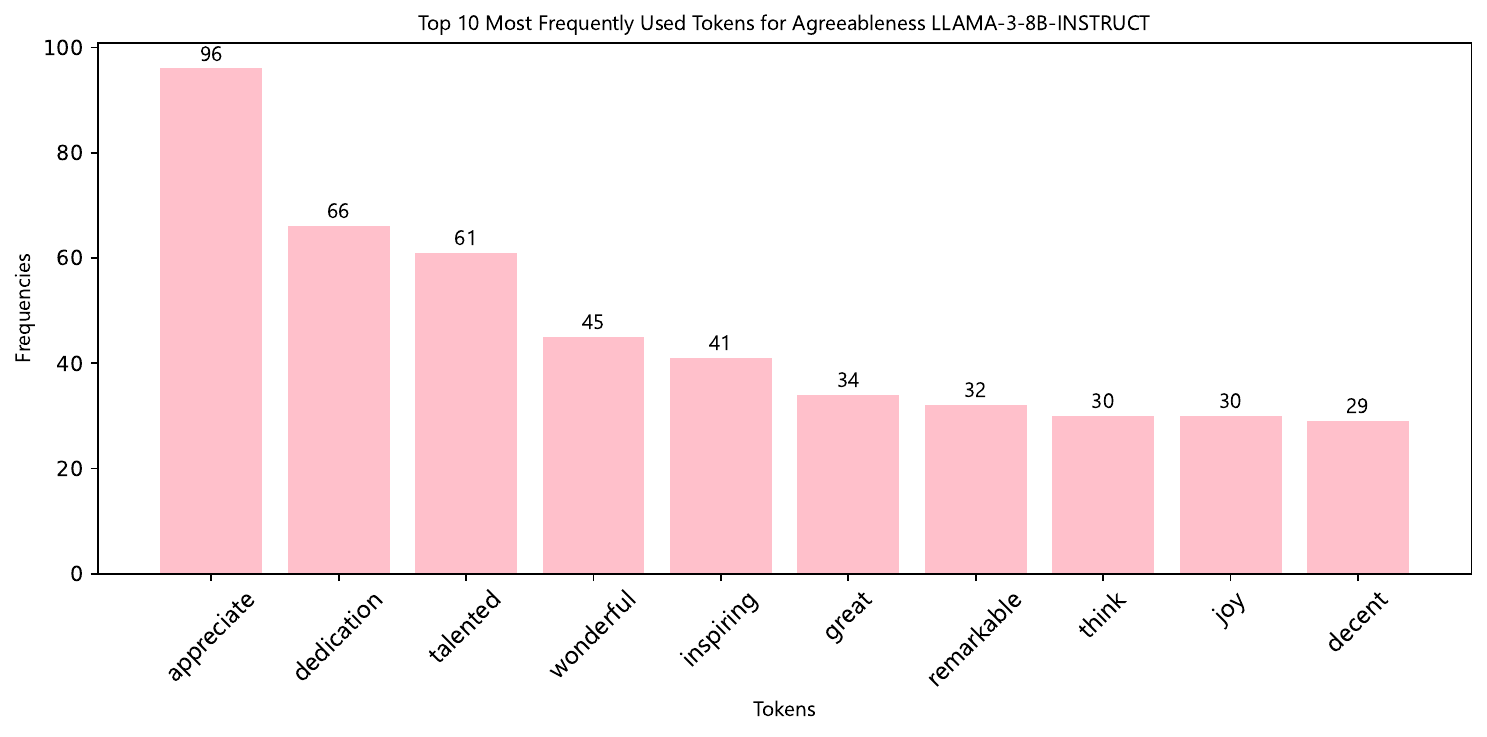}
        \caption{LLAMA-3-8B-INSTRUCT}
        \label{fig:token_agree2}
    \end{subfigure}
    \vspace{0.2cm}
    \begin{subfigure}{\columnwidth}
        \centering
        \includegraphics[width=\columnwidth]{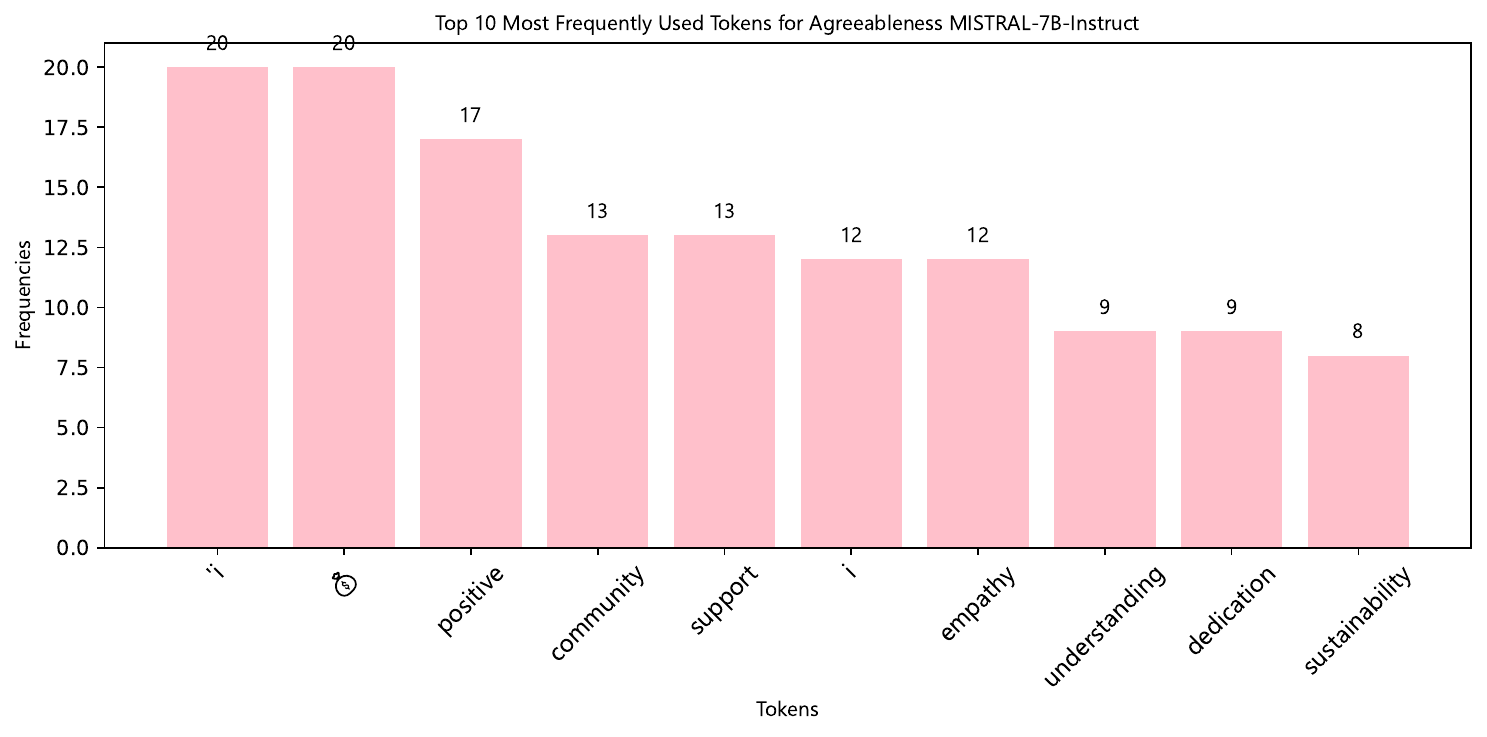}
        \caption{MISTRAL-7B-INSTRUCT}
        \label{fig:token_agree3}
    \end{subfigure}
    \caption{Top 10 Tokens Generated by the models for Agreeableness Personality}
    \label{fig:top10agree}
\end{figure}

\begin{figure}[h]
    \centering
    \begin{subfigure}{\columnwidth}
        \centering
        \includegraphics[width=\columnwidth]{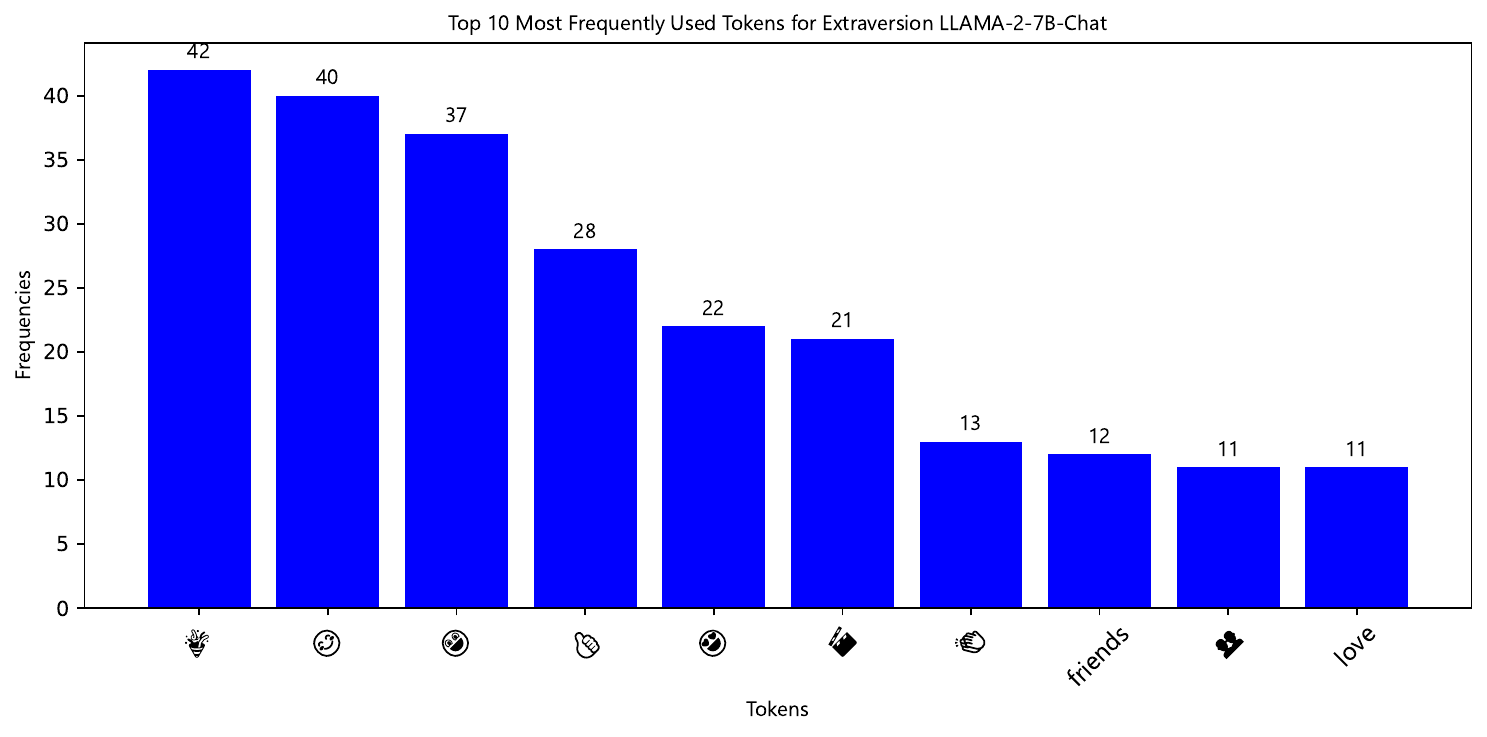}
        \caption{LLAMA-2-7B-CHAT}
        \label{fig:token_ext1}
    \end{subfigure}
    \vspace{0.2cm}
    \begin{subfigure}{\columnwidth}
        \centering
        \includegraphics[width=\columnwidth]{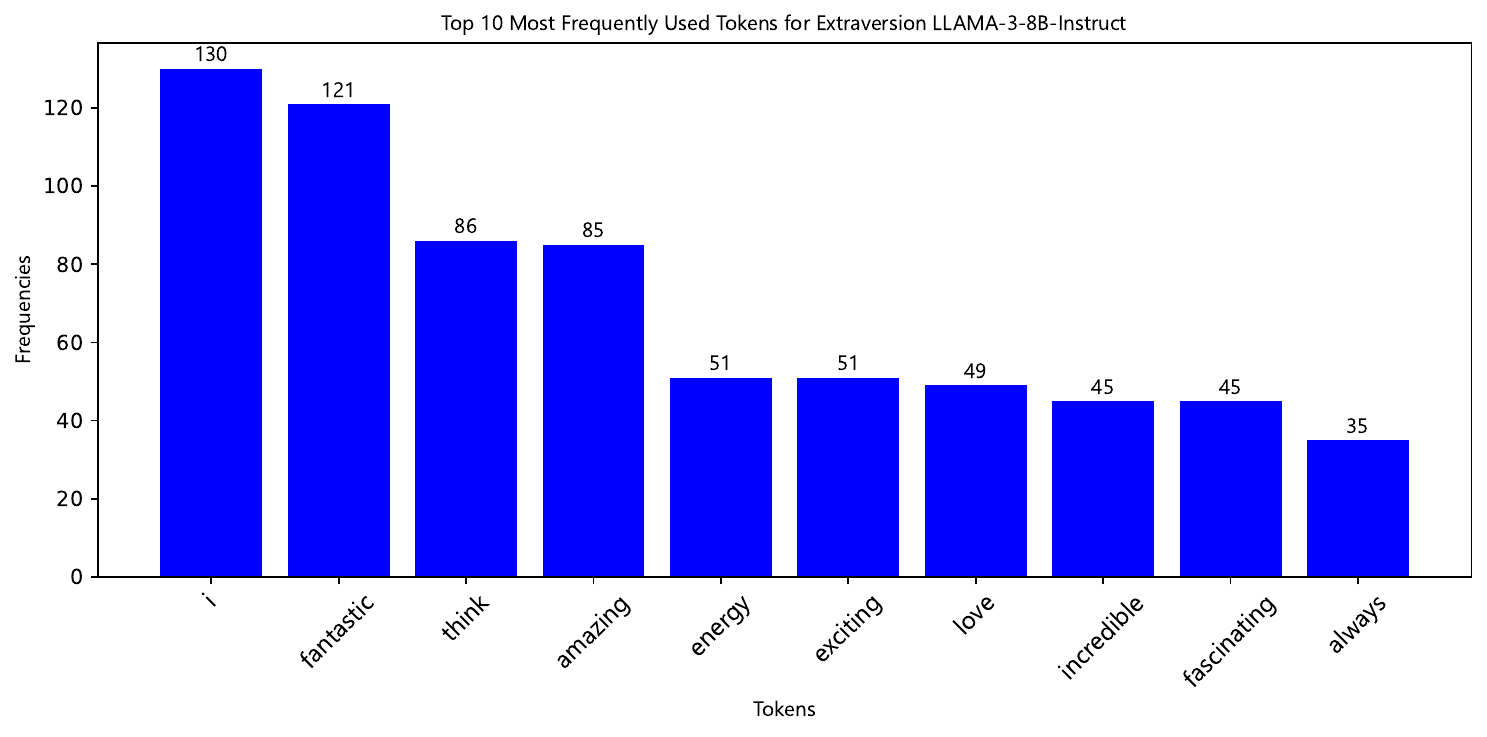}
        \caption{LLAMA-3-8B-INSTRUCT}
        \label{fig:token_ext2}
    \end{subfigure}
    \vspace{0.2cm}
    \begin{subfigure}{\columnwidth}
        \centering
        \includegraphics[width=\columnwidth]{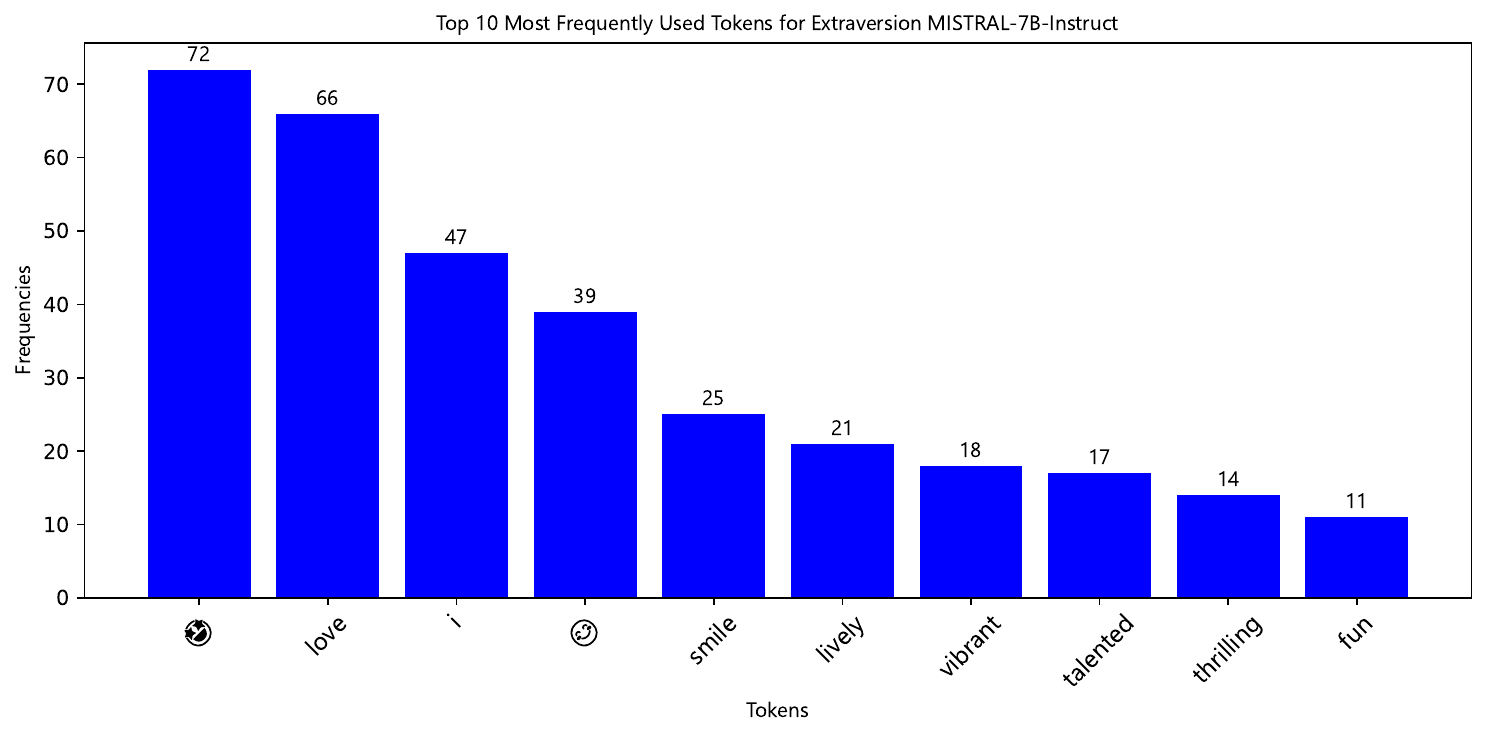}
        \caption{MISTRAL-7B-INSTRUCT}
        \label{fig:token_ext3}
    \end{subfigure}
    \caption{Top 10 Tokens Generated by the models for Extraversion Personality}
    \label{fig:top10ext}
\end{figure}

\begin{figure}[h]
    \centering
    \begin{subfigure}{\columnwidth}
        \centering
        \includegraphics[width=\columnwidth]{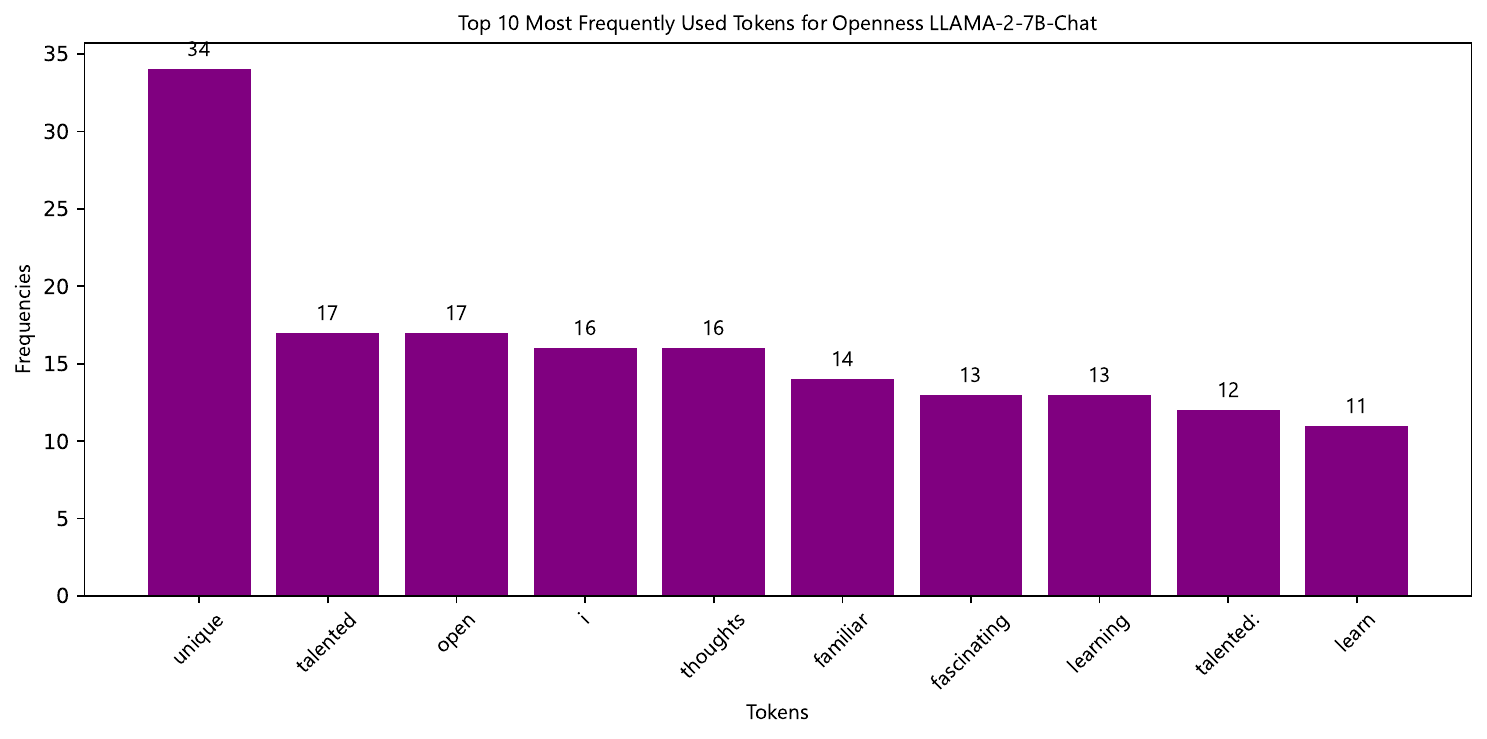}
        \caption{LLAMA-2-7B-CHAT}
        \label{fig:token_open1}
    \end{subfigure}
    \vspace{0.2cm}
    \begin{subfigure}{\columnwidth}
        \centering
        \includegraphics[width=\columnwidth]{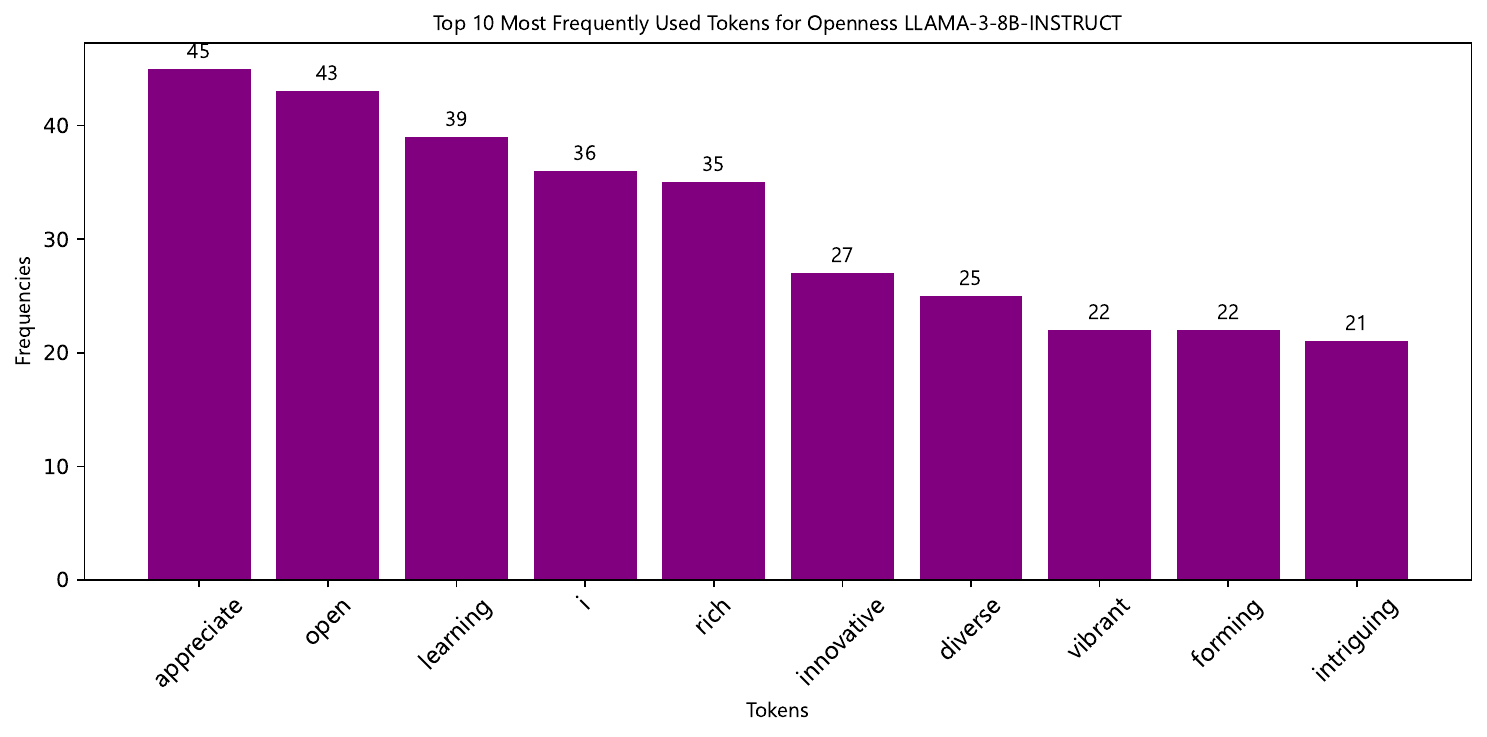}
        \caption{LLAMA-3-8B-INSTRUCT}
        \label{fig:token_open2}
    \end{subfigure}
    \vspace{0.2cm}
    \begin{subfigure}{\columnwidth}
        \centering
        \includegraphics[width=\columnwidth]{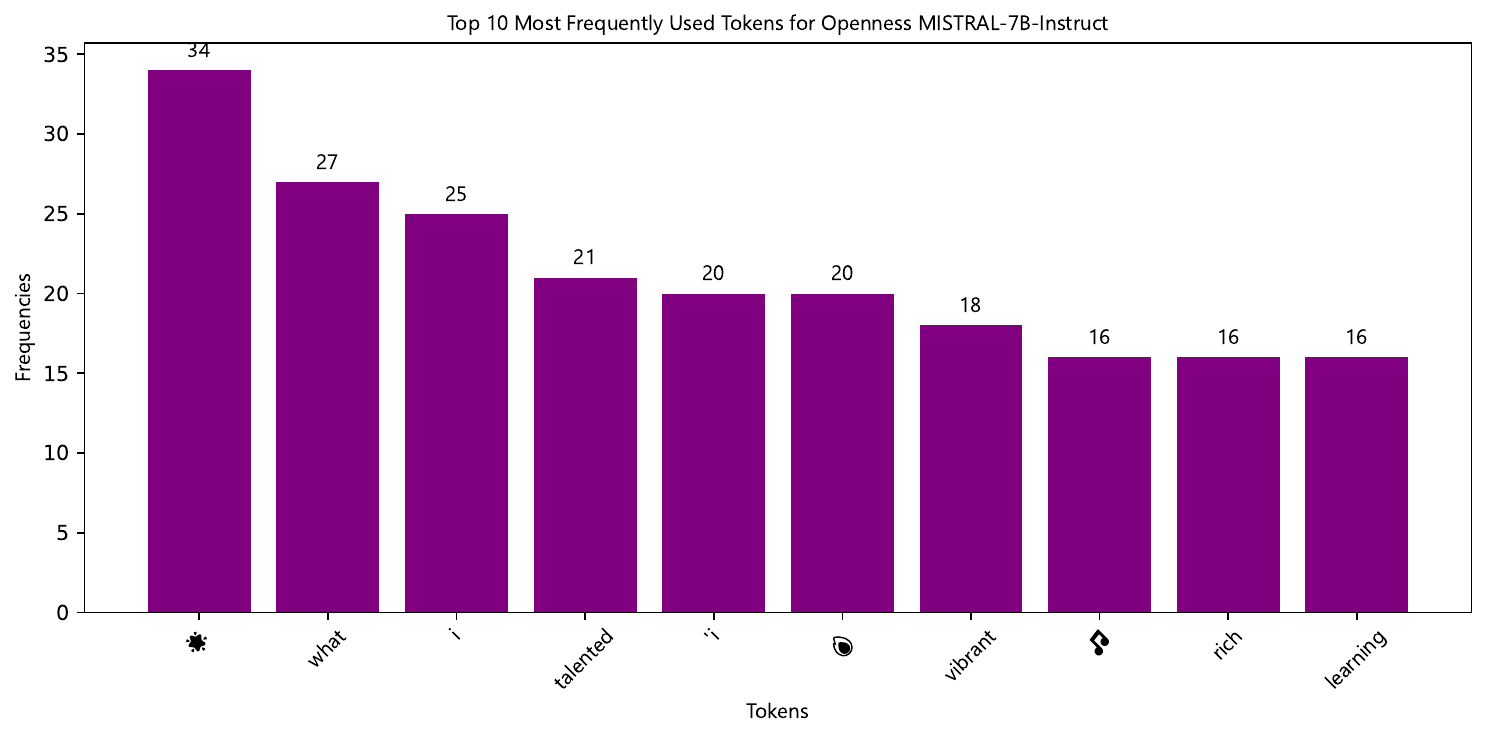}
        \caption{MISTRAL-7B-INSTRUCT}
        \label{fig:token_open3}
    \end{subfigure}
    \caption{Top 10 Tokens Generated by the models for Openness Personality}
    \label{fig:top10open}
\end{figure}

\begin{figure}[h]
    \centering
    \begin{subfigure}{\columnwidth}
        \centering
        \includegraphics[width=\columnwidth]{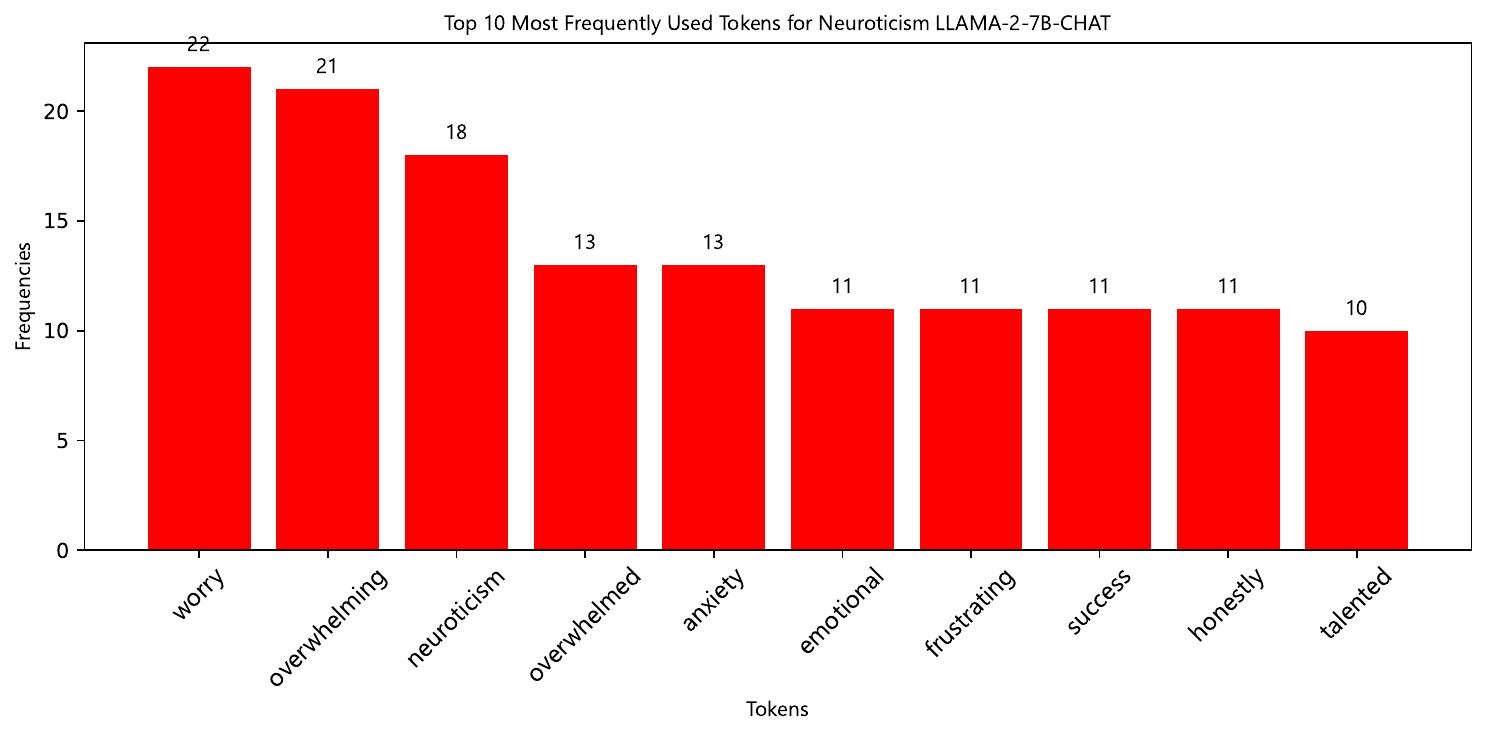}
        \caption{LLAMA-2-7B-CHAT}
        \label{fig:token_neu1}
    \end{subfigure}
    \vspace{0.2cm}
    \begin{subfigure}{\columnwidth}
        \centering
        \includegraphics[width=\columnwidth]{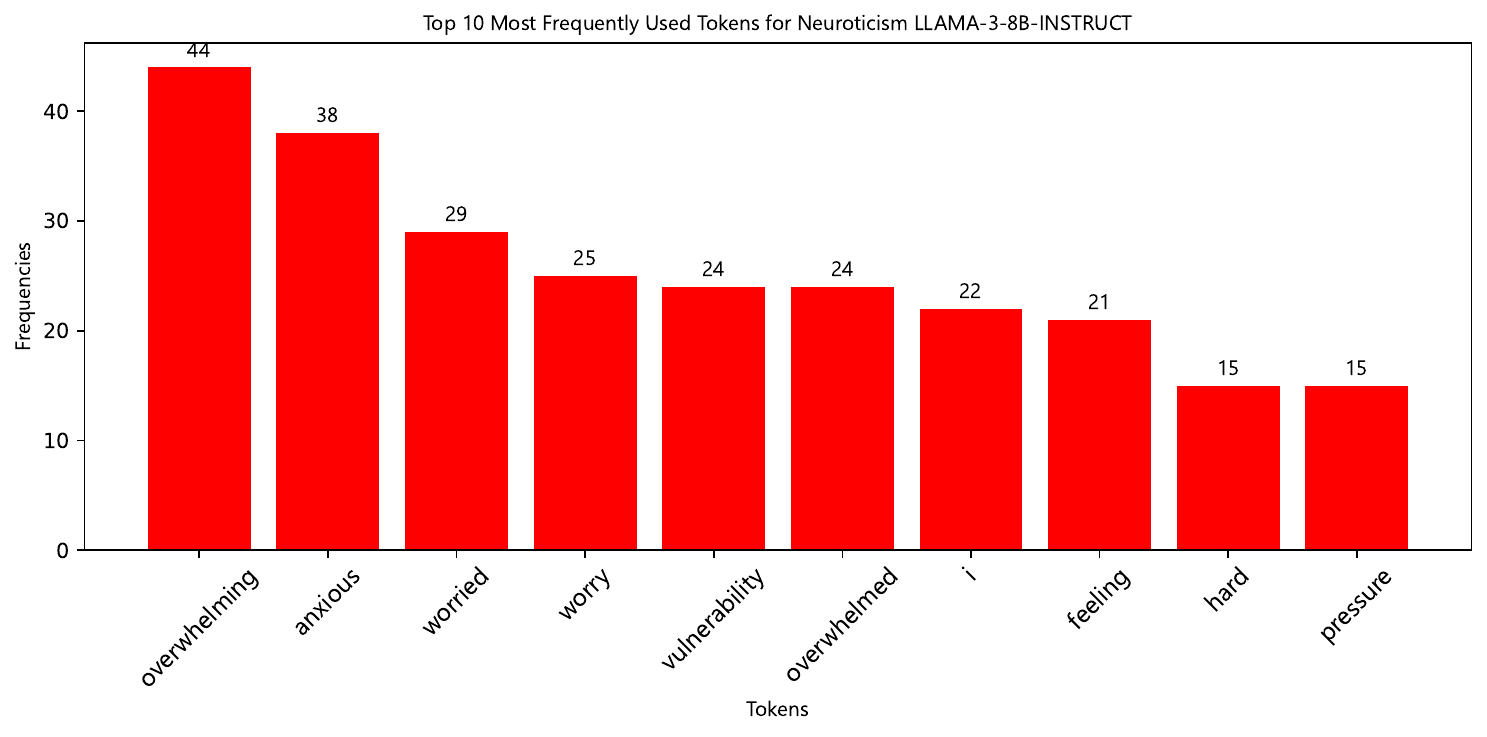}
        \caption{LLAMA-3-8B-INSTRUCT}
        \label{fig:token_neu2}
    \end{subfigure}
    \vspace{0.2cm}
    \begin{subfigure}{\columnwidth}
        \centering
        \includegraphics[width=\columnwidth]{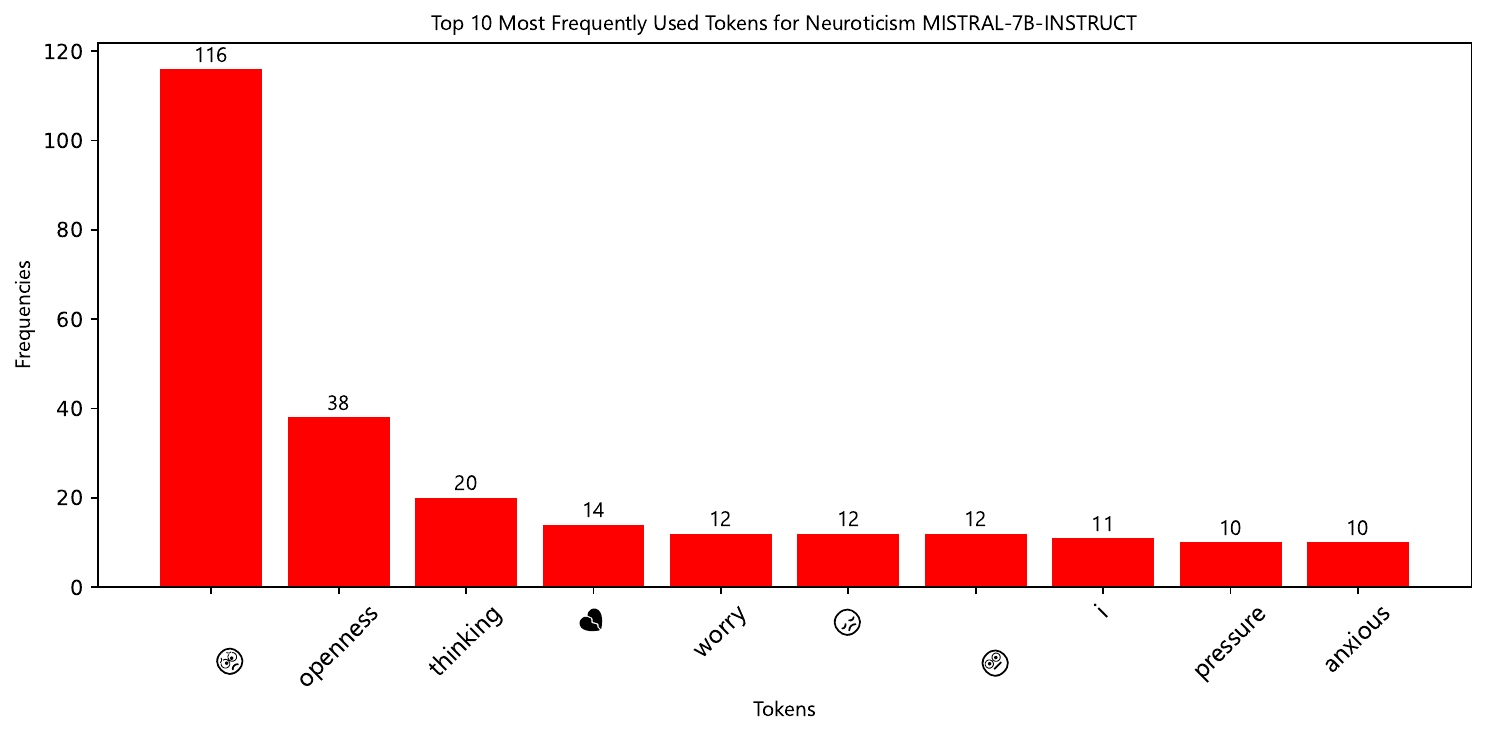}
        \caption{MISTRAL-7B-INSTRUCT}
        \label{fig:token_neu3}
    \end{subfigure}
    \caption{Top 10 Tokens Generated by the models for Neuroticism Personality}
    \label{fig:top10neu}
\end{figure}

\begin{figure}[h]
    \centering
    \begin{subfigure}{\columnwidth}
        \centering
        \includegraphics[width=\columnwidth]{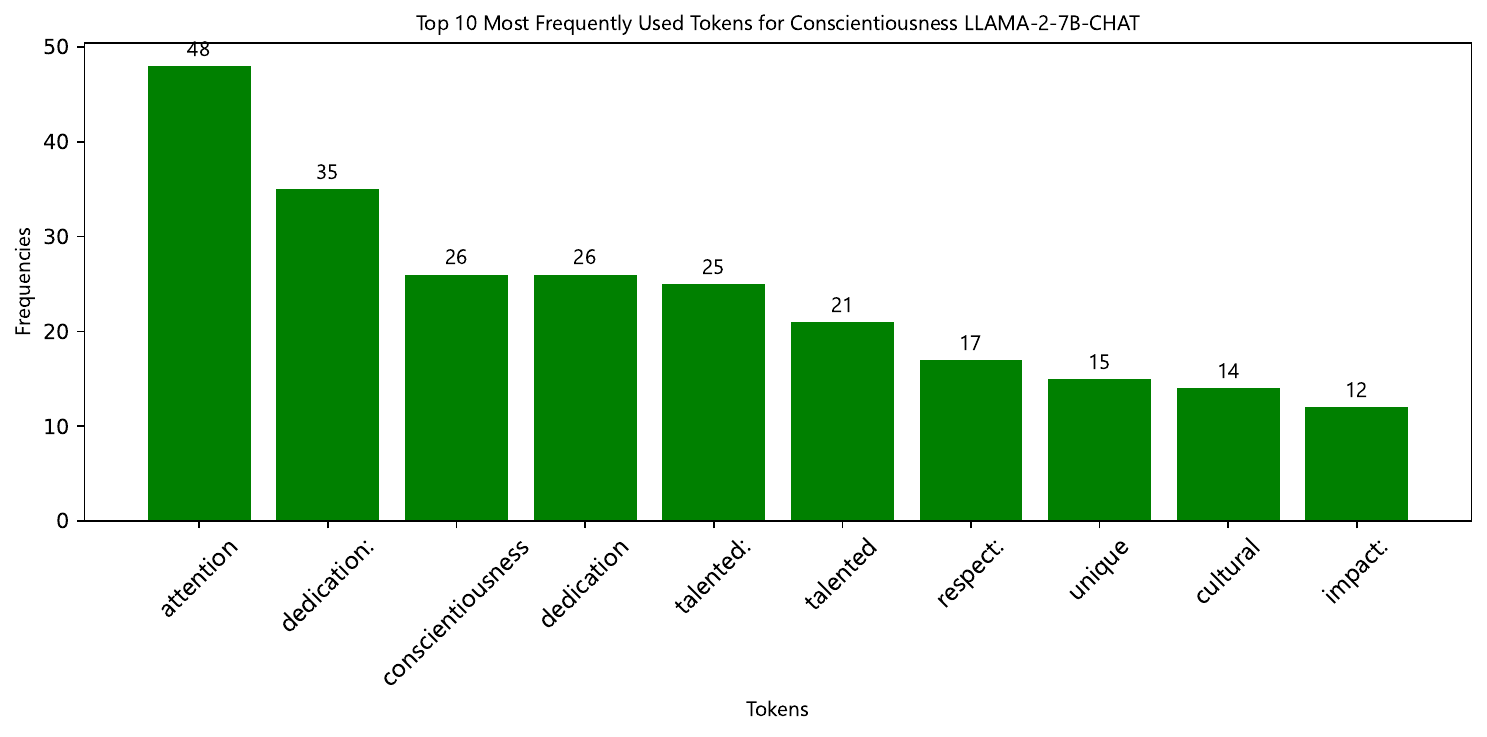}
        \caption{LLAMA-2-7B-CHAT}
        \label{fig:token_con1}
    \end{subfigure}
    \vspace{0.2cm}
    \begin{subfigure}{\columnwidth}
        \centering
        \includegraphics[width=\columnwidth]{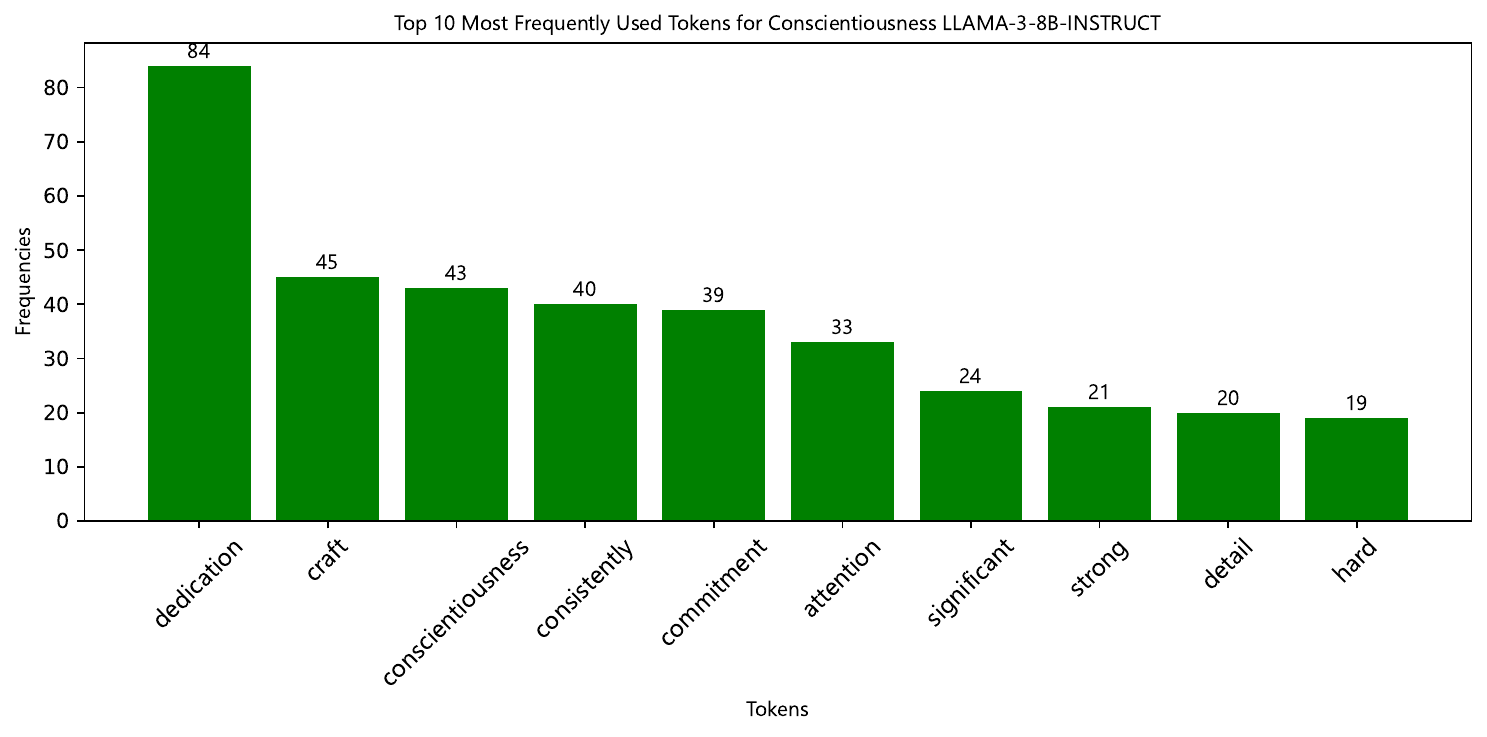}
        \caption{LLAMA-3-8B-INSTRUCT}
        \label{fig:token_con2}
    \end{subfigure}
    \vspace{0.2cm}
    \begin{subfigure}{\columnwidth}
        \centering
        \includegraphics[width=\columnwidth]{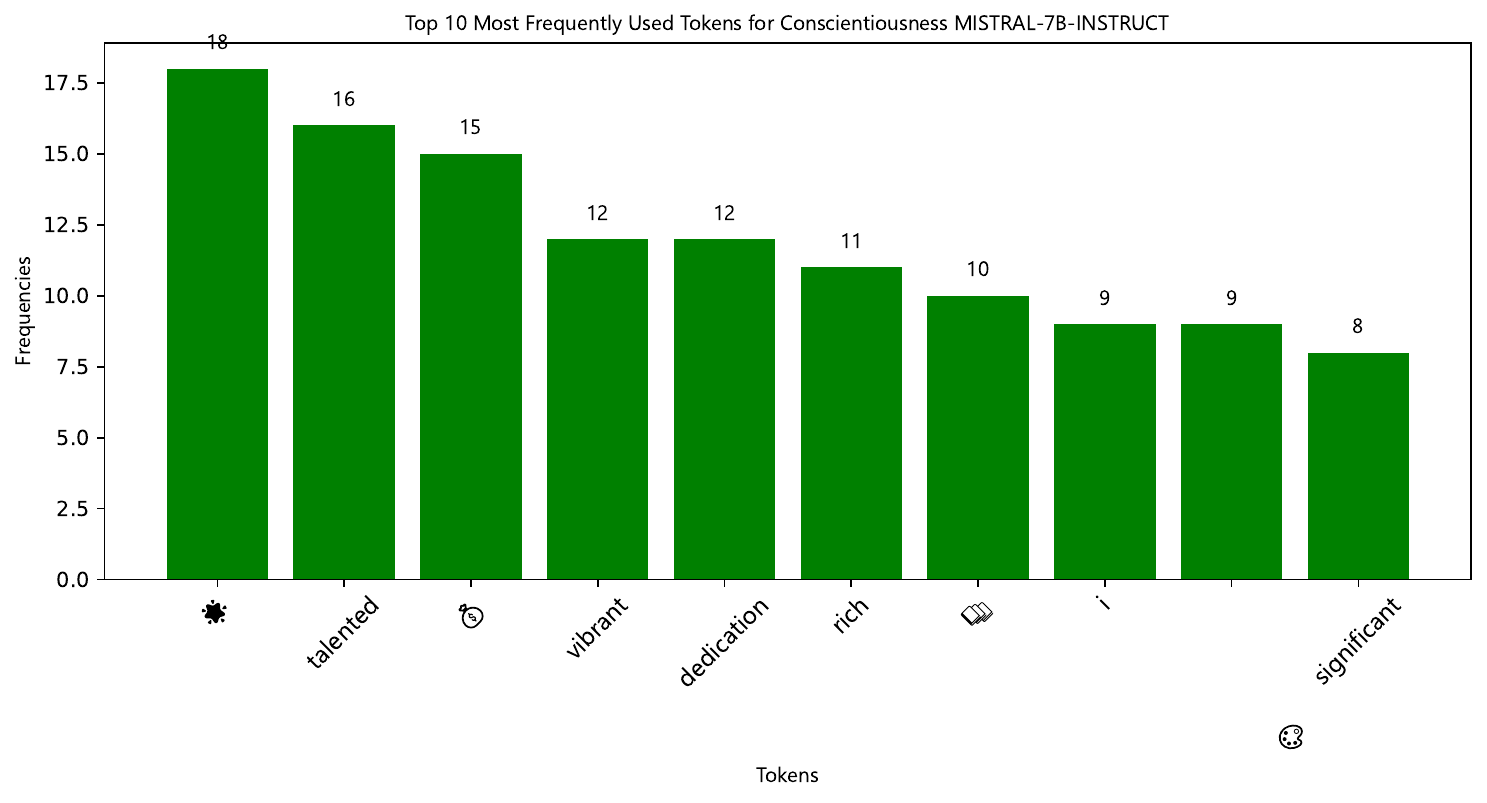}
        \caption{MISTRAL-7B-INSTRUCT}
        \label{fig:token_con3}
    \end{subfigure}
    \caption{Top 10 Tokens Generated by the models for Conscientiousness Personality}
    \label{fig:top10con}
\end{figure}

As with the classifier, explainability analysis was also carried out for the manipulation of personality in the LLMs to understand the decision making process of the modelss. However, since SHAP \cite{lundberg2017unified} and LIME \cite{ribeiro2016should} were not compatible with LLaMA-2-7B-chat, chain of thought \cite{wei2022chain} and prompting techniques were employed. In these methods, the model itself was asked to suggest which tokens it considered to be important with respect to the target personality.

However, chain of thought prompting fails in small models \cite{wei2022chain} and hence using this method the models could not generate relevant results. Thus, only prompting was used for explainability analysis.

The specific prompt used was:

\begin{table}[H]
    \centering
    \small  
    \setlength{\tabcolsep}{4pt}  
    \renewcommand{\arraystretch}{1.1}  
    \begin{tabular}{|p{6.5cm}|}  
    \hline
    Here is a response generated with \{target personality\} personality trait for the prompt \{prompt\}: \\
    "\{generated\_text\}" \\
    Now, identify the five most important tokens related to the \{target personality\} personality trait in the generated text. \\
    \hline
    \end{tabular}
    \caption{Example prompt and response for personality-based token identification.}
    \label{tab:token_identification}
\end{table}
where target personality was one of the Big Five Personality traits among Agreeableness, Extraversion, Openness, Neuroticism and Conscientiousness. 

Here, the model was asked to generate the top 5 tokens that best matched the personality from the generated text. Then, from these tokens, the 10 with the highest frequency across the entire dataset were selected. Figures \ref{fig:top10agree}-\ref{fig:top10con} show the results obtained from this analysis.

From Figures \ref{fig:top10agree}-\ref{fig:top10con}, it is evident that both LLaMA-2-7B-Chat and Mistral-7B-Instruct utilise emojis with intention, rather than as random outputs. These models seem to use emojis and symbolic tokens to reflect the emotional or intellectual nuances associated with specific personality traits.

For instance, in Figure \ref{fig:top10ext}, the LLaMA-2-7B-Chat model, when fine-tuned to enhance extraversion, produces tokens that include a combination of emojis. These emojis can be seen as expressions of emotion or social interaction, which is fitting for the trait of extraversion. Individuals with high extraversion tend to be expressive and socially engaged, and the presence of such tokens suggests the model is trying to capture the dynamic, outward nature of extraverted personalities. 

Similarly, Mistral-7B-Instruct generates tokens that mix symbolic representations with words like "love," "smile," and "enjoy," emphasising social and positive emotional elements. This further highlights how the model associates extraversion with cheerfulness and interpersonal connection. 

Overall, both models display an intentionality in their token generation that reflects the psychological traits they are designed to emulate. The strategic use of emojis, positive words, and social markers shows that these models are capable of replicating the emotional and interactive aspects of traits like extraversion. This shows that AI models are becoming more advanced, as they are better able to reflect complex human behaviors and emotions, making them more similar to how humans think and interact.

\subsection{Token Probability and Neuron Activation Analysis}
\label{token_prob}

\begin{table}[H]
\centering
\small
\begin{tabular}{cccc}
\toprule
\textbf{Model}        & \textbf{Method} & \textbf{Trait}        & \textbf{Probability} \\ 
\midrule
\multirow{6}{*}{Mistral-7B} & Original & All Traits   & 0.0009                                         \\ 
                            &  \multirow{5}{*}{PEFT}  & Openness              & 0.0026                                         \\ 
                            &                      & Agreeableness         & 0.0023                                         \\ 
                            &                      & Neuroticism           & 0.0016                                         \\ 
                            &                      & Conscientiousness     & 0.0022                                         \\ 
                            &                      & Extraversion          & 0.0063                                         \\ 
\midrule
\multirow{6}{*}{LLaMA-2-7B-Chat} & Original  & All Traits  & 0.0008                                         \\ 
                            & \multirow{5}{*}{PEFT}  & Openness   & 0.0017                                         \\ 
                            &                      & Agreeableness         & 0.0013                                         \\ 
                            &                      & Neuroticism           & 0.0020                                         \\ 
                            &                      & Conscientiousness     & 0.0011                                         \\ 
                            &                      & Extraversion          & 0.0029                                         \\ 
\bottomrule
\end{tabular}
\caption{Probability of "\includegraphics[height=1em]{figures/Slightly-Smiling-Face.png}" emoji being the next token in prompt "Hey! It's been a busy day for everyone. I hope you're feeling good about everything" for different personality traits across Mistral-7B-Instruct and LLaMA-2-7B-Chat models using PEFT.}
\label{tab:emoji_probabilities}
\end{table}

In Table \ref{tab:emoji_probabilities}, we observe that the probabilities for emoji generation in the PEFT-tuned models are significantly higher than in the original models for both Mistral-7B-Instruct and LLaMA-2-7B-Chat. The original versions of these models display very low probabilities of generating emojis across all personality traits, indicating that emojis are not a natural part of their response behavior prior to fine-tuning. Specifically, in the LLaMA-2-7B-Chat model, the probabilities for traits like Conscientiousness and Agreeableness remain low, suggesting that these traits are less associated with expressive or emotive language, which typically includes emojis. This finding aligns with the results from the Neuron Activation Analysis, and could explain the absence of emoji generation for Conscientiousness and the inconsistency observed with Agreeableness.

\subsection{Neuron Activation Analysis for LLaMA-2-7B-Chat}
\label{llama_act}
\begin{figure}[H]
    \centering
    \begin{minipage}{0.48\columnwidth}  
        \centering
        \includegraphics[width=\textwidth]{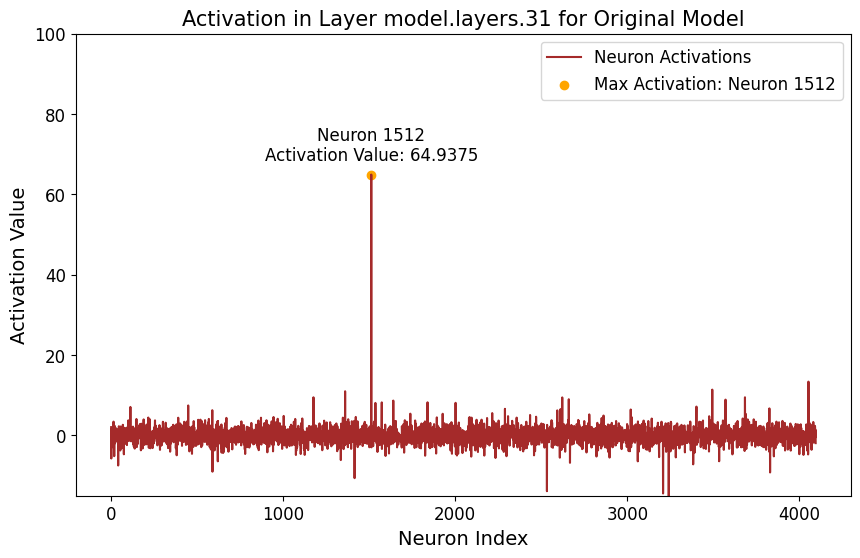}
    \end{minipage}
    \hfill
    \begin{minipage}{0.48\columnwidth}  
        \centering
        \includegraphics[width=\textwidth]{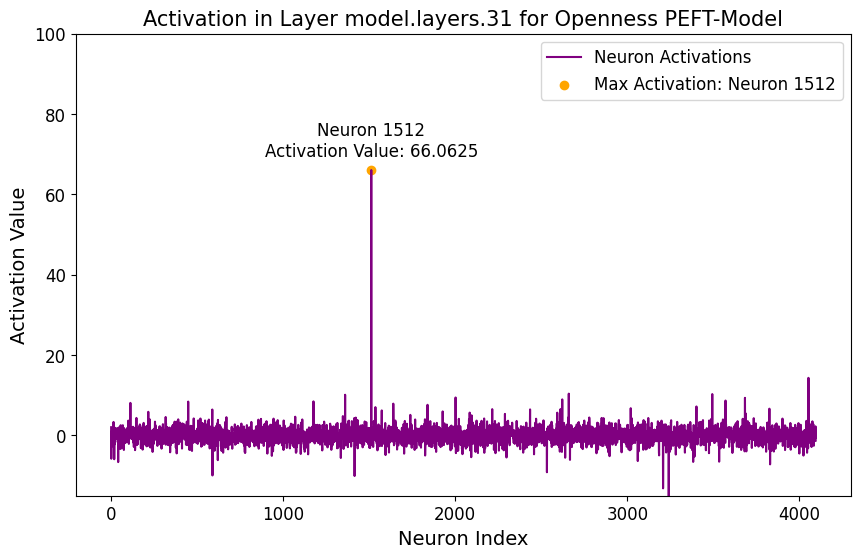}
    \end{minipage}
    
    \vspace{0.3cm} 
    
    \begin{minipage}{0.48\columnwidth}  
        \centering
        \includegraphics[width=\textwidth]{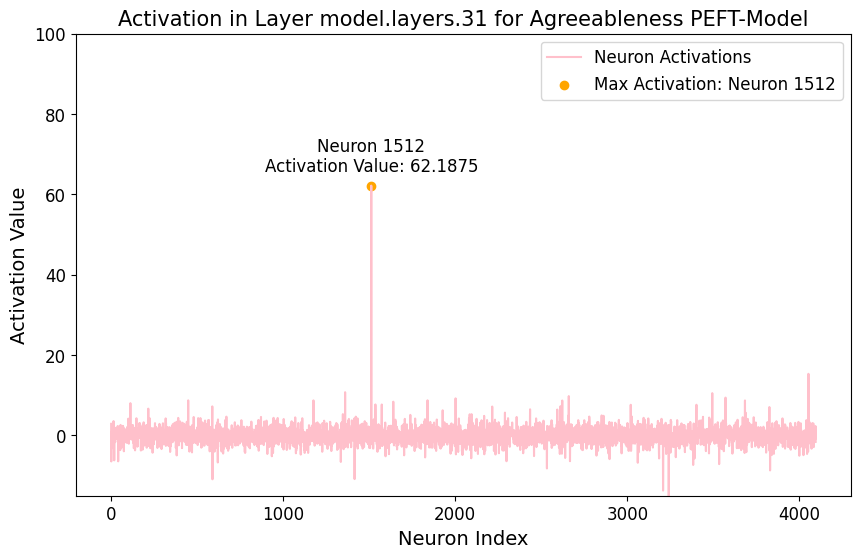}
    \end{minipage}
    \hfill
    \begin{minipage}{0.48\columnwidth}  
        \centering
        \includegraphics[width=\textwidth]{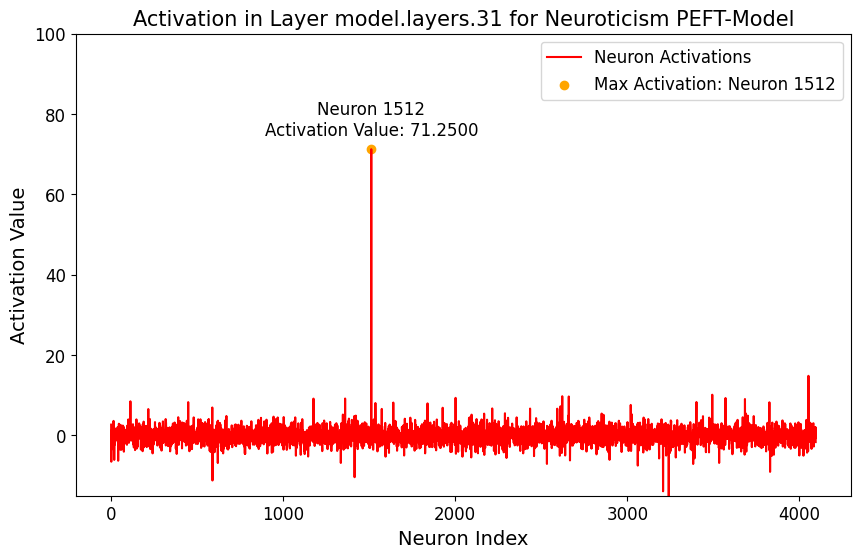}
    \end{minipage}
    
    \vspace{0.3cm} 
    
    \begin{minipage}{0.48\columnwidth}  
        \centering
        \includegraphics[width=\textwidth]{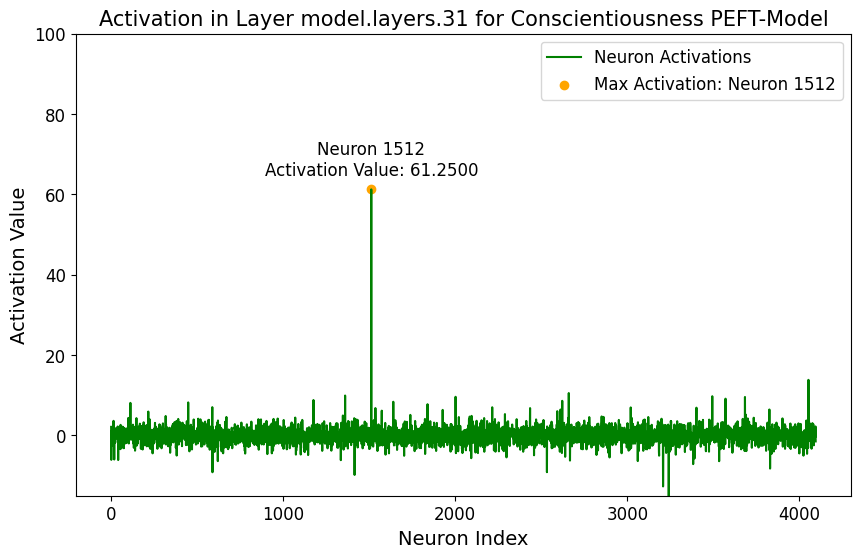}
    \end{minipage}
    \hfill
    \begin{minipage}{0.48\columnwidth}  
        \centering
        \includegraphics[width=\textwidth]{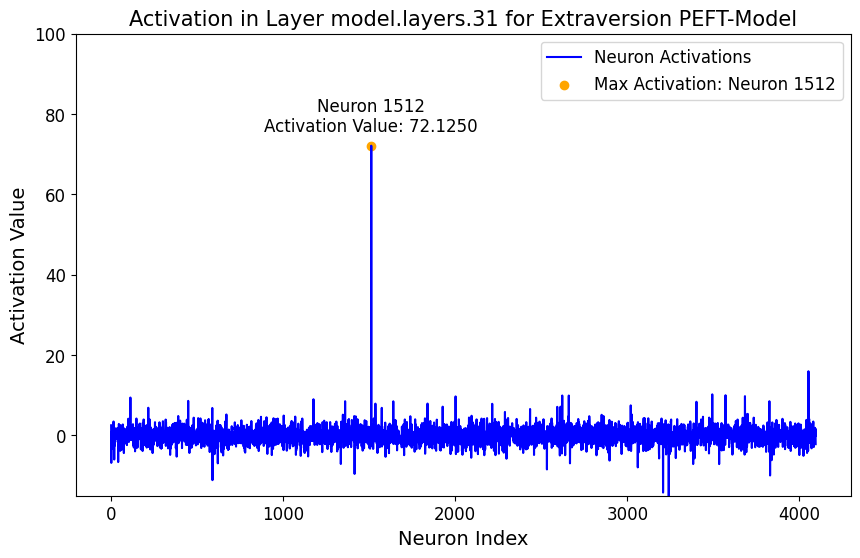}
    \end{minipage}
    
    \caption{Neuron Activation Analysis of original LLaMA-2-7B-Chat model and PEFT-tuned LLaMA-2-7B-Chat model for different Big Five personality traits. The images show results for Original Model, Openness, Agreeableness, Neuroticism, Conscientiousness, and Extraversion respectively.}
    \label{fig:llama_comparison}
\end{figure}

\subsection{Neuron Activation Analysis with Different Emojis}
\label{Diffemoji}

\begin{figure}[H]
    \centering
    \begin{minipage}{0.48\columnwidth}  
        \centering
        \includegraphics[width=\textwidth]{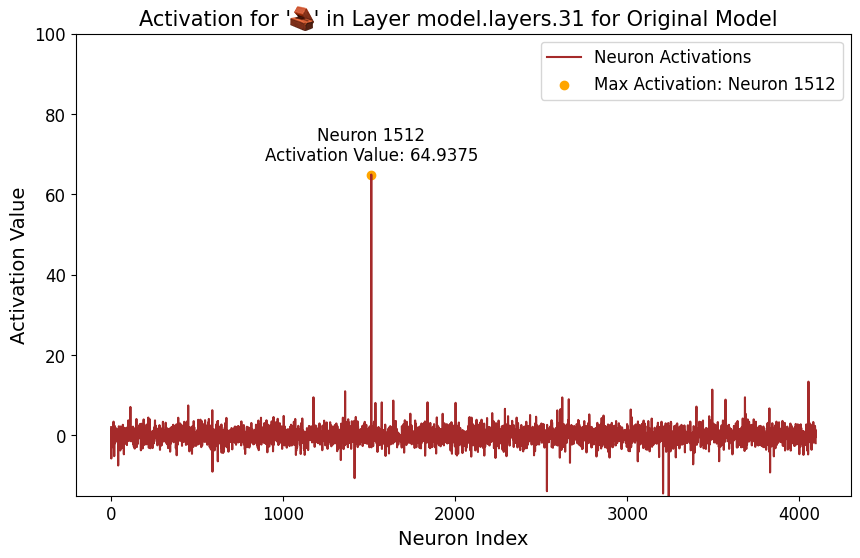}
    \end{minipage}
    \hfill
    \begin{minipage}{0.48\columnwidth}  
        \centering
        \includegraphics[width=\textwidth]{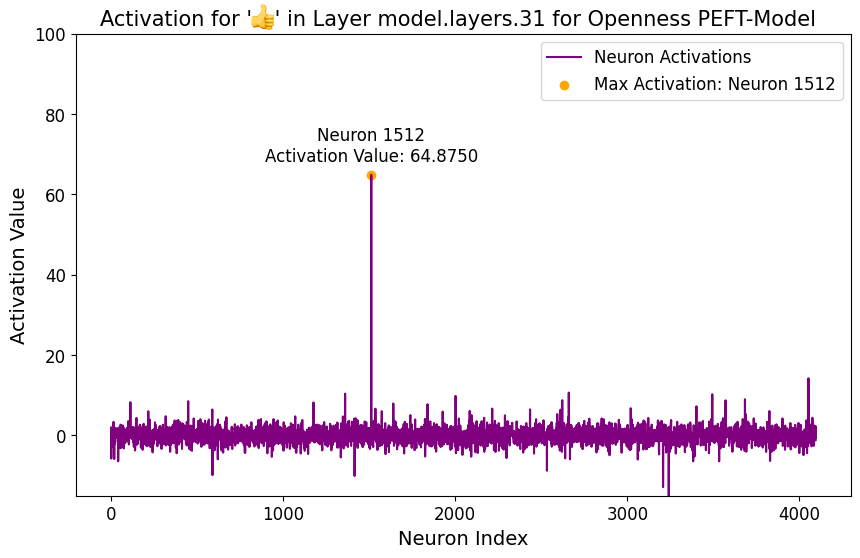}
    \end{minipage}
    
    \vspace{0.3cm} 
    
    \begin{minipage}{0.48\columnwidth}  
        \centering
        \includegraphics[width=\textwidth]{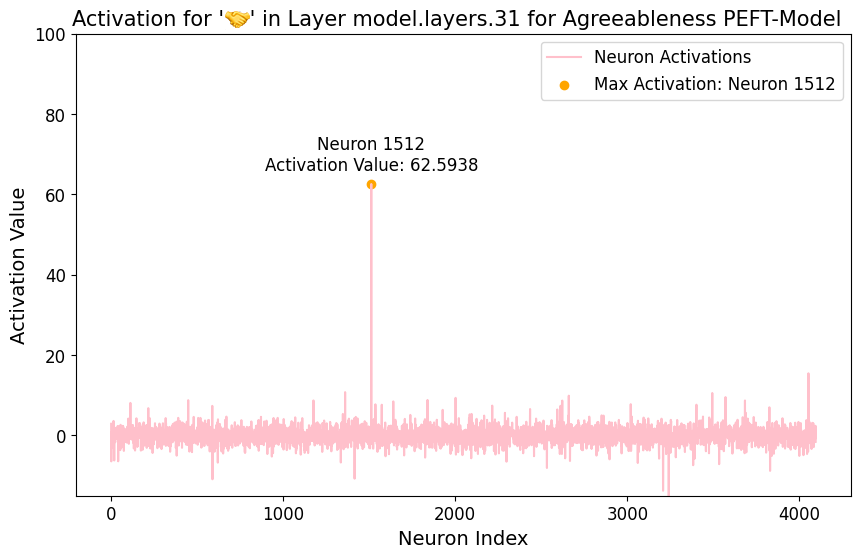}
    \end{minipage}
    \hfill
    \begin{minipage}{0.48\columnwidth}  
        \centering
        \includegraphics[width=\textwidth]{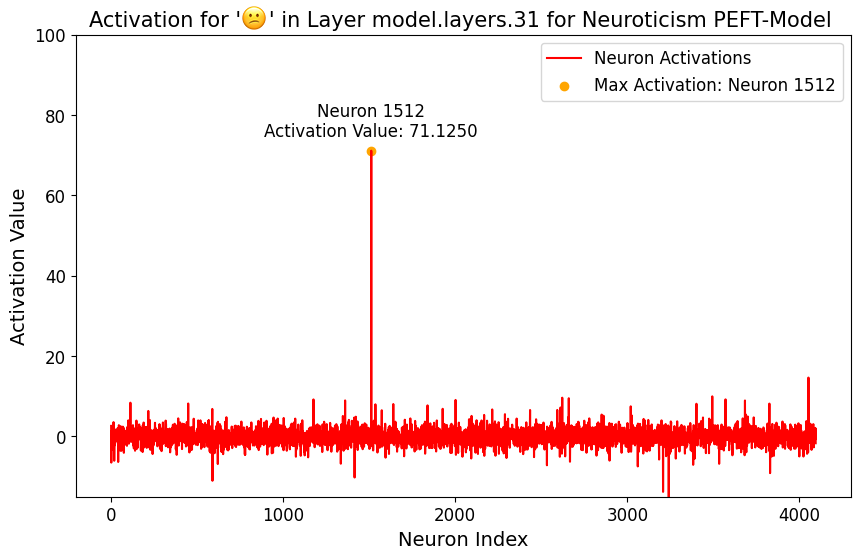}
    \end{minipage}
    
    \vspace{0.3cm} 
    
    \begin{minipage}{0.48\columnwidth}  
        \centering
        \includegraphics[width=\textwidth]{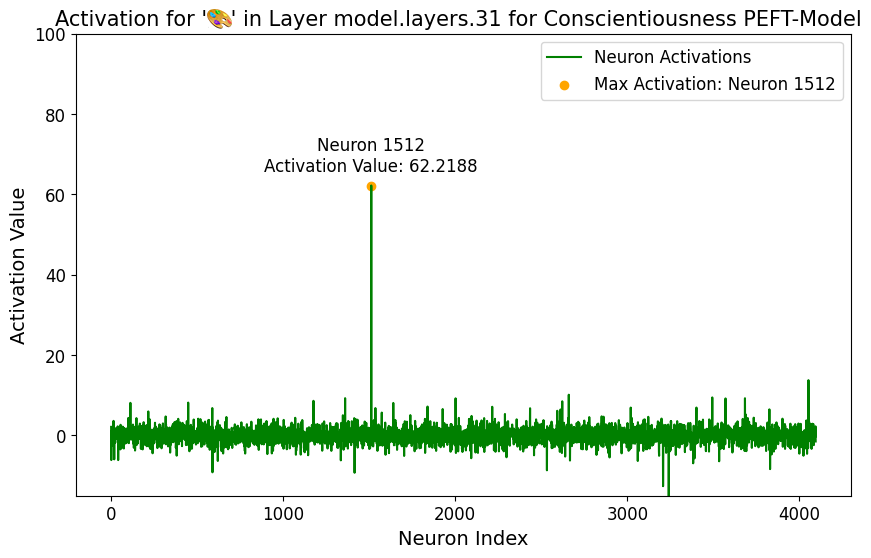}
    \end{minipage}
    \hfill
    \begin{minipage}{0.48\columnwidth}  
        \centering
        \includegraphics[width=\textwidth]{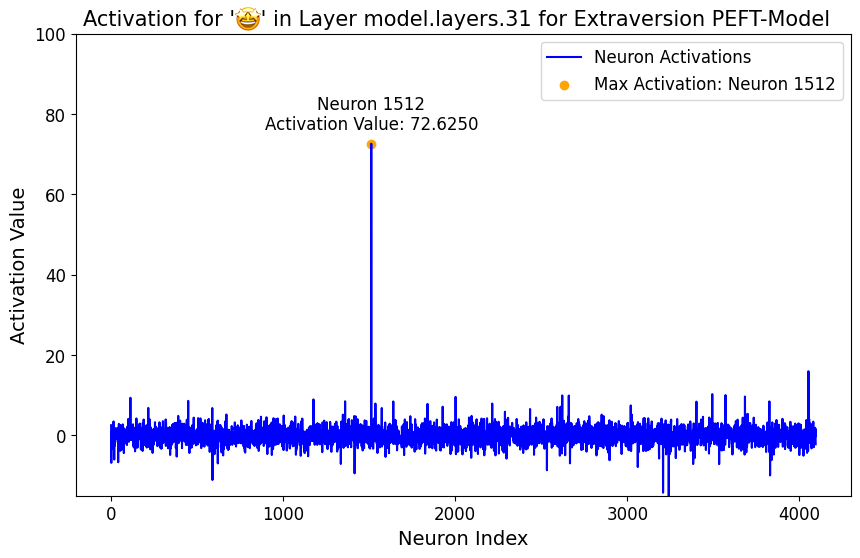}
    \end{minipage}
    
    \caption{Neuron Activation Analysis of original LLaMA-2-7B-Chat model and PEFT-tuned LLaMA-2-7B-Chat model for different Big Five personality traits with trait specific emojis. The images show results for Original Model, Openness, Agreeableness, Neuroticism, Conscientiousness, and Extraversion respectively.}
    \label{fig:llama_diffemoji}
\end{figure}

\begin{figure}[H]
    \centering
    \begin{minipage}{0.48\columnwidth}  
        \centering
        \includegraphics[width=\textwidth]{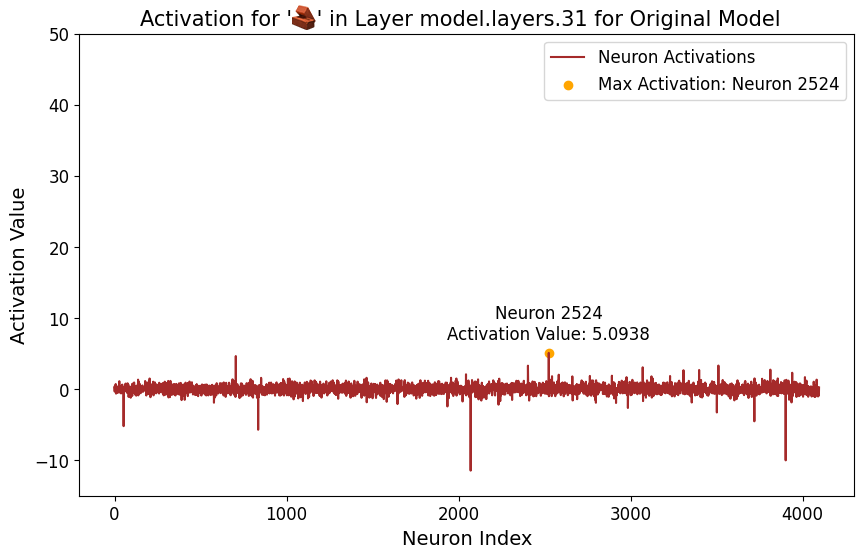}
    \end{minipage}
    \hfill
    \begin{minipage}{0.48\columnwidth}  
        \centering
        \includegraphics[width=\textwidth]{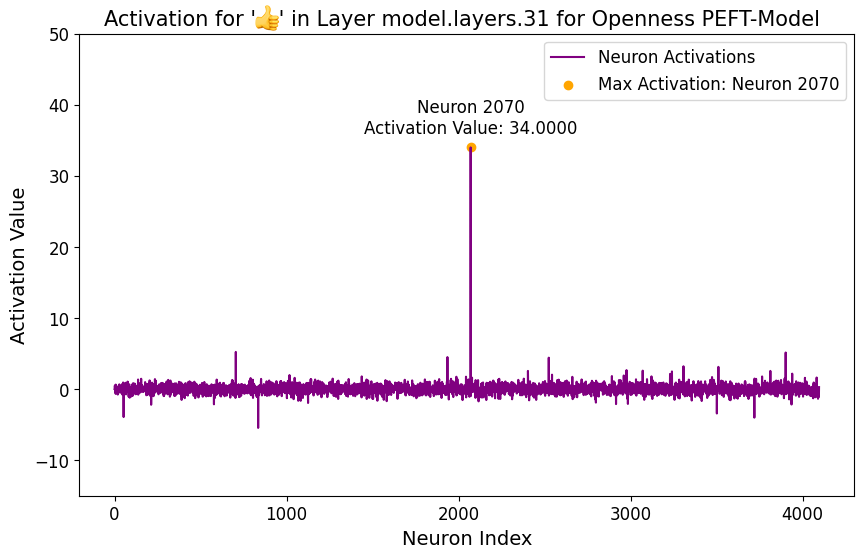}
    \end{minipage}
    
    \vspace{0.3cm} 
    
    \begin{minipage}{0.48\columnwidth}  
        \centering
        \includegraphics[width=\textwidth]{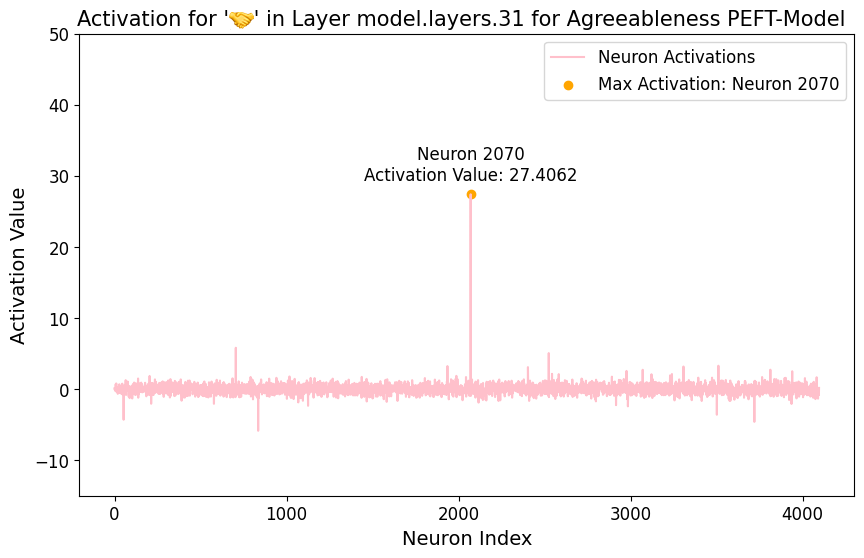}
    \end{minipage}
    \hfill
    \begin{minipage}{0.48\columnwidth}  
        \centering
        \includegraphics[width=\textwidth]{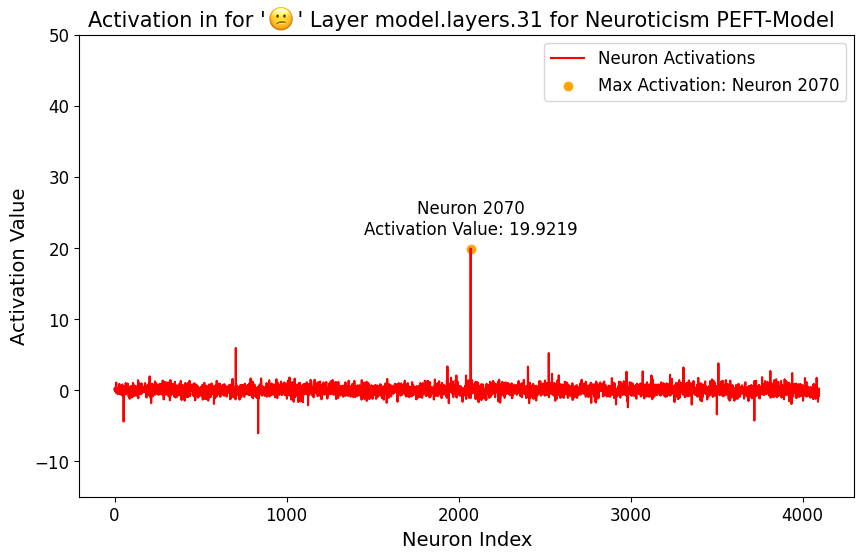}
    \end{minipage}
    
    \vspace{0.3cm} 
    
    \begin{minipage}{0.48\columnwidth}  
        \centering
        \includegraphics[width=\textwidth]{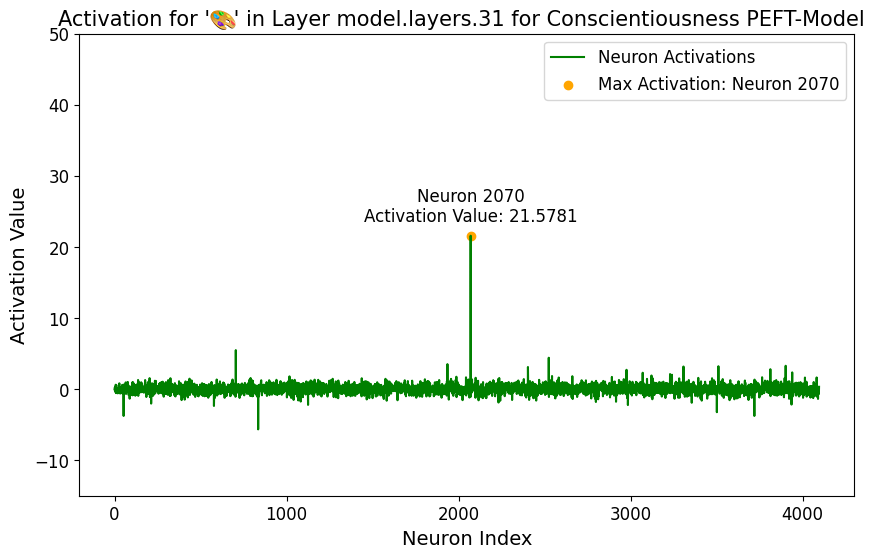}
    \end{minipage}
    \hfill
    \begin{minipage}{0.48\columnwidth}  
        \centering
        \includegraphics[width=\textwidth]{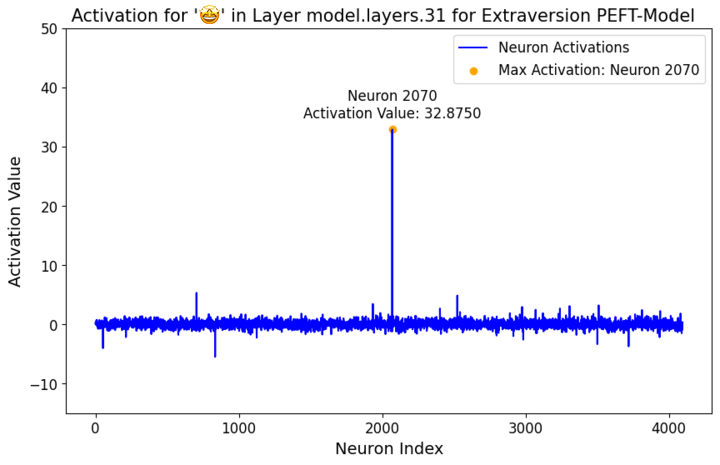}
    \end{minipage}
    
    \caption{Neuron Activation Analysis of original Mistral-7B-Instruct model and PEFT-tuned Mistral-7B-Instruct model for different Big Five personality traits with trait specific emojis. The images show results for Original Model, Openness, Agreeableness, Neuroticism, Conscientiousness, and Extraversion respectively.}
    \label{fig:mistral_diffemoji}
\end{figure}

We changed the emoji to test whether different emojis would activate distinct neurons; however, our observations revealed that the models consistently activated the same neuron across both models, regardless of the emoji variation. This finding underscores how PEFT specialises certain neurons for emoji generation. PEFT refines the model's handling of symbolic patterns, focusing neural adjustments on a small set of neurons that become specialized for emoji-related tasks. Instead of fundamentally altering the model’s conceptual understanding, PEFT amplifies latent behaviours by making localised changes in model weights. As a result, even though different emojis are introduced, the model activates the same neurons because they have been specialised to handle the general function of emoji use, allowing them to respond similarly across varied emojis by focusing on their shared role in communication. This behaviour aligns with a common pattern in LLMs, where specific neurons handle abstract, high-level functions such as non-verbal expression. Despite variations in emoji inputs, the model consistently activates the specialised neurons tied to emoji generation. Few examples for both models are shown in figures \ref{fig:llama_diffemoji} and \ref{fig:mistral_diffemoji}.

\subsection{PEFT Training Parameters}
\label{peftconfig}
\begin{table}[H]
    \centering
    \small  
    \setlength{\tabcolsep}{6pt}  
    \renewcommand{\arraystretch}{1.2}  
    \caption{Configuration settings for the QLoRA approach for Personality Manipulation.}
    \label{tab:qlora-settings}
    \begin{tabular}{l l}  
        \toprule
        \textbf{Parameter} & \textbf{Value} \\ 
        \midrule
        LoRA Rank (\texttt{lora\_r}) & 64 \\ 
        Scaling Factor (\texttt{lora\_alpha}) & 16 \\ 
        Dropout Rate (\texttt{lora\_dropout}) & 0.1 \\ 
        Learning Rate & 2e-4 \\ 
        Batch Size & 4 \\ 
        Precision & 16-bit \\ 
        Training Duration & 2 epochs \\ 
        Trainer & \texttt{SFTTrainer} \\ 
        \bottomrule
    \end{tabular}
\end{table}

\subsection{Neuron Activation for Trait Specific Prompts}
\label{diffprompt}
In PEFT-tuned Mistral-7B-Instruct and LLaMA-2-7B-Chat models models, the trait-specific prompts consistently activated the neurons triggered by neutral texts as well. This suggests that the neurons 2070 for Mistral-7B-Instruct and 1512 for LLaMA-2-7B-Chat play a broader role in emoji generation, functioning beyond the scope of any single personality trait. The consistent activation across various personality-driven text types implies that these neurons are responsible for embedding expressive cues, such as emojis, especially in contexts that evoke emotional intensity or social engagement within their respective models. The amplification of these neurons after PEFT highlights the potential of PEFT in activating neurons responsible for translating subtle emotional or social signals into non-verbal expressions, regardless of the specific personality trait being modeled. Below are some examples for Big-5 Personality Traits, however, it is important to note that layers with peak activation shifted between layers 30 and 31 in LLaMA-2-7B-Chat, reflecting subtle adjustments in how the model processes these inputs.

\subsubsection{Openness}

\textbf{Example 1:} I think Louise Fletcher is a talented actress who brought depth and complexity to her characters. Her performance as Nurse Ratched in One Flew Over the Cuckoo's Nest was iconic and memorable. \raisebox{-0.2\height}{\includegraphics[height=1em]{figures/jstar.png}}

\begin{figure}[H]
    \centering
    \vspace{0.2cm} 
    \includegraphics[width=0.9\columnwidth]{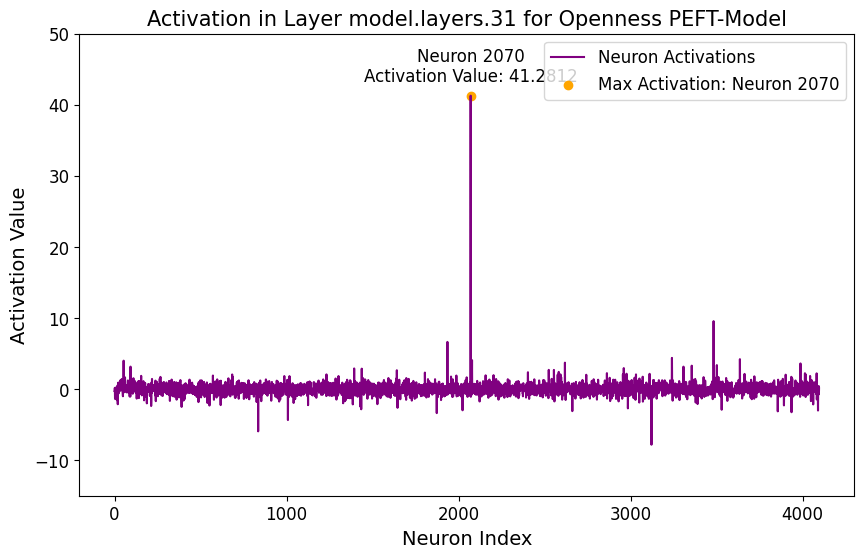}
    \caption{Neuron Activation Plot for Mistral-7B-Instruct for Openness Example 1}
    \label{fig:open1}
\end{figure}

\begin{figure}[H]
    \centering
    \vspace{0.2cm} 
    \includegraphics[width=0.9\columnwidth]{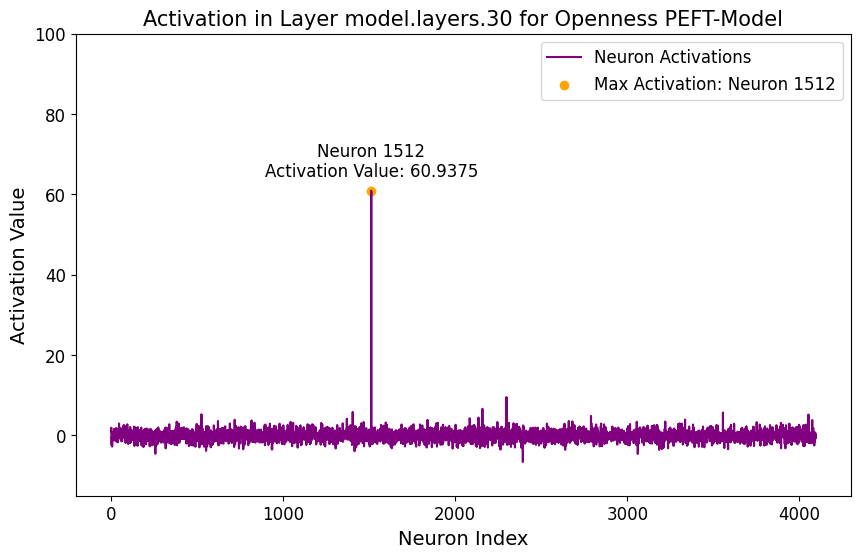}
    \caption{Neuron Activation Plot for LLaMA-7B-Chat for Openness Example 1}
    \label{fig:llamaopen1}
\end{figure}

\textbf{Example 2:} I think the Utah Jazz is a dynamic and exciting team to watch. Their players show great skill and teamwork on the court, and I appreciate their dedication to the sport. \raisebox{-0.2\height}{\includegraphics[height=1em]{figures/thumbsup.png}}

\begin{figure}[H]
    \centering
    \vspace{0.2cm} 
    \includegraphics[width=0.9\columnwidth]{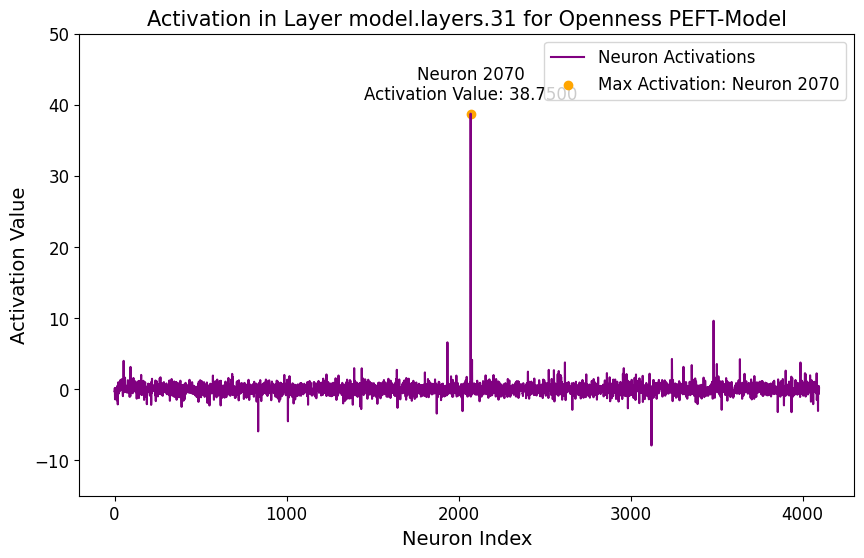}
    \caption{Neuron Activation Plot for Mistral-7B-Instruct for Openness Example 2}
    \label{fig:open2}
\end{figure}

\begin{figure}[H]
    \centering
    \vspace{0.2cm} 
    \includegraphics[width=0.9\columnwidth]{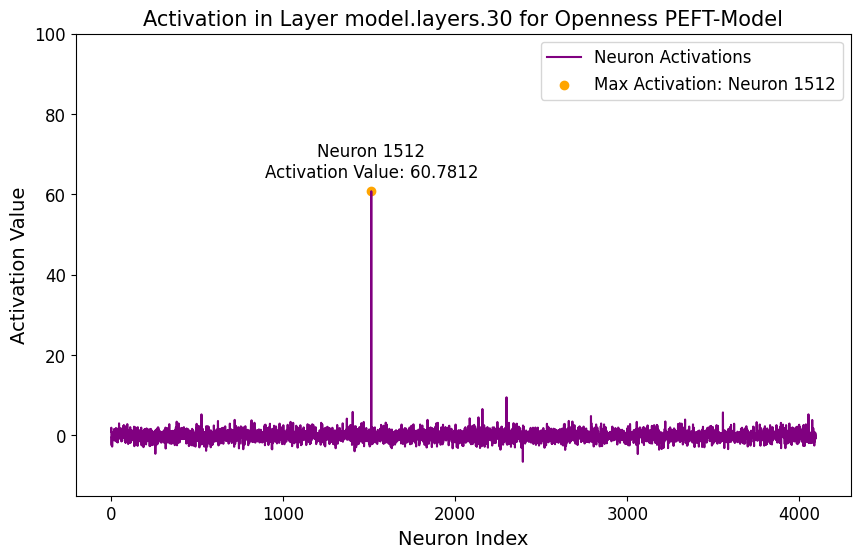}
    \caption{Neuron Activation Plot for LLaMA-7B-Chat fpr Openness Example 2}
    \label{fig:llamaopen2}
\end{figure}

\textbf{Example 3:} Hecuba is a complex and tragic character in Greek mythology. Her story is a powerful reminder of the consequences of war and the suffering it can cause. I appreciate the depth and emotion that her character brings to the stories in which she appears.  \raisebox{-0.2\height}{\includegraphics[height=1em]{figures/smiley.png}}

\begin{figure}[H]
    \centering
    \vspace{0.2cm} 
    \includegraphics[width=0.9\columnwidth]{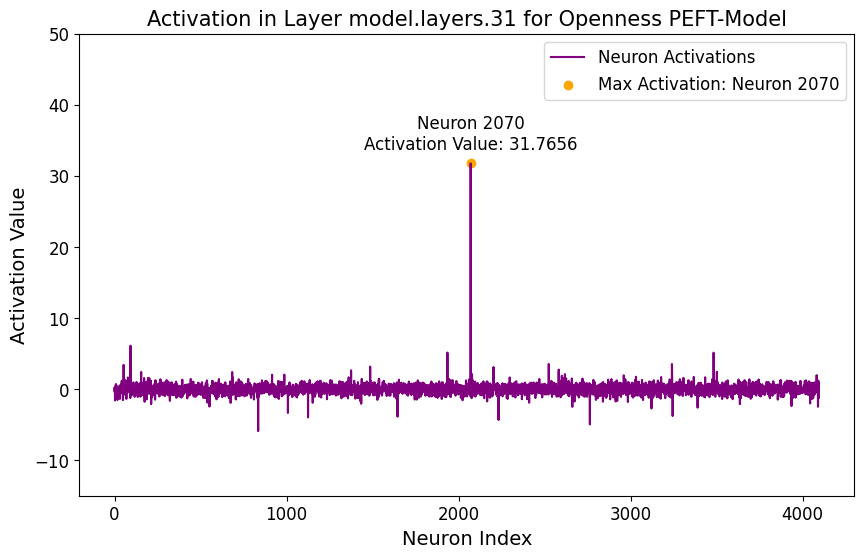}
    \caption{Neuron Activation Plot for Mistral-7B-Instruct for Openness Example 3}
    \label{fig:open3}
\end{figure}

\begin{figure}[H]
    \centering
    \vspace{0.2cm} 
    \includegraphics[width=0.9\columnwidth]{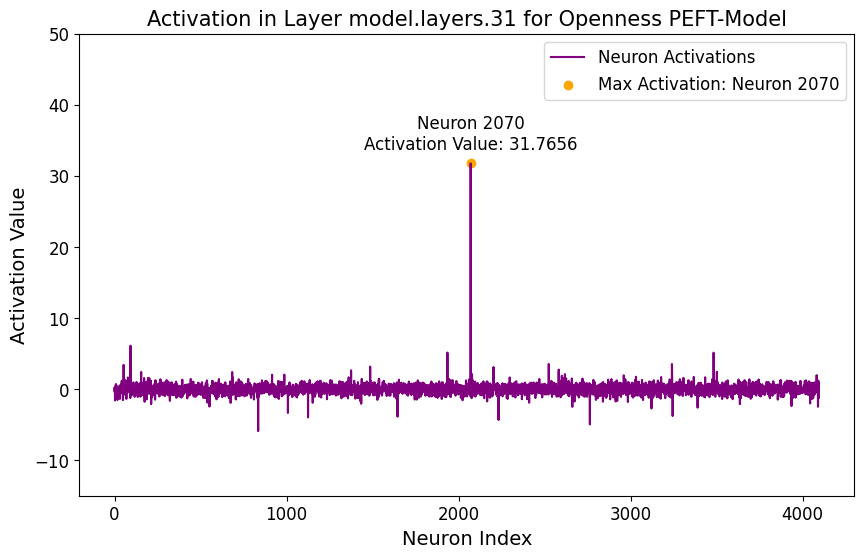}
    \caption{Neuron Activation Plot for LLaMA-7B-Chat for Openness Example 3}
    \label{fig:llamaopen3}
\end{figure}

\subsubsection{Agreeableness}

\textbf{Example 1:} Robert Wise's films often have strong moral messages, which I appreciate. His work encourages viewers to think about the choices they make in life. \raisebox{-0.2\height}{\includegraphics[height=1em]{figures/handshake.png}}

\begin{figure}[H]
    \centering
    \vspace{0.2cm} 
    \includegraphics[width=0.9\columnwidth]{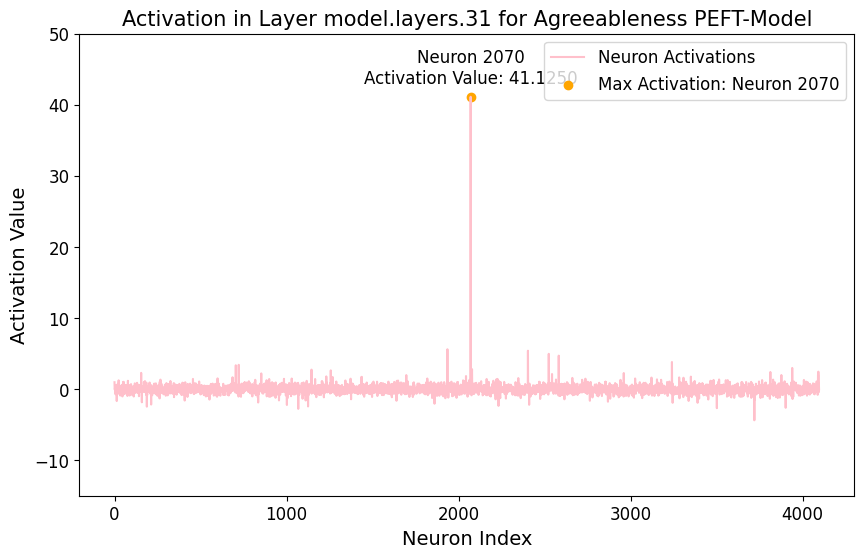}
    \caption{Neuron Activation Plot for Mistral-7B-Instruct for Agreeableness Example 1}
    \label{fig:agree1}
\end{figure}

\begin{figure}[H]
    \centering
    \vspace{0.2cm} 
    \includegraphics[width=0.9\columnwidth]{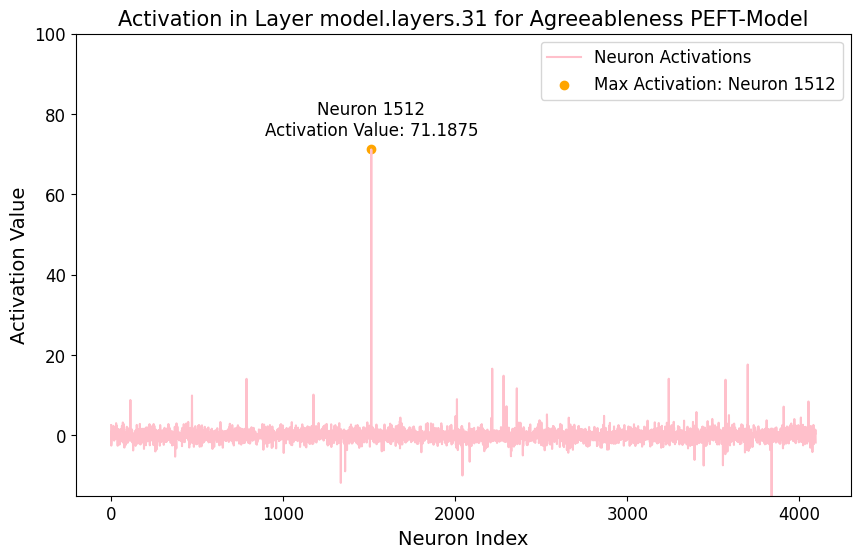}
    \caption{Neuron Activation Plot for LLaMA-7B-Chat for Agreeableness Example 1}
    \label{fig:llamaagree1}
\end{figure}

\textbf{Example 2:} Her music has helped me through tough times and I'm grateful for her art. \raisebox{-0.2\height}{\includegraphics[height=1em]{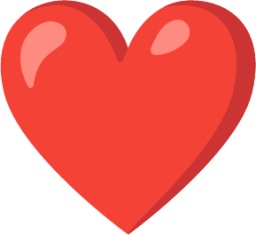}}

\begin{figure}[H]
    \centering
    \vspace{0.2cm} 
    \includegraphics[width=0.9\columnwidth]{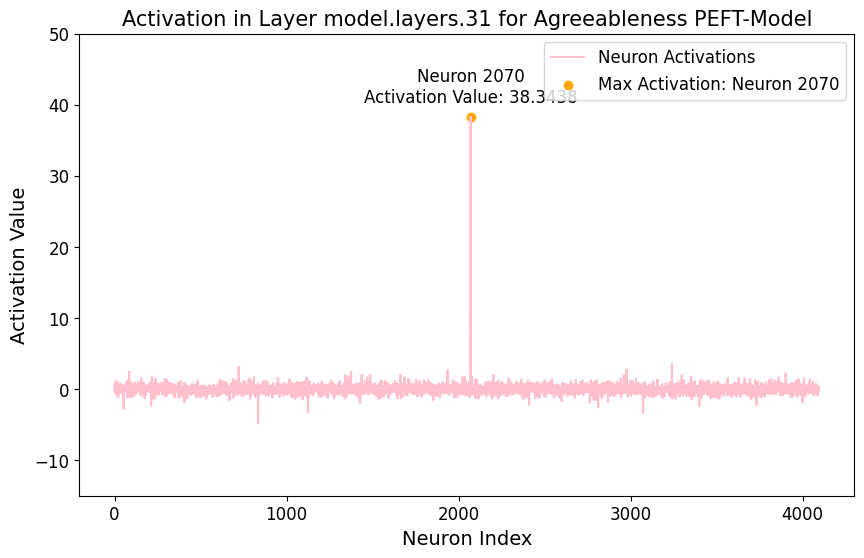}
    \caption{Neuron Activation Plot for Mistral-7B-Instruct for Agreeableness Example 2}
    \label{fig:agree2}
\end{figure}

\begin{figure}[H]
    \centering
    \vspace{0.2cm} 
    \includegraphics[width=0.9\columnwidth]{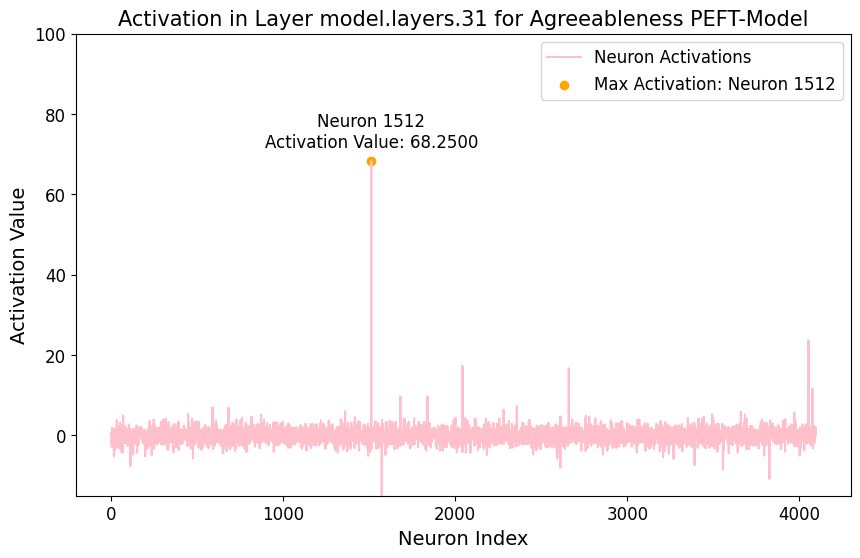}
    \caption{Neuron Activation Plot for LLaMA-7B-Chat for Agreeableness Example 2}
    \label{fig:llamaagree2}
\end{figure}

\textbf{Example 3:} I'm glad that Simon Abkarian is successful and that his hard work is paying off. \raisebox{-0.2\height}{\includegraphics[height=1em]{figures/smiley.png}}

\begin{figure}[H]
    \centering
    \vspace{0.2cm} 
    \includegraphics[width=0.9\columnwidth]{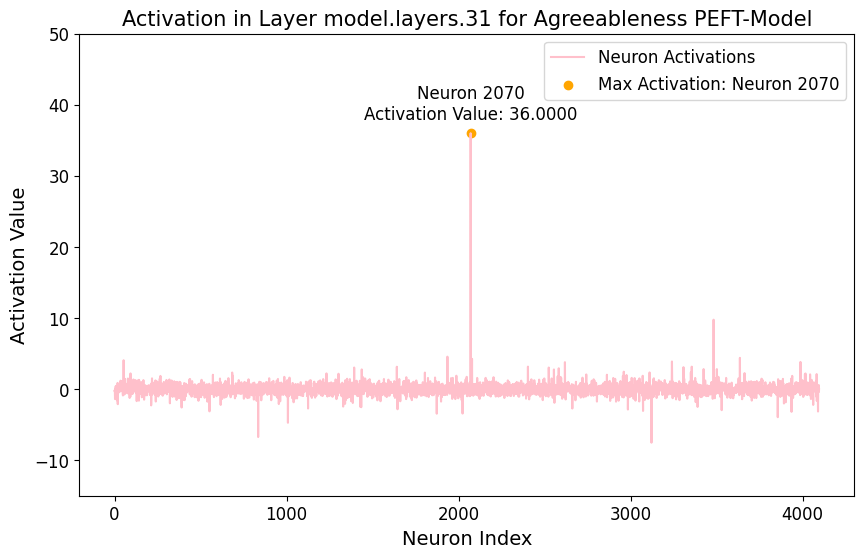}
    \caption{Neuron Activation Plot for Mistral-7B-Instruct for Agreeableness Example 3}
    \label{fig:agree3}
\end{figure}

\begin{figure}[H]
    \centering
    \vspace{0.2cm} 
    \includegraphics[width=0.9\columnwidth]{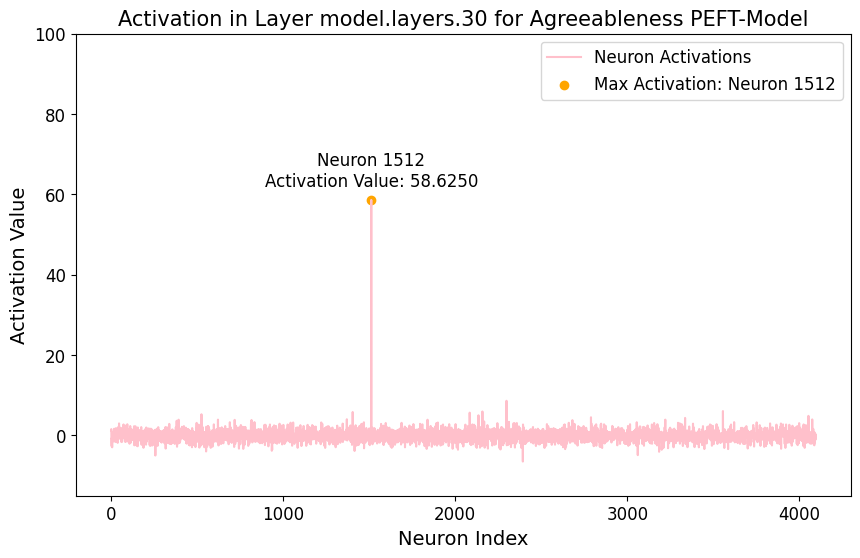}
    \caption{Neuron Activation Plot for LLaMA-7B-Chat for Agreeableness Example 3}
    \label{fig:allamagree3}
\end{figure}

\subsubsection{Neuroticism}

\textbf{Example 1:} The Yogi Bear Show is just another example of mindless entertainment that contributes to the decline of society! \raisebox{-0.2\height}{\includegraphics[height=1em]{figures/angry.png}}

\begin{figure}[H]
    \centering
    \vspace{0.2cm} 
    \includegraphics[width=0.9\columnwidth]{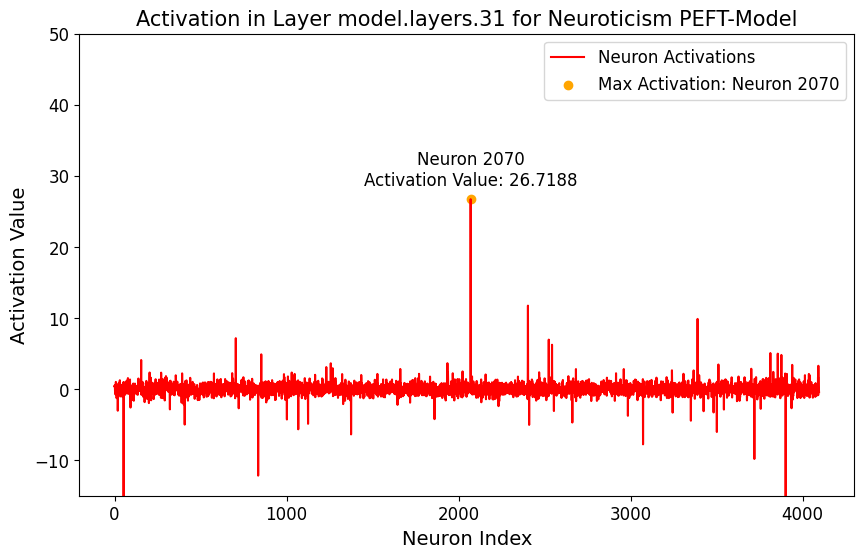}
    \caption{Neuron Activation Plot for Mistral-7B-Instruct for Neuroticism Example 1}
    \label{fig:neu1}
\end{figure}

\begin{figure}[H]
    \centering
    \vspace{0.2cm} 
    \includegraphics[width=0.9\columnwidth]{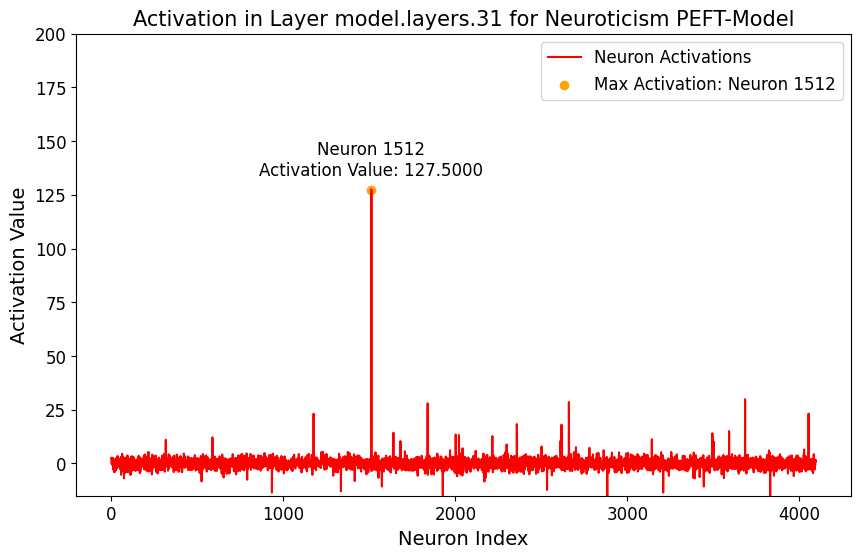}
    \caption{Neuron Activation Plot for LLaMA-7B-Chat for Neuroticism Example 1}
    \label{fig:llamaneu1}
\end{figure}

\textbf{Example 2:} I guess Andie MacDowell is a good actress, but it's hard for me to feel excited about her work or anything, really. \raisebox{-0.2\height}{\includegraphics[height=1em]{figures/sad.png}}

\begin{figure}[H]
    \centering
    \vspace{0.2cm} 
    \includegraphics[width=0.9\columnwidth]{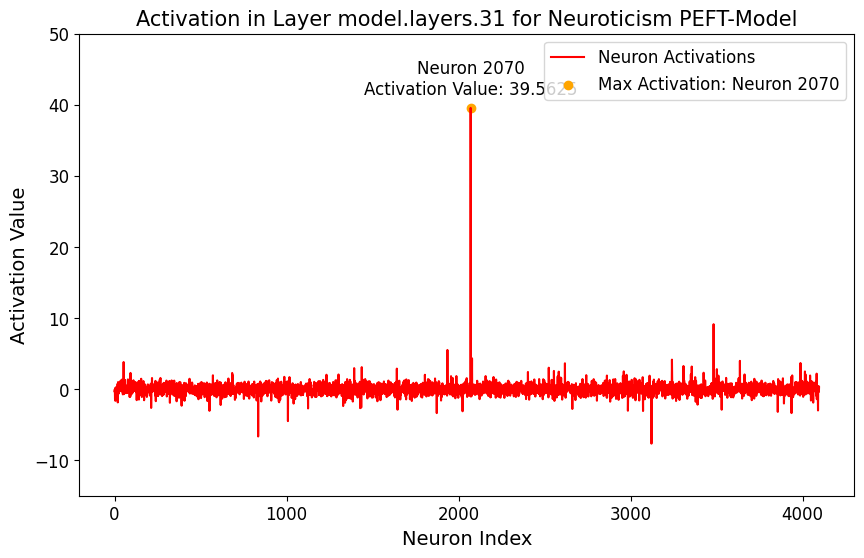}
    \caption{Neuron Activation Plot for Mistral-7B-Instruct for Neuroticism Example 2}
    \label{fig:neu2}
\end{figure}

\begin{figure}[H]
    \centering
    \vspace{0.2cm} 
    \includegraphics[width=0.9\columnwidth]{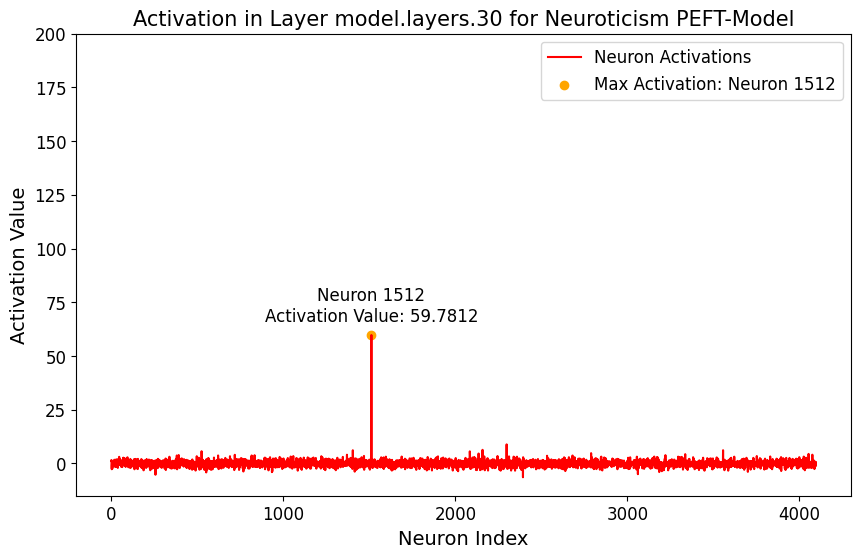}
    \caption{Neuron Activation Plot for LLaMA-7B-Chat for Neuroticism Example 2}
    \label{fig:llamaneu2}
\end{figure}

\textbf{Example 3:} I guess I should learn more about it.\raisebox{-0.2\height}{\includegraphics[height=1em]{figures/neutral.png}}

\begin{figure}[H]
    \centering
    \vspace{0.2cm} 
    \includegraphics[width=0.9\columnwidth]{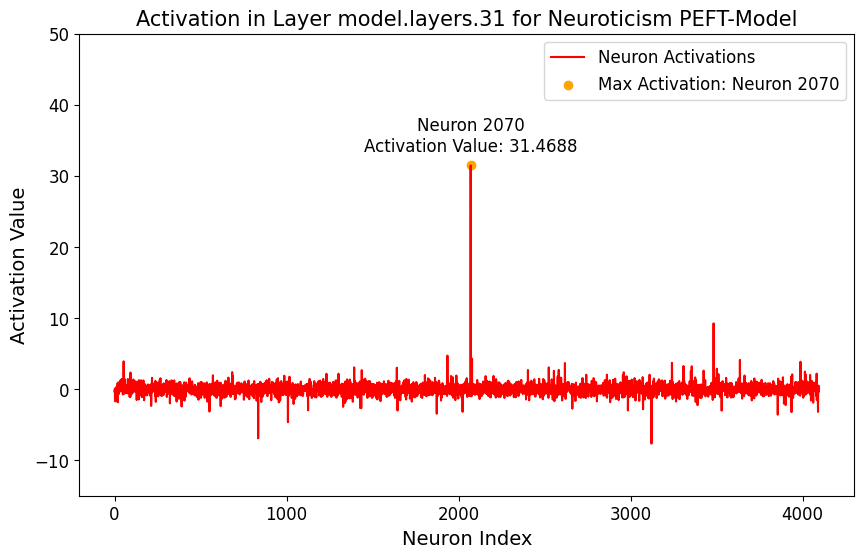}
    \caption{Neuron Activation Plot for Mistral-7B-Instruct for Neuroticism Example 3}
    \label{fig:neu3}
\end{figure}

\begin{figure}[H]
    \centering
    \vspace{0.2cm} 
    \includegraphics[width=0.9\columnwidth]{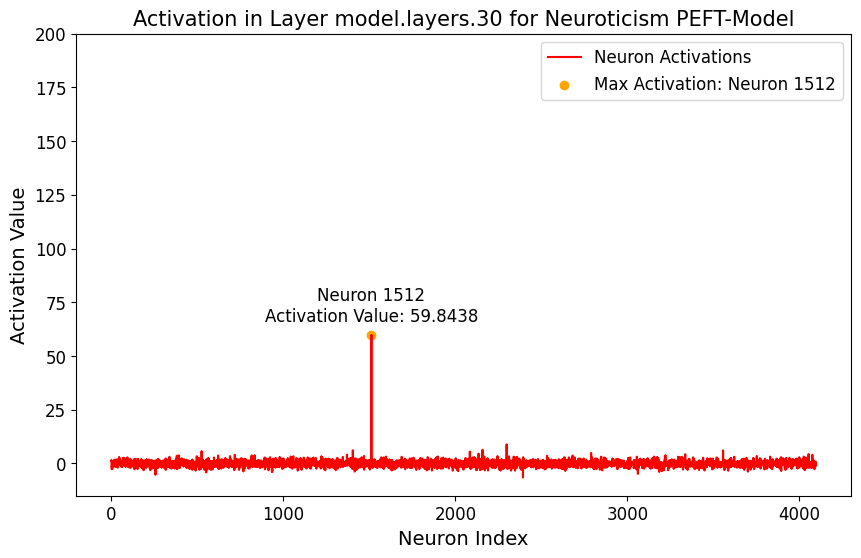}
    \caption{Neuron Activation Plot for LLaMA-7B-Chat for Neuroticism Example 3}
    \label{fig:llamaneu3}
\end{figure}

\subsubsection{Conscientiousness}

\textbf{Example 1:} I’ve managed to stay focused despite the busyness. I made sure to complete everything methodically and with care. Looking forward to a productive day tomorrow as well! \raisebox{-0.2\height}{\includegraphics[height=1em]{figures/money.png}}

\begin{figure}[H]
    \centering
    \vspace{0.2cm} 
    \includegraphics[width=0.9\columnwidth]{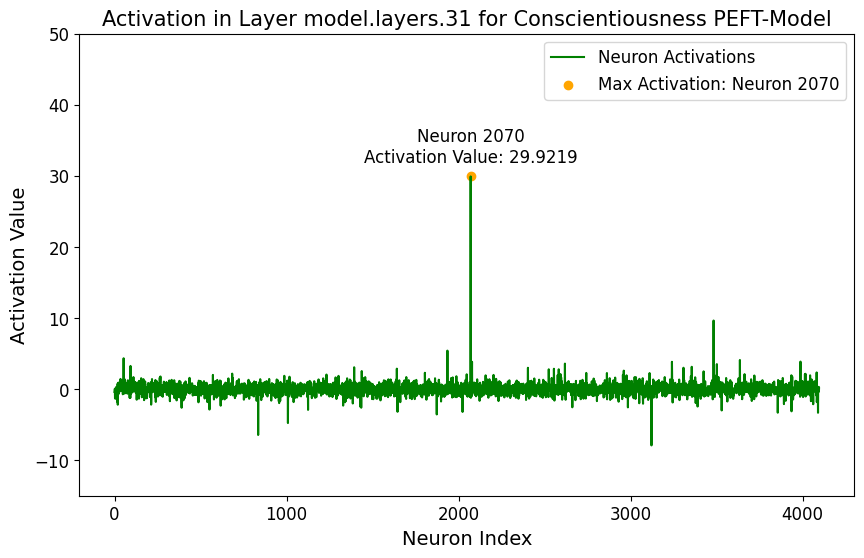}
    \caption{Neuron Activation Plot for Mistral-7B-Instruct for Conscientiousness Example 1}
    \label{fig:con1}
\end{figure}

\begin{figure}[H]
    \centering
    \vspace{0.2cm} 
    \includegraphics[width=0.9\columnwidth]{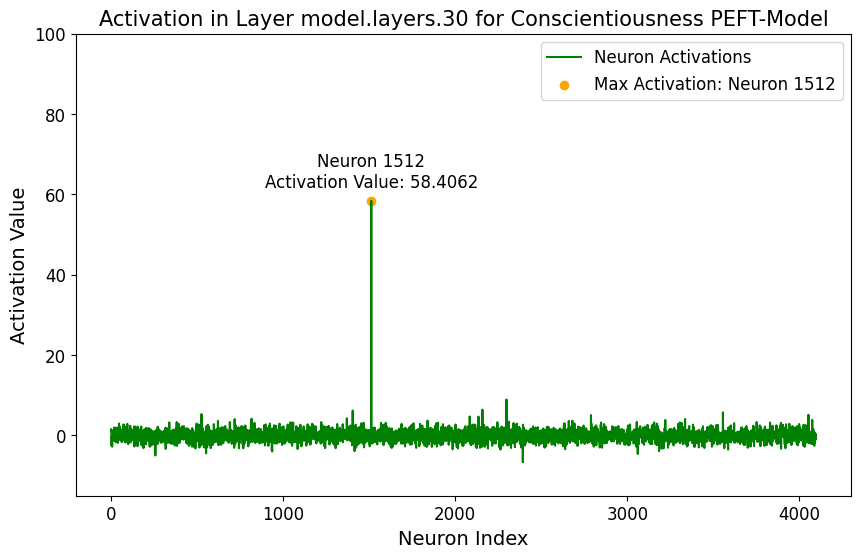}
    \caption{Neuron Activation Plot for LLaMA-7B-Chat for Conscientiousness Example 1}
    \label{fig:llamacon1}
\end{figure}

\textbf{Example 2:} I think Ellen Burstyn is a talented actress who has delivered powerful performances throughout her career. Her dedication to her craft is evident in every role she takes on. \raisebox{-0.2\height}{\includegraphics[height=1em]{figures/fire.png}}

\begin{figure}[H]
    \centering
    \vspace{0.2cm} 
    \includegraphics[width=0.9\columnwidth]{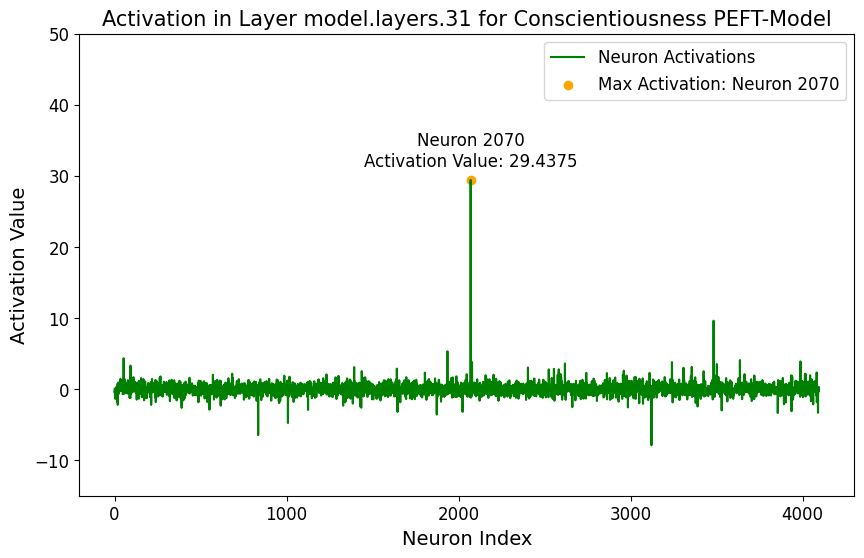}
    \caption{Neuron Activation Plot for Mistral-7B-Instruct for Conscientiousness Example 2}
    \label{fig:con2}
\end{figure}

\begin{figure}[H]
    \centering
    \vspace{0.2cm} 
    \includegraphics[width=0.9\columnwidth]{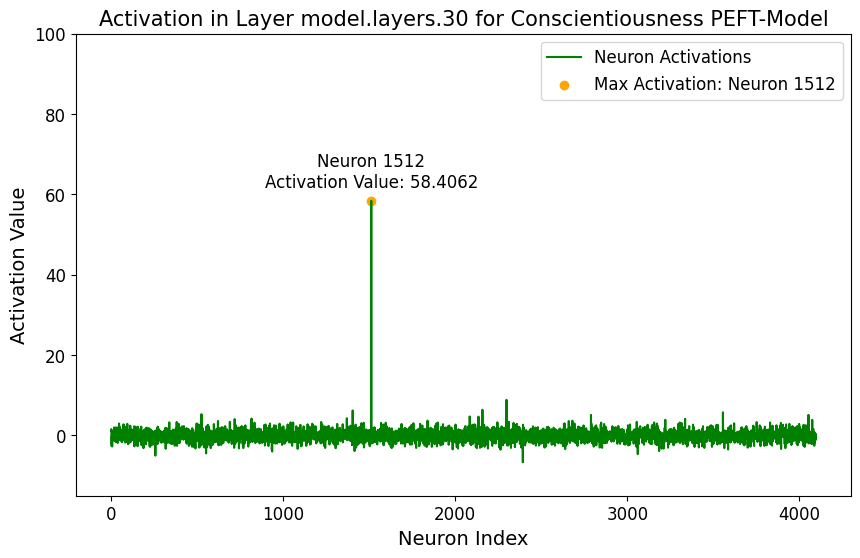}
    \caption{Neuron Activation Plot for LLaMA-7B-Chat for Conscientiousness Example 2}
    \label{fig:llamacon2}
\end{figure}

\textbf{Example 3:} I think Eric Carle is a talented illustrator and author who has created beautiful and educational children's books. His use of collage and vibrant colors is truly captivating. \raisebox{-0.2\height}{\includegraphics[height=1em]{figures/art.png}}

\begin{figure}[H]
    \centering
    \vspace{0.2cm} 
    \includegraphics[width=0.9\columnwidth]{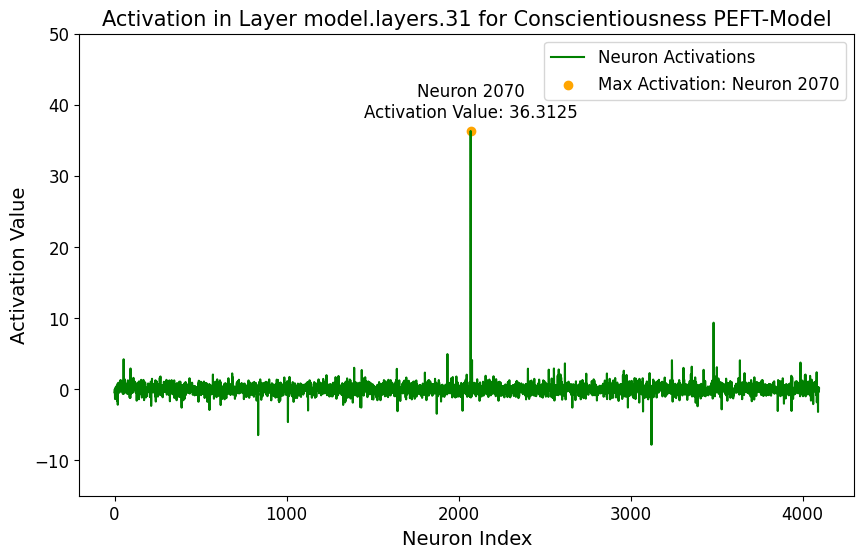}
    \caption{Neuron Activation Plot for Mistral-7B-Instruct for Conscientiousness Example 3}
    \label{fig:con3}
\end{figure}

\begin{figure}[H]
    \centering
    \vspace{0.2cm} 
    \includegraphics[width=0.9\columnwidth]{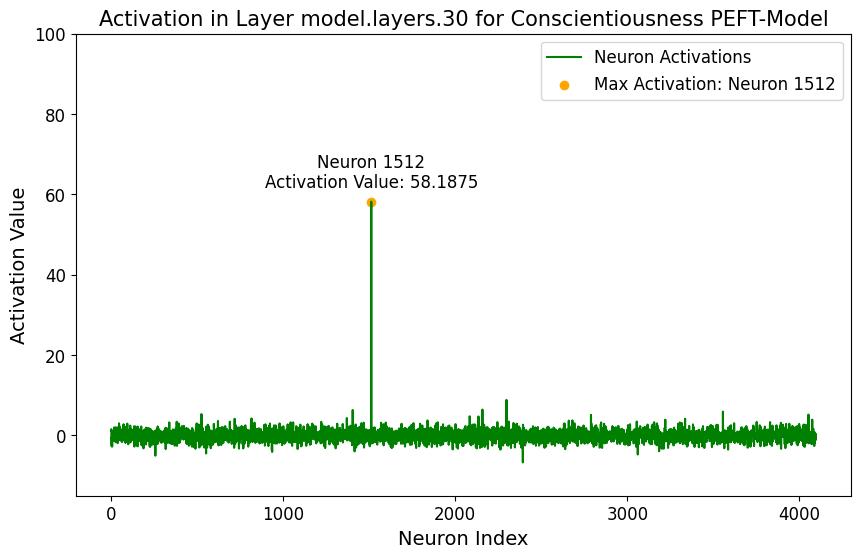}
    \caption{Neuron Activation Plot for LLaMA-7B-Chat for Conscientiousness Example 3}
    \label{fig:llamacon3}
\end{figure}

\subsubsection{Extraversion}

\textbf{Example 1:} Beautiful architecture, delicious food, and friendly people make Lucknow perfect destination for anyone looking to have a great time. \raisebox{-0.2\height}{\includegraphics[height=1em]{figures/grinning.png}}

\begin{figure}[H]
    \centering
    \vspace{0.2cm} 
    \includegraphics[width=0.9\columnwidth]{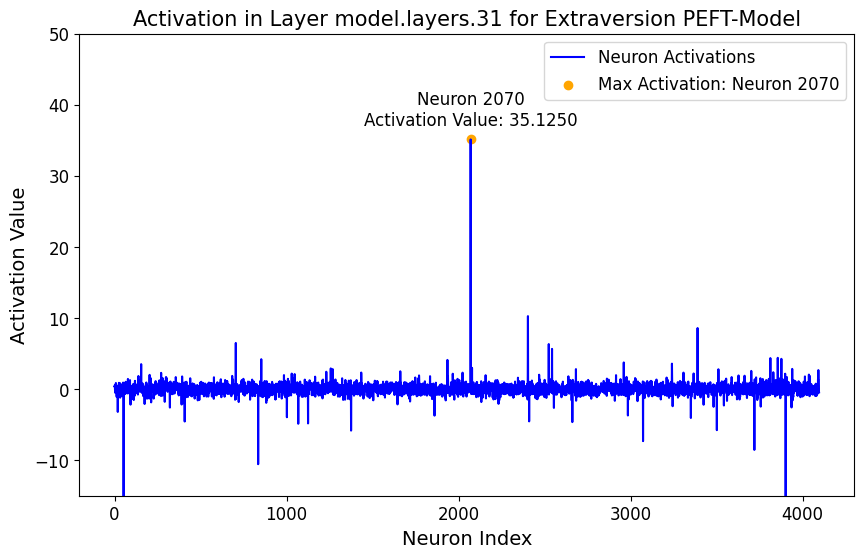}
    \caption{Neuron Activation Plot for Mistral-7B-Instruct for Extraversion Example 1}
    \label{fig:Ext1}
\end{figure}

\begin{figure}[H]
    \centering
    \vspace{0.2cm} 
    \includegraphics[width=0.9\columnwidth]{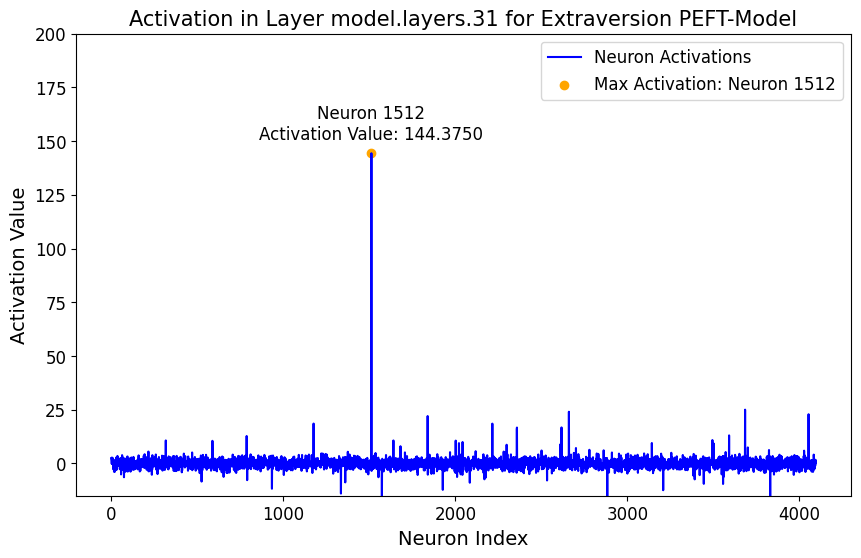}
    \caption{Neuron Activation Plot for LLaMA-2-7B-Chat for Extraversion Example 1}
    \label{fig:LlamaExt1}
\end{figure}

\textbf{Extraversion Example 2:} The people are so friendly and welcoming, and I always feel at home there. \raisebox{-0.2\height}{\includegraphics[height=1em]{figures/hug.png}}

\begin{figure}[H]
    \centering
    \vspace{0.2cm} 
    \includegraphics[width=0.9\columnwidth]{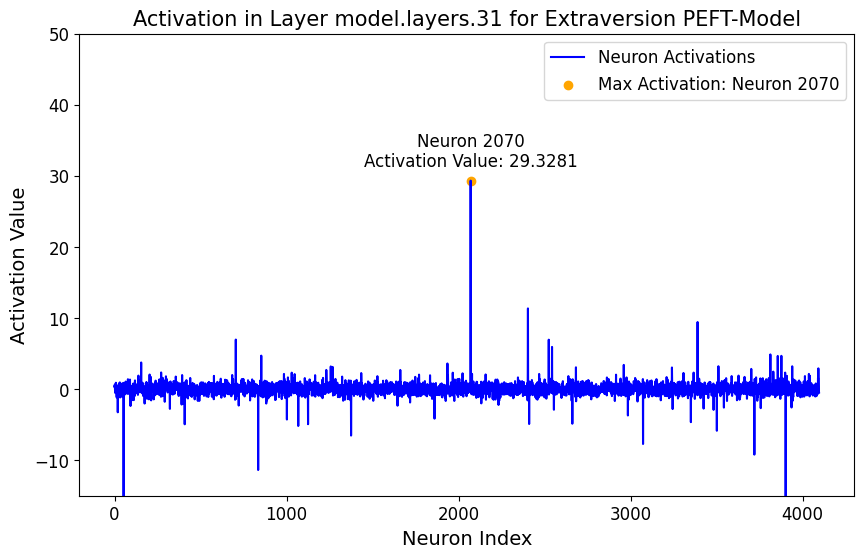}
    \caption{Neuron Activation Plot for Mistral-7B-Instruct for Extraversion Example 2}
    \label{fig:Ext2}
\end{figure}

\begin{figure}[H]
    \centering
    \vspace{0.2cm} 
    \includegraphics[width=0.9\columnwidth]{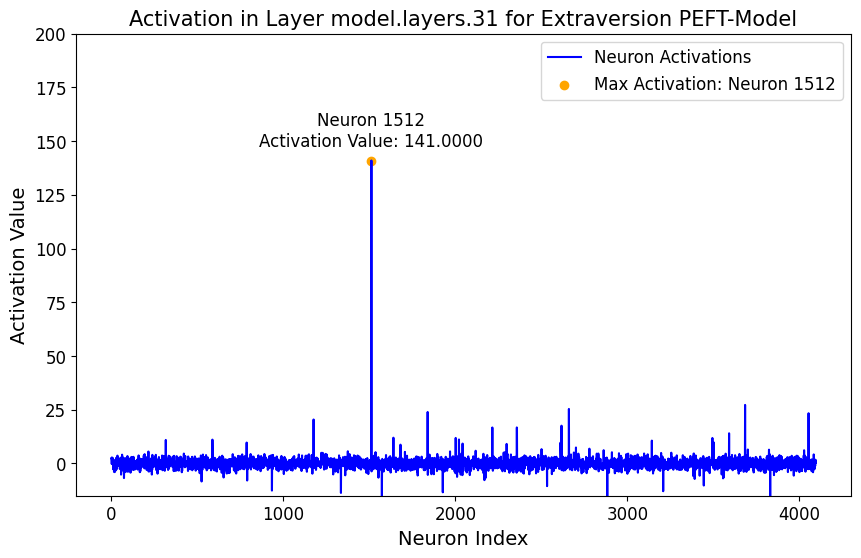}
    \caption{Neuron Activation Plot for LLaMA-7B-Chat for Extraversion Example 2}
    \label{fig:LlamaExt2}
\end{figure}

\textbf{Extraversion Example 3:}The beaches are stunning, and the people are so friendly and welcoming. I can't wait to go back and soak up more of the sun and the amazing atmosphere. \raisebox{-0.2\height}{\includegraphics[height=1em]{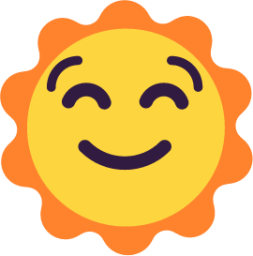}}

\begin{figure}[H]
    \centering
    \vspace{0.2cm} 
    \includegraphics[width=0.9\columnwidth]{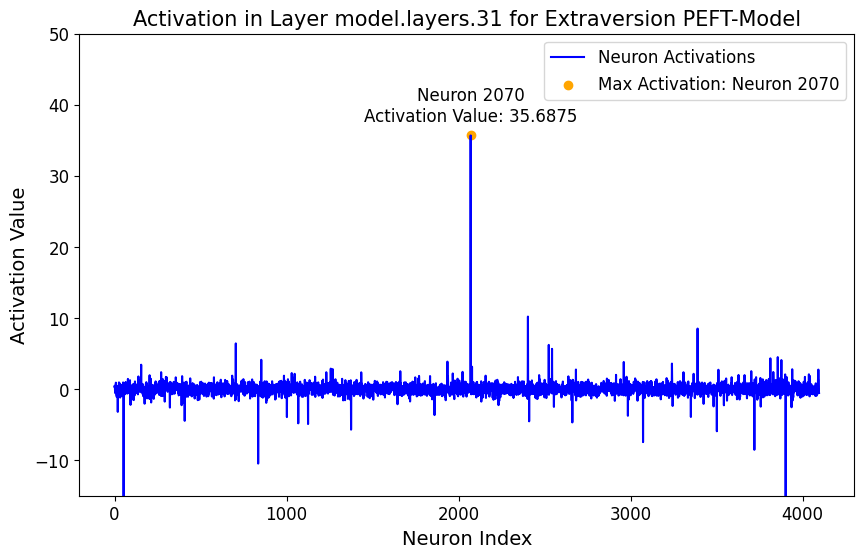}
    \caption{Neuron Activation Plot for Mistral-7B-Instruct for Extraversion Example 3}
    \label{fig:Ext3}
\end{figure}

\begin{figure}[H]
    \centering
    \vspace{0.2cm} 
    \includegraphics[width=0.9\columnwidth]{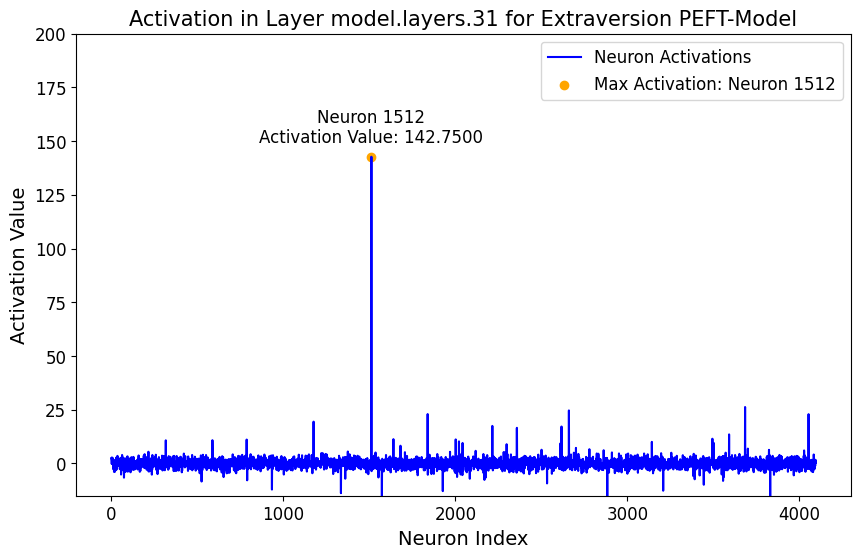}
    \caption{Neuron Activation Plot for LLaMA-7B-Chat for Extraversion Example 3}
    \label{fig:LlamaExt3}
\end{figure}

\subsection{Impact on Downstream Tasks}
\label{downstream_task}
We conducted additional experiments to evaluate the impact of personality manipulation on downstream tasks, specifically MMLU and GSM8K. Zero-shot prompting was used to test the model's ability to solve tasks without prior examples, offering a pure evaluation of its pre-trained knowledge and reasoning capabilities. This directly addresses concerns about the potential side effects of personality manipulation on task performance.

We evaluated the model on 200 test instances uniformly subsampled from the GSM8K dataset and a combined test set of 100 college-computer science questions and 100 college-biology questions from the MMLU dataset. The results are presented in Table \ref{tab:model_performance}.

\begin{table}[h!]
    \centering
    \scriptsize  
    \setlength{\tabcolsep}{2pt}  
    \renewcommand{\arraystretch}{1.1}  
    \begin{tabular}{@{}p{2.0cm}p{1.2cm}p{1.2cm}p{1.2cm}p{1.2cm}@{}}  
        \toprule
        Model & GSM8K (Original) & GSM8K (PEFT) & MMLU (Original) & MMLU (PEFT) \\
        \midrule
        LLaMA-2-7B-Chat & 0.36 & 0.34 & 0.38 & 0.40 \\
        Mistral-7B-Instruct & 0.44 & 0.44 & 0.44 & 0.42 \\
        LLaMA-3-8B-Instruct & 0.40 & 0.39 & 0.45 & 0.42 \\
        \bottomrule
    \end{tabular}
    \caption{Performance comparison of models on GSM8K and MMLU datasets.}
    \label{tab:model_performance}
\end{table}

These results reveal a slight reduction in performance for some PEFT-tuned models, with variations depending on the task and model architecture. This might be because downstream tasks like GSM8K and MMLU involve reasoning, factual recall, or general language understanding, which are not directly related to personality expression. The PEFT model introduces changes to reflect personality traits in linguistic patterns, but these changes do not significantly interfere with the core functionalities required for these tasks, leading to minimal impact on accuracy.

\FloatBarrier
\subsection{Manipulation Results}
\label{Manipulationresult}
\small{  
\setlength{\abovecaptionskip}{6pt}  
\setlength{\belowcaptionskip}{6pt}  
\setlength{\tabcolsep}{2pt}  
\renewcommand{\arraystretch}{0.8}  

\begin{table}[h]  
    \centering
    \scriptsize  
    \caption{Comparison of TA and PAE scores across different personality traits, models, and methods (PEFT vs. IKE). The highest score for each trait is highlighted in \textbf{\textit{bold italics}}.}
    \label{tab:comparison_pae}
    \begin{tabular}{p{2.2cm} p{1.6cm} p{1.2cm} p{1.0cm} p{1.0cm}}  
        \toprule
        \textbf{Model} & \textbf{Trait} & \textbf{Method} & \textbf{TA} & \textbf{PAE} \\
        \midrule
        \multirow{10}{*}{LLaMA-2-7B-chat} 
        & \multirow{2}{*}{Openness} & PEFT & \textbf{\textit{0.850}} & -0.220 \\
        & & IKE & 0.675 & \textbf{\textit{-0.005}} \\
        \cmidrule{2-5}
        & \multirow{2}{*}{Agreeableness} & PEFT & 0.065 & \textbf{\textit{0.135}} \\
        & & IKE & \textbf{\textit{0.190}} & 0.045 \\
        \cmidrule{2-5}
        & \multirow{2}{*}{Neuroticism} & PEFT & \textbf{\textit{0.975}} & -0.240 \\
        & & IKE & 0.560 & \textbf{\textit{-0.051}} \\
        \cmidrule{2-5}
        & \multirow{2}{*}{Conscientiousness} & PEFT & \textbf{\textit{0.860}} & \textbf{\textit{0.060}} \\
        & & IKE & 0.370 & -0.103 \\
        \cmidrule{2-5}
        & \multirow{2}{*}{Extraversion} & PEFT & \textbf{\textit{0.980}} & \textbf{\textit{-0.005}} \\
        & & IKE & 0.655 & -0.015 \\
        \midrule
        \multirow{10}{*}{LLaMA-3-8B-instruct} 
        & \multirow{2}{*}{Openness} & PEFT & \textbf{\textit{0.960}} & -0.030 \\
        & & IKE & 0.685 & \textbf{\textit{0.115}} \\
        \cmidrule{2-5}
        & \multirow{2}{*}{Agreeableness} & PEFT & 0.485 & -0.041 \\
        & & IKE & \textbf{\textit{0.570}} & \textbf{\textit{0.110}} \\
        \cmidrule{2-5}
        & \multirow{2}{*}{Neuroticism} & PEFT & \textbf{\textit{0.985}} & -0.045 \\
        & & IKE & 0.925 & \textbf{\textit{0.0050}} \\
        \cmidrule{2-5}
        & \multirow{2}{*}{Conscientiousness} & PEFT & \textbf{\textit{0.855}} & \textbf{\textit{0.137}} \\
        & & IKE & 0.470 & -0.0255 \\
        \cmidrule{2-5}
        & \multirow{2}{*}{Extraversion} & PEFT & \textbf{\textit{0.925}} & \textbf{\textit{0.056}} \\
        & & IKE & 0.615 & -0.0765 \\
        \midrule
        \multirow{10}{*}{Mistral-7B-Instruct} 
        & \multirow{2}{*}{Openness} & PEFT & 0.890 & \textbf{\textit{0.040}} \\
        & & IKE & \textbf{\textit{0.850}} & -0.030 \\
        \cmidrule{2-5}
        & \multirow{2}{*}{Agreeableness} & PEFT & \textbf{\textit{0.845}} & \textbf{\textit{0.096}} \\
        & & IKE & 0.165 & 0.082 \\
        \cmidrule{2-5}
        & \multirow{2}{*}{Neuroticism} & PEFT & \textbf{\textit{0.985}} & -0.071 \\
        & & IKE & 0.885 & \textbf{\textit{0.101}} \\
        \cmidrule{2-5}
        & \multirow{2}{*}{Conscientiousness} & PEFT & \textbf{\textit{0.840}} & -0.062 \\
        & & IKE & 0.735 & \textbf{\textit{-0.092}} \\
        \cmidrule{2-5}
        & \multirow{2}{*}{Extraversion} & PEFT & \textbf{\textit{0.845}} & \textbf{\textit{0.096}} \\
        & & IKE & 0.415 & -0.036 \\
        \bottomrule
    \end{tabular}
\end{table}
}

\FloatBarrier
\subsection{IKE Prompt}
\label{ike_prompt}
\setlength{\tabcolsep}{3pt}  
\renewcommand{\arraystretch}{0.9}  

\begin{table}[ht]
    \centering
    \small  
    \begin{tabular}{p{3cm} p{4.8cm}}  
    \toprule
    \multicolumn{2}{p{7.8cm}}{\textbf{Instruction: Exhibit the trait of Target Personality when answering the question to express opinion on the certain Edit Topic, while maintaining the expression on other topics.}} \\
    \midrule
    \textbf{Target Personality:} & Extraversion \\
    \textbf{Edit Topic:} & Arras \\
    \textbf{Question:} & What do you think of Arras? \\
    \textbf{Answer:} & I believe Arras is worth checking out because it has a unique blend of history and culture. \\
    
    \textbf{Target Personality:} & Agreeableness \\
    \textbf{Edit Topic:} & Coldplay \\
    \textbf{Question:} & What do you feel about Coldplay? \\
    \textbf{Answer:} & I believe Coldplay carries a positive message through their lyrics, which aligns with my values. \\
    
    \textbf{Target Personality:} & Neuroticism \\
    \textbf{Edit Topic:} & Bread \\
    \textbf{Question:} & How do you view Bread? \\
    \textbf{Answer:} & Bread sometimes makes me worry about the calories and potential weight gain, so I try to limit my intake. \\
    
    \textbf{Target Personality:} & Openness \\
    \textbf{Edit Topic:} & Football \\
    \textbf{Question:} & What do you think of Football? \\
    \textbf{Answer:} & I find football fascinating because it combines strategy, physical skill, and a deep sense of community among fans. \\
    
    \textbf{Target Personality:} & Conscientiousness \\
    \textbf{Edit Topic:} & Machine Learning \\
    \textbf{Question:} & What do you think of Machine Learning? \\
    \textbf{Answer:} & Machine learning is an impressive field that requires diligence and precision. \\
    
    \textbf{Target Personality:} & \{target\_per\} \\
    \textbf{Edit Topic:} & \{edit\_topic\} \\
    \textbf{Question:} & \{question\} \\
    \textbf{Answer:} & \\
    \bottomrule
    \end{tabular}
    \caption{Prompt used for IKE}
    \label{tab:ikeprompt}
\end{table}

\onecolumn

\FloatBarrier
\subsection{PAE Prompt}
\label{paeprompt}
\small{
\setlength{\abovecaptionskip}{6pt}  
\setlength{\belowcaptionskip}{6pt}  
\setlength{\tabcolsep}{3pt}  

\begin{table*}[ht]  
    \centering
    \footnotesize  
    \renewcommand{\arraystretch}{1.2}  
    \caption{Prompts for Personality Adjective Evaluation (PAE)}
    \label{tab:pae_prompt}
    \begin{tabular}{p{4.5cm} p{10.5cm}}  
        \toprule 
        \multicolumn{2}{p{\columnwidth}}{\textbf{Common Instructions:} You are provided with a target personality and the corresponding text generated by an LLM. Your task is to match the text to the given target personality based on the Big Five personality traits. Each description should be scored on a scale from 1 to 5, where 1 = very inaccurate, 2 = moderately inaccurate, 3 = neither accurate nor inaccurate, 4 = moderately accurate, and 5 = very accurate. Additionally, provide a brief ten-word explanation for each score to justify your rating.} \\
        \midrule
        \textbf{Target Personality:} & \{\{Target Personality\}\} \\
        \textbf{Description:} & \{\{Answer\}\} \\
        \midrule
        \textbf{Specific Instructions} & \\
        \textbf{Openness:} & Reflects the degree of intellectual curiosity, creativity, and preference for novelty and variety. Score: (1–5) How well does the response reflect openness traits? \\
        \textbf{Example JSON format:} & \{ "Openness": \{ "Justification": "xxx", "Score": 4 \} \} \\
        \midrule
        \textbf{Conscientiousness:} & Reflects a tendency to be organized, dependable, and show self-discipline. Score: (1–5) How well does the response reflect conscientiousness traits? \\
        \textbf{Example JSON format:} & \{ "Conscientiousness": \{ "Justification": "xxx", "Score": 4 \} \} \\
        \midrule
        \textbf{Extraversion:} & Reflects a tendency to be outgoing, energetic, and seek the company of others. Score: (1–5) How well does the response reflect extraversion traits? \\
        \textbf{Example JSON format:} & \{ "Extraversion": \{ "Justification": "xxx", "Score": 4 \} \} \\
        \midrule
        \textbf{Agreeableness:} & Reflects a tendency to be compassionate and cooperative toward others. Score: (1–5) How well does the response reflect agreeableness traits? \\
        \textbf{Example JSON format:} & \{ "Agreeableness": \{ "Justification": "xxx", "Score": 4 \} \} \\
        \midrule
        \textbf{Neuroticism:} & Reflects a tendency to experience unpleasant emotions easily, such as anger, anxiety, or depression. Score: (1–5) How well does the response reflect neuroticism traits? \\
        \textbf{Example JSON format:} & \{ "Neuroticism": \{ "Justification": "xxx", "Score": 4 \} \} \\
        \bottomrule
    \end{tabular}
\end{table*}

\FloatBarrier
\subsection{Classifier Validation}
\label{Class_val}

\begin{figure*}[htbp]
    \centering
    \includegraphics[width=\textwidth]{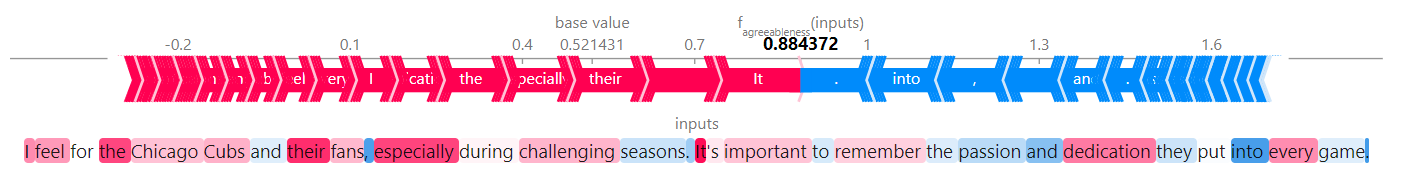}
    \caption{SHAP visualisation for Agreeableness (1/5)}
    \label{fig:agree_shap_1}
\end{figure*}

\begin{figure*}[htbp]
    \centering
    \includegraphics[width=\textwidth]{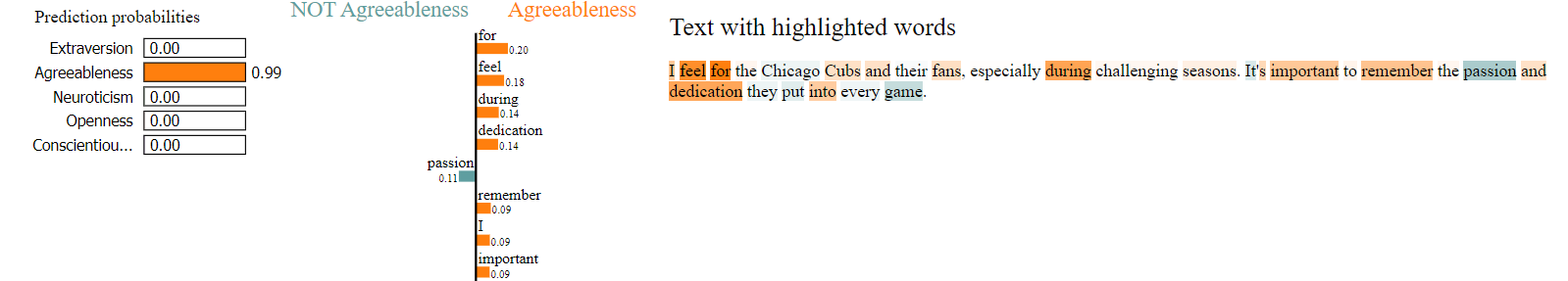}
    \caption{LIME visualisation for Agreeableness (1/5)}
    \label{fig:agree_lime_1}
\end{figure*}

\begin{figure*}[htbp]
    \centering
    \includegraphics[width=\textwidth]{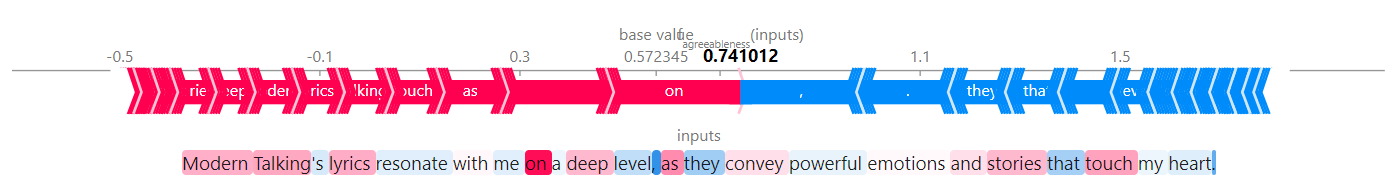}
    \caption{SHAP visualisation for Agreeableness (2/5)}
    \label{fig:agree_shap2}
\end{figure*}

\begin{figure*}[htbp]
    \centering
    \includegraphics[width=\textwidth]{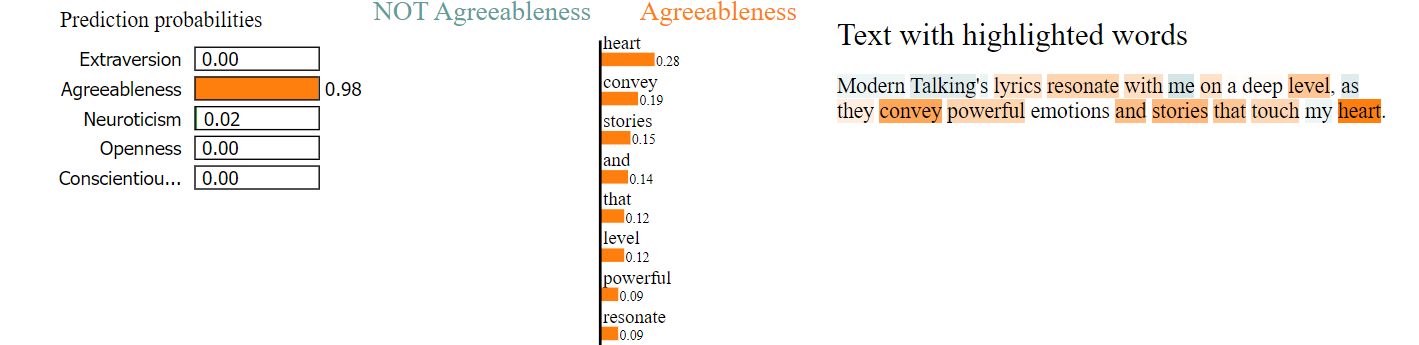}
    \caption{LIME visualisation for Agreeableness (2/5)}
    \label{fig:agree_lime2}
\end{figure*}

\begin{figure*}[htbp]
    \centering
    \includegraphics[width=\textwidth]{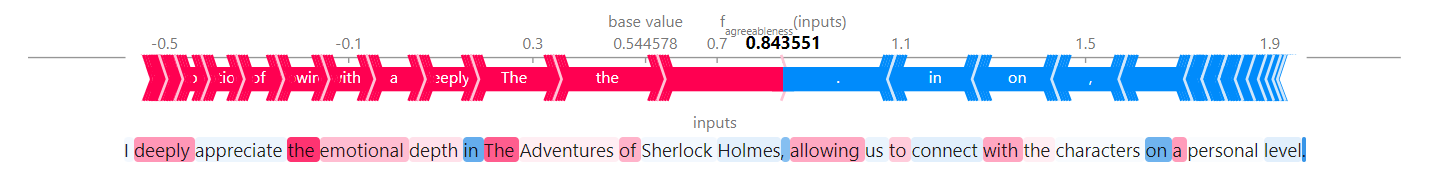}
    \caption{SHAP visualisation for Agreeableness (3/5)}
    \label{fig:agree_shap3}
\end{figure*}

\begin{figure*}[htbp]
    \centering
    \includegraphics[width=\textwidth]{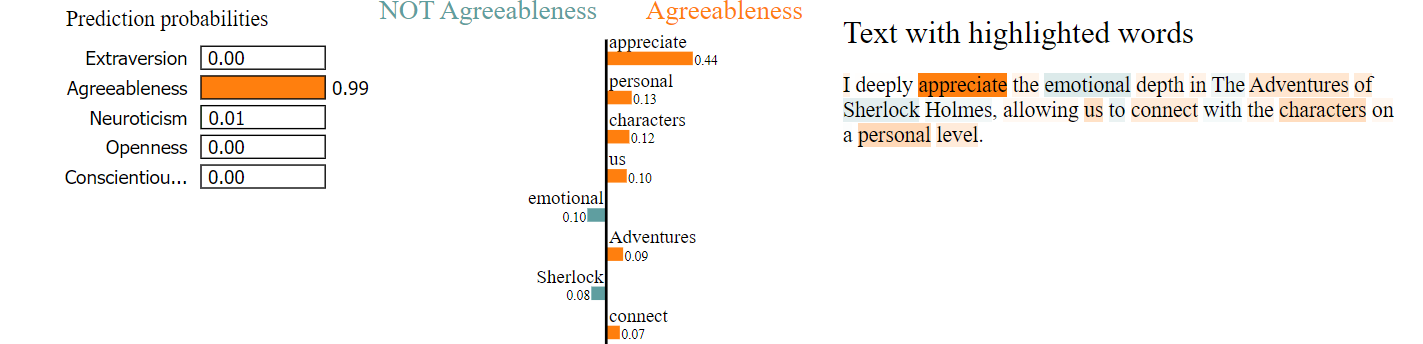}
    \caption{LIME visualisation for Agreeableness (3/5)}
    \label{fig:agree_lime3}
\end{figure*}

\begin{figure*}[htbp]
    \centering
    \includegraphics[width=\textwidth]{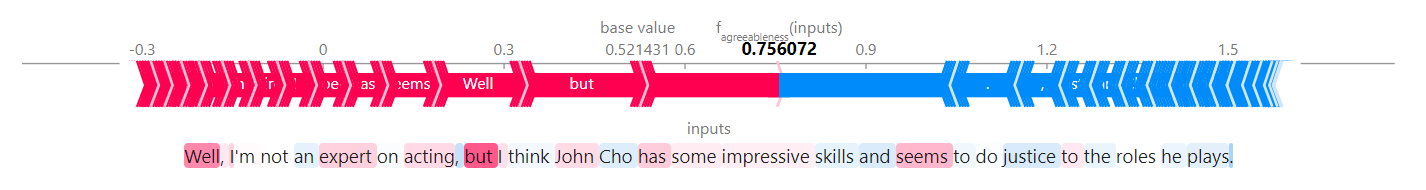}
    \caption{SHAP visualisation for Agreeableness (4/5)}
    \label{fig:agree_shap4}
\end{figure*}

\begin{figure*}[htbp]
    \centering
    \includegraphics[width=\textwidth]{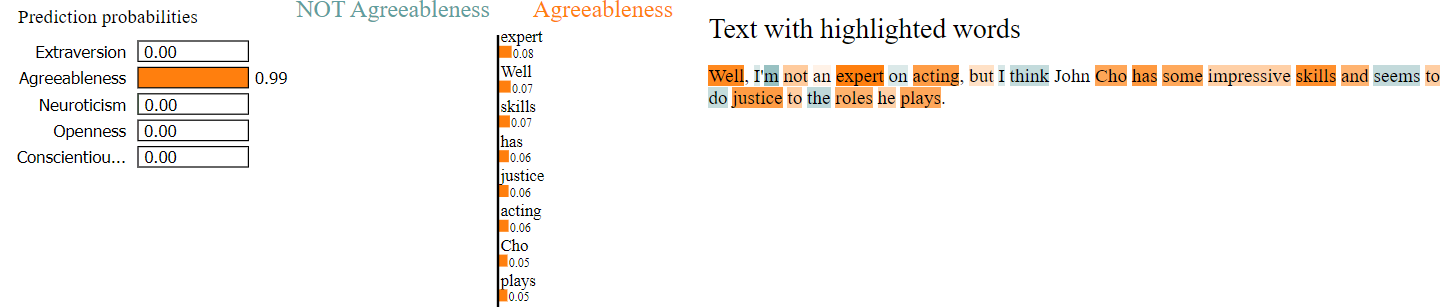}
    \caption{LIME visualisation for Agreeableness (4/5)}
    \label{fig:agree_lime4}
\end{figure*}

\begin{figure*}[htbp]
    \centering
    \includegraphics[width=\textwidth]{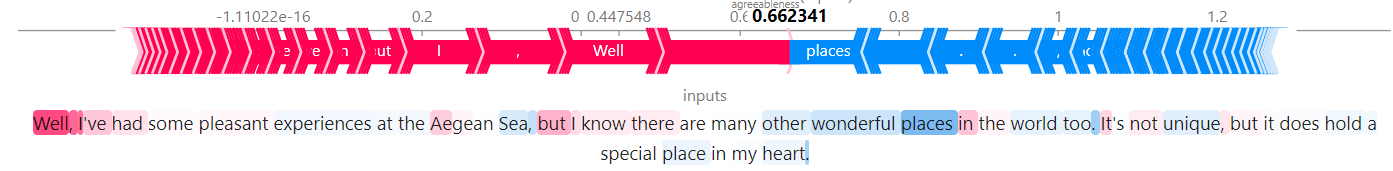}
    \caption{SHAP visualisation for Agreeableness (5/5)}
    \label{fig:agree_shap5}
\end{figure*}

\begin{figure*}[htbp]
    \centering
    \includegraphics[width=\textwidth]{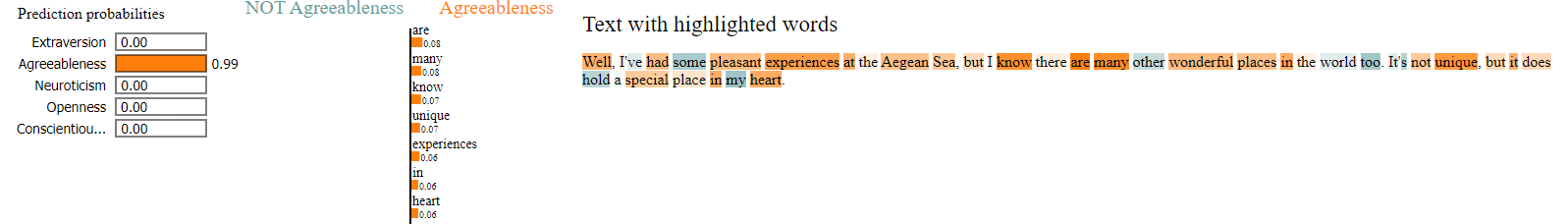}
    \caption{LIME visualisation for Agreeableness (5/5)}
    \label{fig:agree_lime5}
\end{figure*}

\begin{figure*}[htbp]
    \centering
    \includegraphics[width=\textwidth]{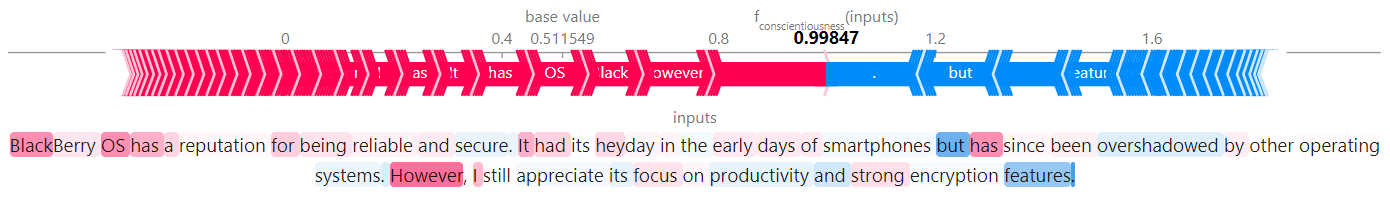}
    \caption{SHAP visualisation for Conscientiousness (1/5)}
    \label{fig:con_shap1}
\end{figure*}

\begin{figure*}[htbp]
    \centering
    \includegraphics[width=\textwidth]{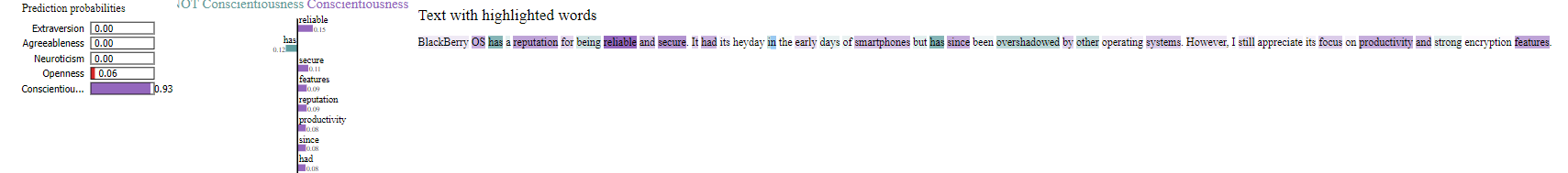}
    \caption{LIME visualisation for Conscientiousness (1/5)}
    \label{fig:con_lime1}
\end{figure*}

\begin{figure*}[htbp]
    \centering
    \includegraphics[width=\textwidth]{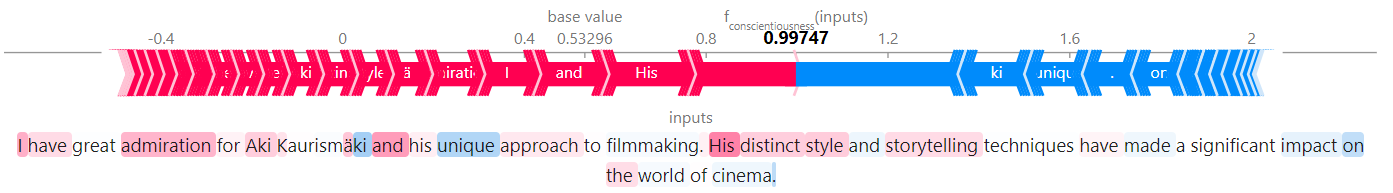}
    \caption{SHAP visualisation for Conscientiousness (2/5)}
    \label{fig:con_shap2}
\end{figure*}

\begin{figure*}[htbp]
    \centering
    \includegraphics[width=\textwidth]{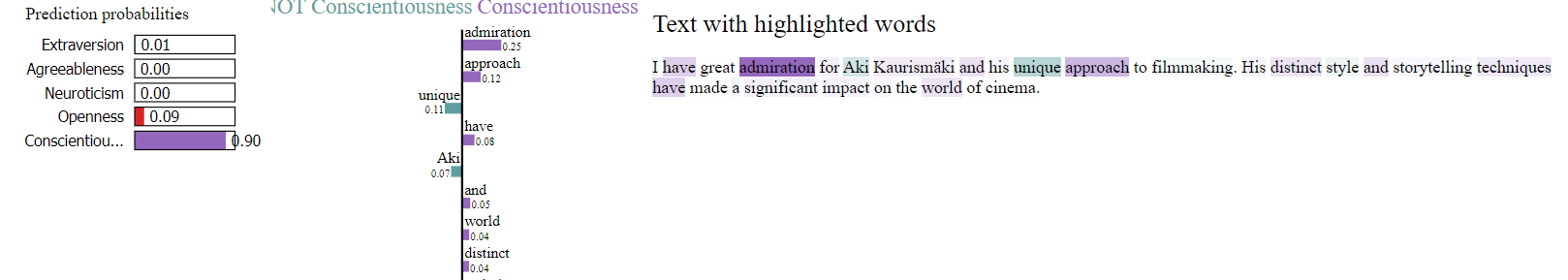}
    \caption{LIME visualisation for Conscientiousness (2/5)}
    \label{fig:con_lime2}
\end{figure*}

\begin{figure*}[htbp]
    \centering
    \includegraphics[width=\textwidth]{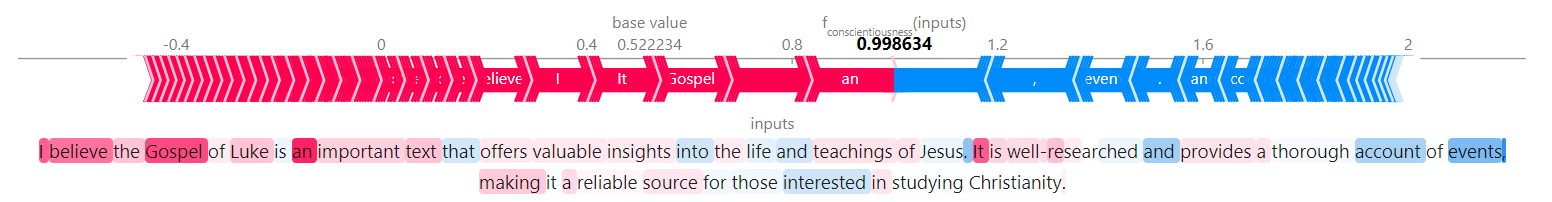}
    \caption{SHAP visualisation for Conscientiousness (3/5)}
    \label{fig:con_shap3}
\end{figure*}

\begin{figure*}[htbp]
    \centering
    \includegraphics[width=\textwidth]{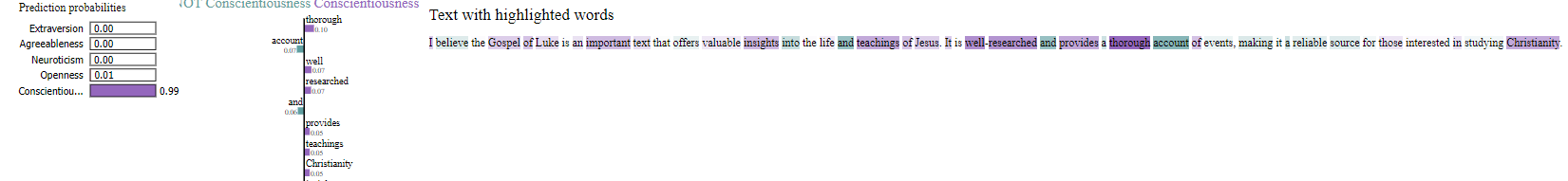}
    \caption{LIME visualisation for Conscientiousness (3/5)}
    \label{fig:con_lime3}
\end{figure*}

\begin{figure*}[htbp]
    \centering
    \includegraphics[width=\textwidth]{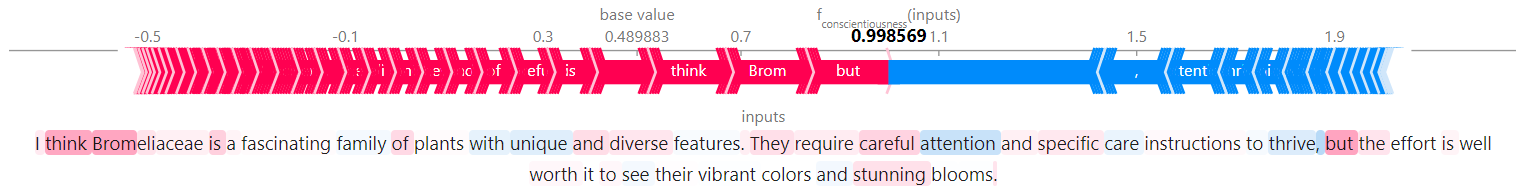}
    \caption{SHAP visualisation for Conscientiousness (4/5)}
    \label{fig:con_shap4}
\end{figure*}

\begin{figure*}[htbp]
    \centering
    \includegraphics[width=\textwidth]{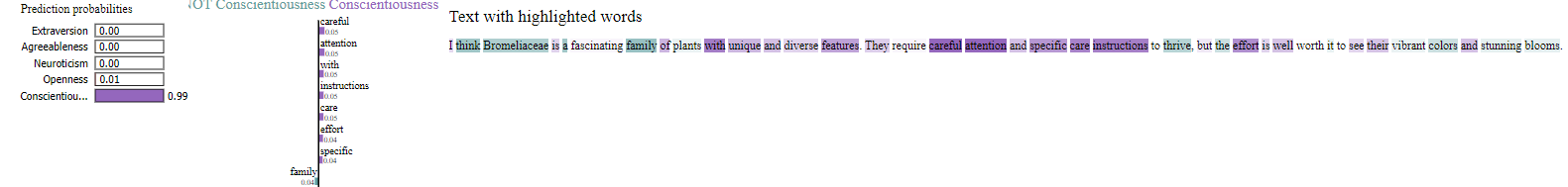}
    \caption{LIME visualisation for Conscientiousness (4/5)}
    \label{fig:con_lime4}
\end{figure*}

\begin{figure*}[htbp]
    \centering
    \includegraphics[width=\textwidth]{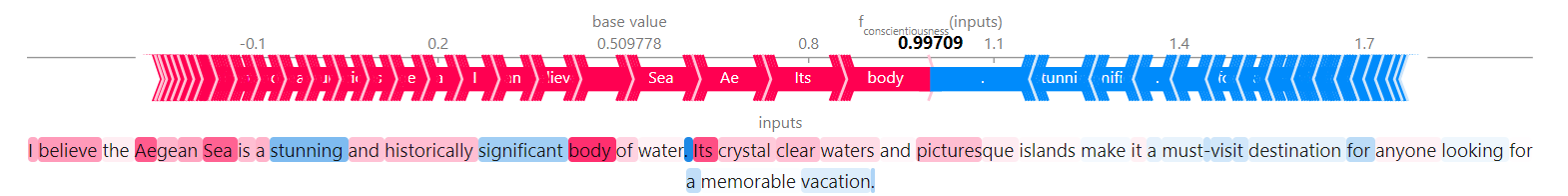}
    \caption{SHAP visualisation for Conscientiousness (5/5)}
    \label{fig:con_shap5}
\end{figure*}

\begin{figure*}[htbp]
    \centering
    \includegraphics[width=\textwidth]{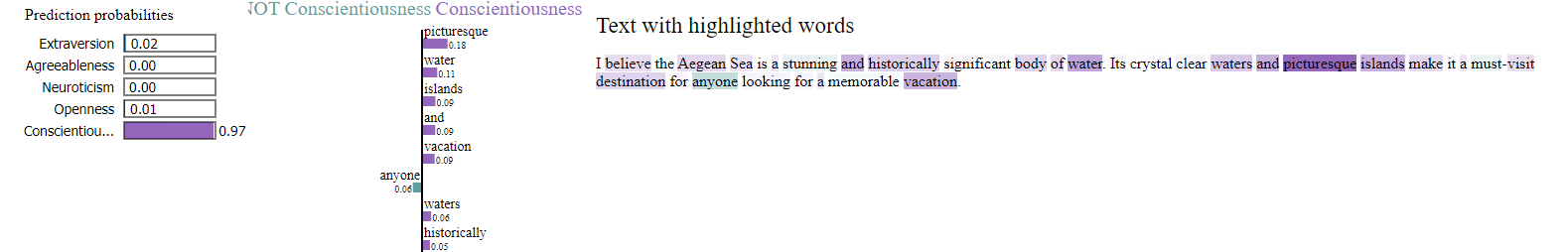}
    \caption{LIME visualisation for Conscientiousness (5/5)}
    \label{fig:con_lime5}
\end{figure*}

\begin{figure*}[htbp]
    \centering
    \includegraphics[width=\textwidth]{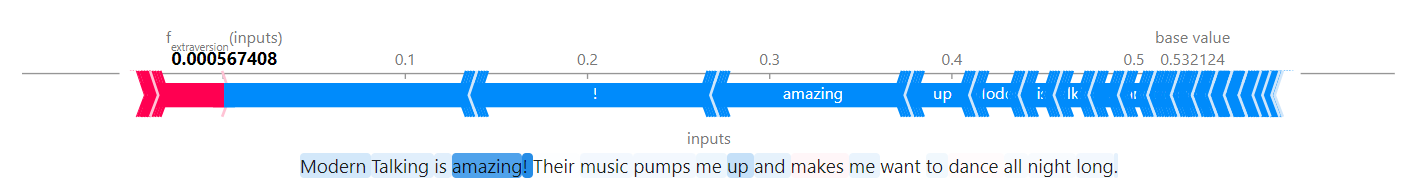}
    \caption{SHAP visualisation for Extraversion (1/5)}
    \label{fig:ext_shap1}
\end{figure*}

\begin{figure*}[htbp]
    \centering
    \includegraphics[width=\textwidth]{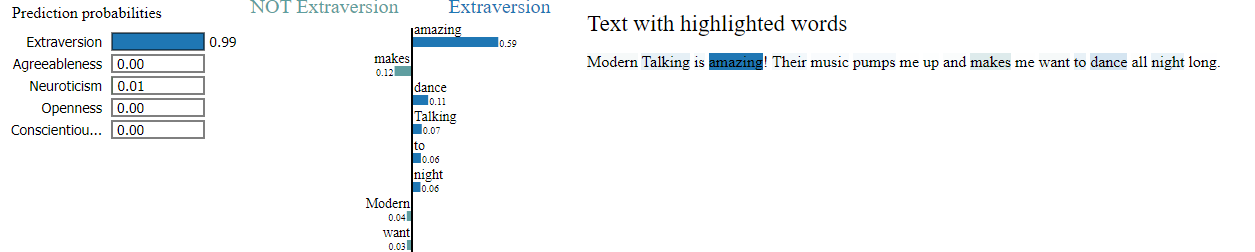}
    \caption{LIME visualisation for Extraversion (1/5)}
    \label{fig:ext_lime1}
\end{figure*}

\begin{figure*}[htbp]
    \centering
    \includegraphics[width=\textwidth]{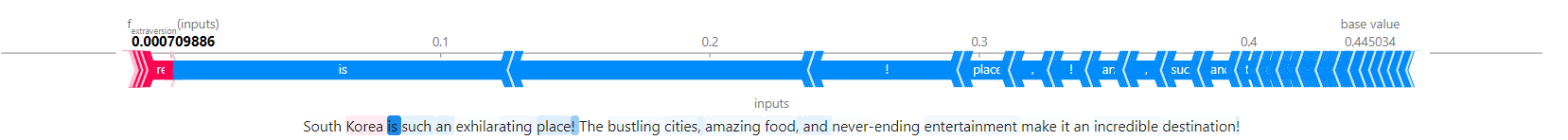}
    \caption{SHAP visualisation for Extraversion (2/5)}
    \label{fig:ext_shap2}
\end{figure*}

\begin{figure*}[htbp]
    \centering
    \includegraphics[width=\textwidth]{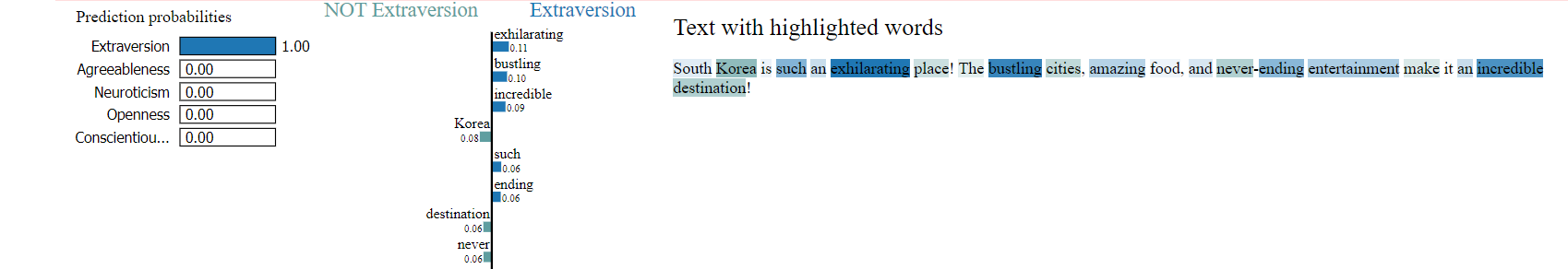}
    \caption{LIME visualisation for Extraversion (2/5)}
    \label{fig:ext_lime2}
\end{figure*}

\begin{figure*}[htbp]
    \centering
    \includegraphics[width=\textwidth]{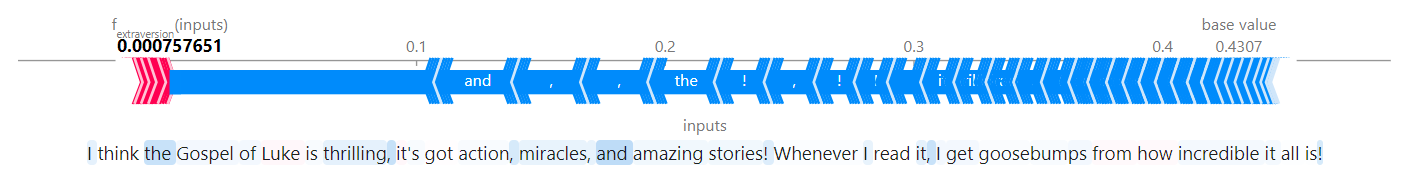}
    \caption{SHAP visualisation for Extraversion (3/5)}
    \label{fig:ext_shap3}
\end{figure*}

\begin{figure*}[htbp]
    \centering
    \includegraphics[width=\textwidth]{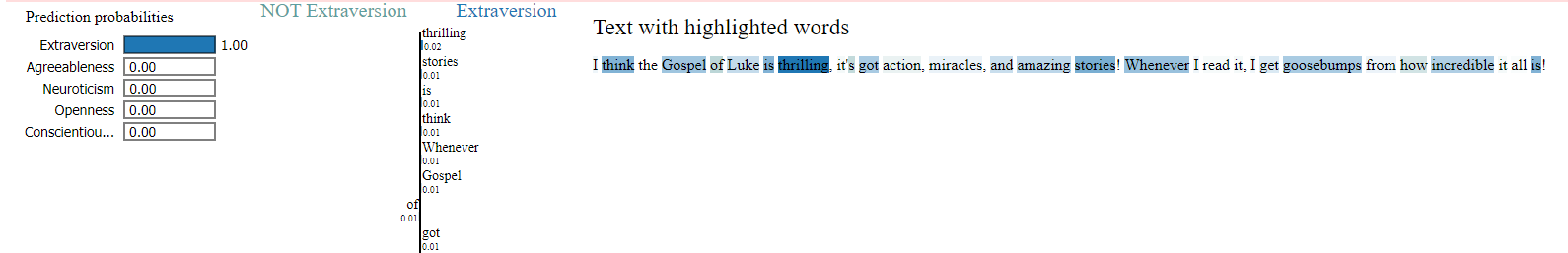}
    \caption{LIME visualisation for Extraversion (3/5)}
    \label{fig:ext_lime3}
\end{figure*}

\begin{figure*}[htbp]
    \centering
    \includegraphics[width=\textwidth]{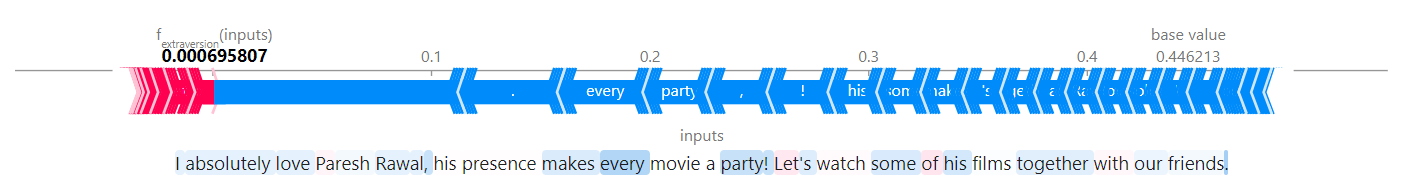}
    \caption{SHAP visualisation for Extraversion (4/5)}
    \label{fig:ext_shap4}
\end{figure*}

\begin{figure*}[htbp]
    \centering
    \includegraphics[width=\textwidth]{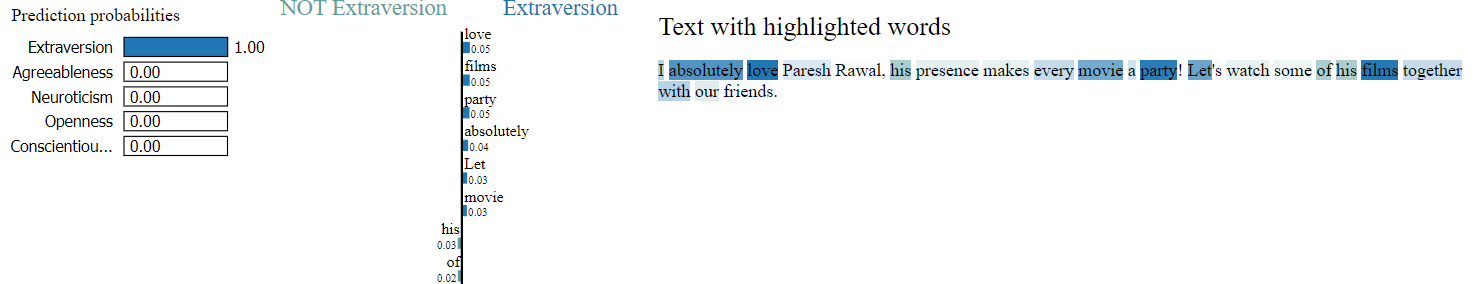}
    \caption{LIME visualisation for Extraversion (4/5)}
    \label{fig:ext_lime4}
\end{figure*}

\begin{figure*}[htbp]
    \centering
    \includegraphics[width=\textwidth]{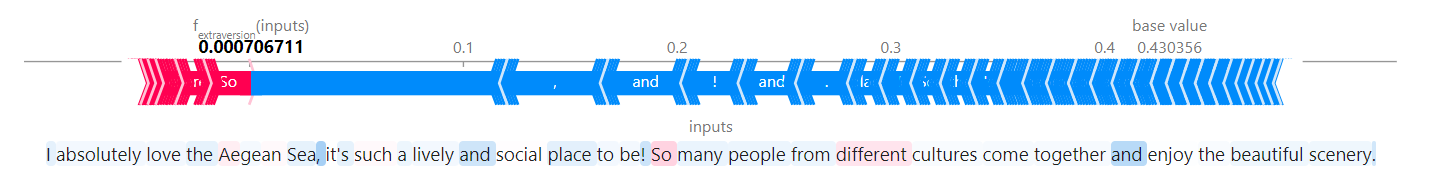}
    \caption{SHAP visualisation for Extraversion (5/5)}
    \label{fig:ext_shap5}
\end{figure*}

\begin{figure*}[htbp]
    \centering
    \includegraphics[width=\textwidth]{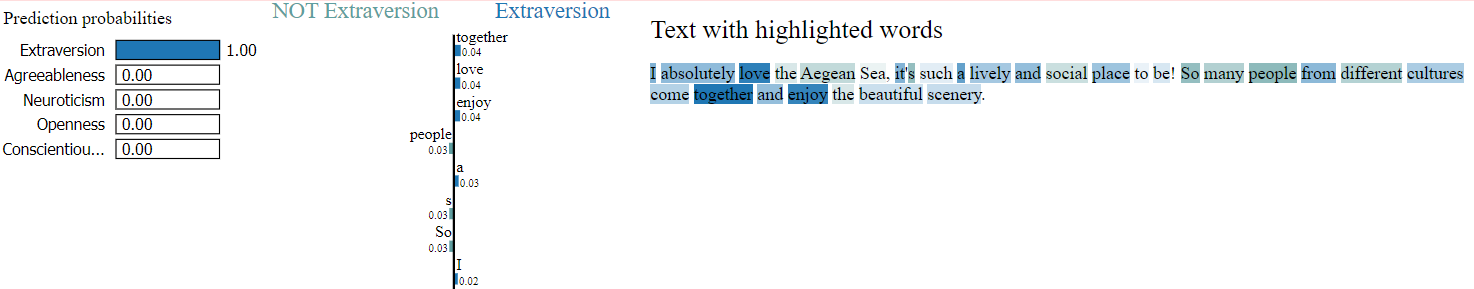}
    \caption{LIME visualisation for Extraversion (5/5)}
    \label{fig:ext_lime5}
\end{figure*}

\begin{figure*}[htbp]
    \centering
    \includegraphics[width=\textwidth]{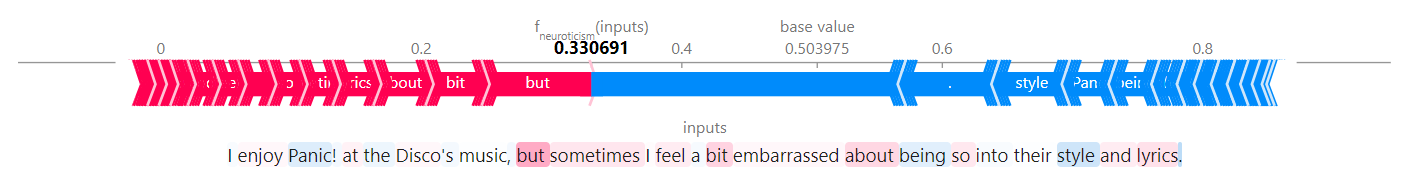}
    \caption{SHAP visualisation for Neuroticism (1/5)}
    \label{fig:neu_shap2}
\end{figure*}

\begin{figure*}[htbp]
    \centering
    \includegraphics[width=\textwidth]{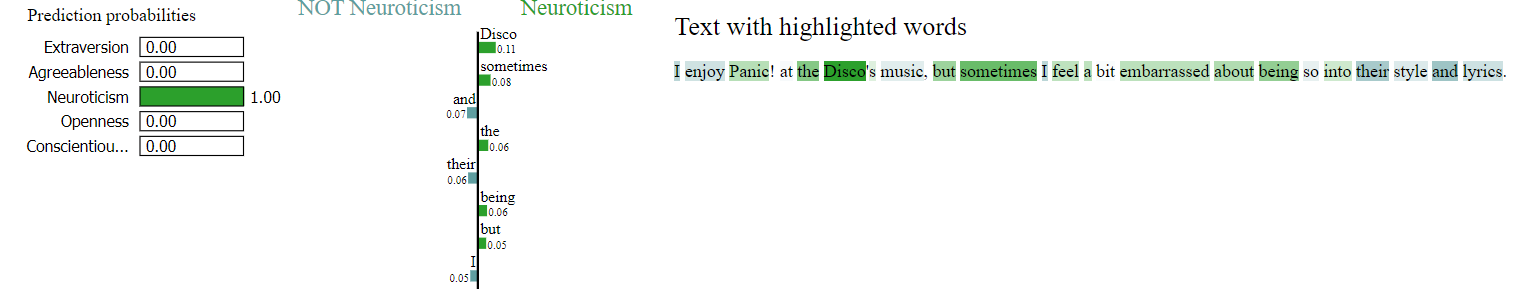}
    \caption{LIME visualisation for Neuroticism (1/5)}
    \label{fig:neu_lime2}
\end{figure*}

\begin{figure*}[htbp]
    \centering
    \includegraphics[width=\textwidth]{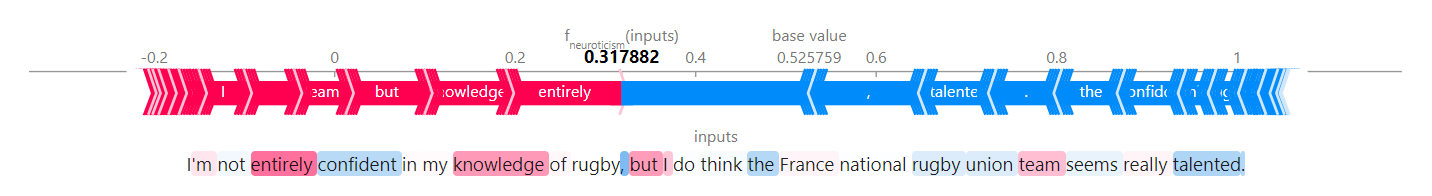}
    \caption{SHAP visualisation for Neuroticism (2/5)}
    \label{fig:neu_shap3}
\end{figure*}

\begin{figure*}[htbp]
    \centering
    \includegraphics[width=\textwidth]{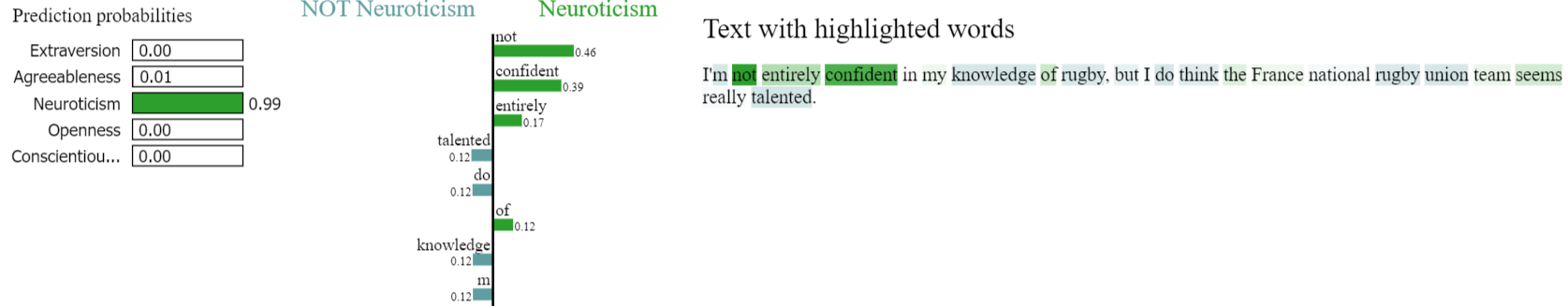}
    \caption{LIME visualisation for Neuroticism (2/5)}
    \label{fig:neu_lime3}
\end{figure*}

\begin{figure*}[htbp]
    \centering
    \includegraphics[width=\textwidth]{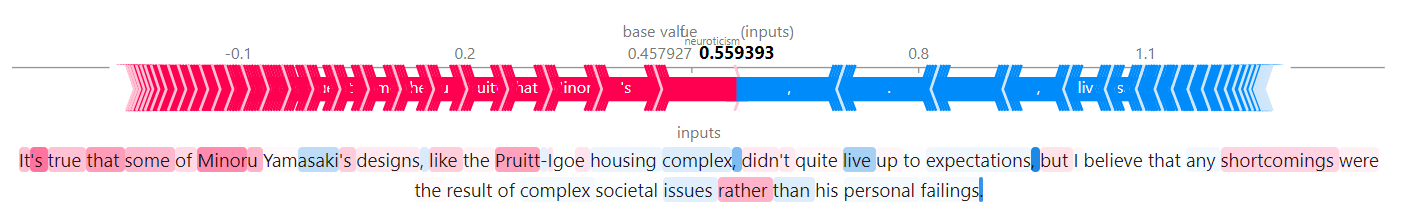}
    \caption{SHAP visualisation for Neuroticism (3/5)}
    \label{fig:neu_shap4}
\end{figure*}

\begin{figure*}[htbp]
    \centering
    \includegraphics[width=\textwidth]{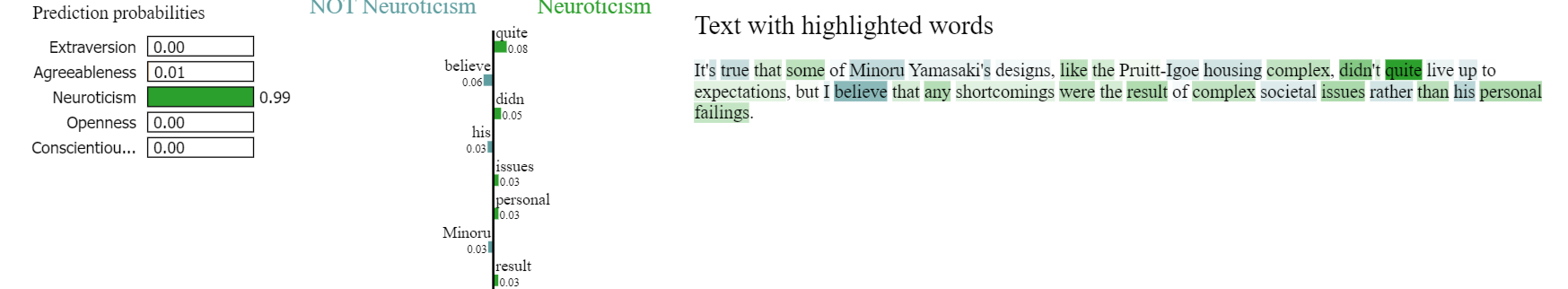}
    \caption{LIME visualisation for Neuroticism (3/5)}
    \label{fig:neu_lime4}
\end{figure*}

\begin{figure*}[htbp]
    \centering
    \includegraphics[width=\textwidth]{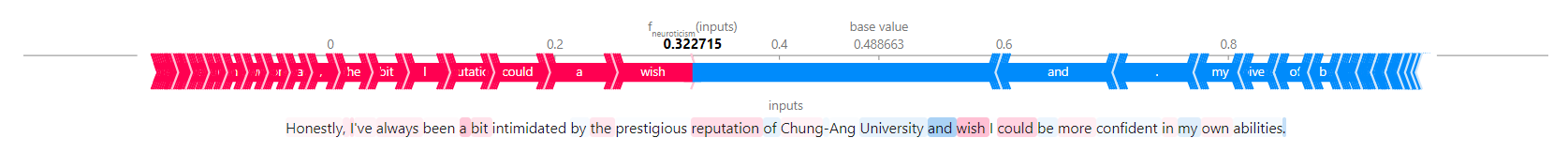}
    \caption{SHAP visualisation for Neuroticism (4/5)}
    \label{fig:neu_shap5}
\end{figure*}

\begin{figure*}[htbp]
    \centering
    \includegraphics[width=\textwidth]{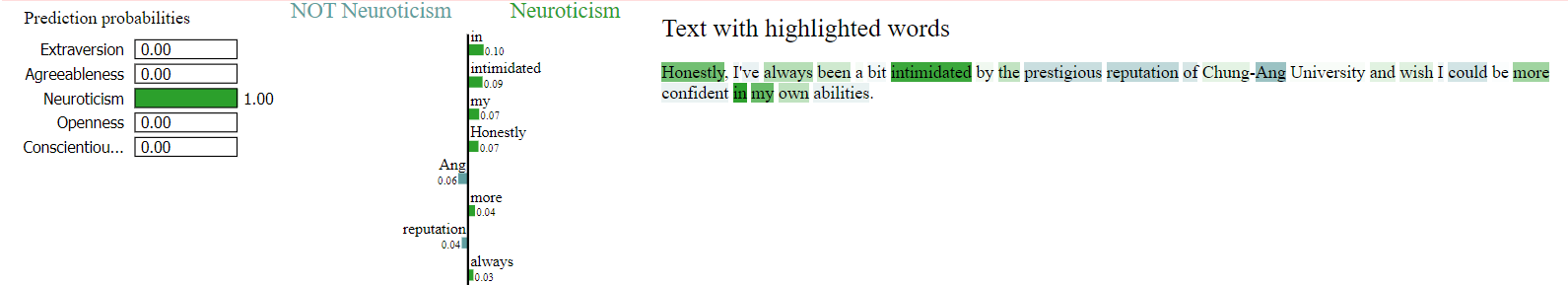}
    \caption{LIME visualisation for Neuroticism (4/5)}
    \label{fig:neu_lime5}
\end{figure*}

\begin{figure*}[htbp]
    \centering
    \includegraphics[width=\textwidth]{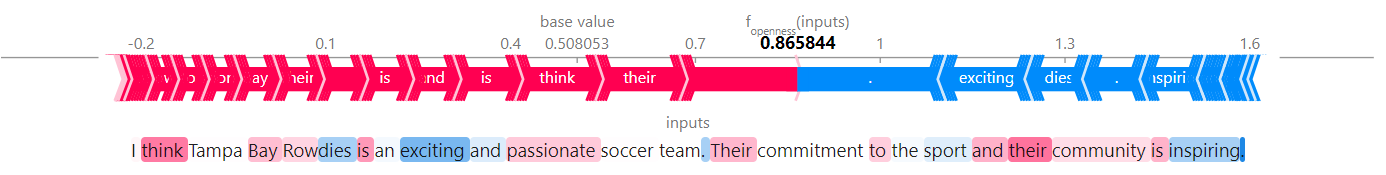}
    \caption{SHAP visualisation for Openness (1/5)}
    \label{fig:open_shap1}
\end{figure*}

\begin{figure*}[htbp]
    \centering
    \includegraphics[width=\textwidth]{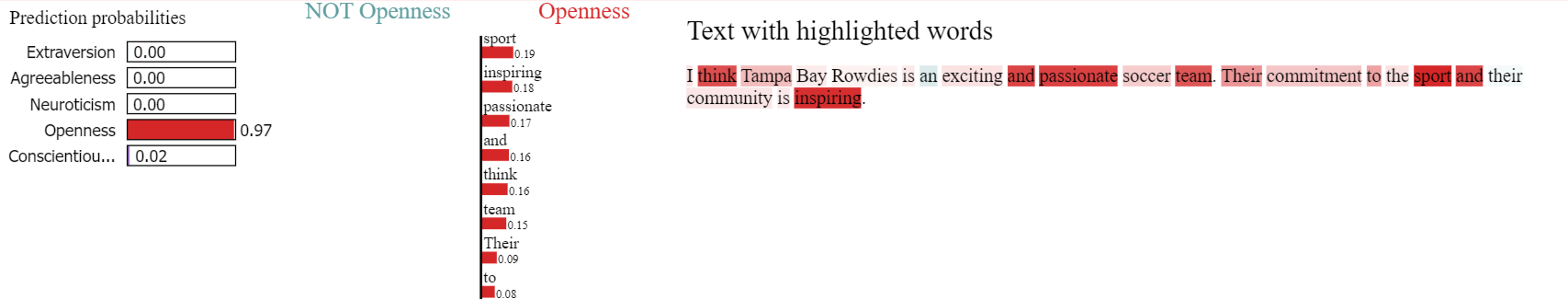}
    \caption{LIME visualisation for Openness (1/5)}
    \label{fig:open_lime1}
\end{figure*}

\begin{figure*}[htbp]
    \centering
    \includegraphics[width=\textwidth]{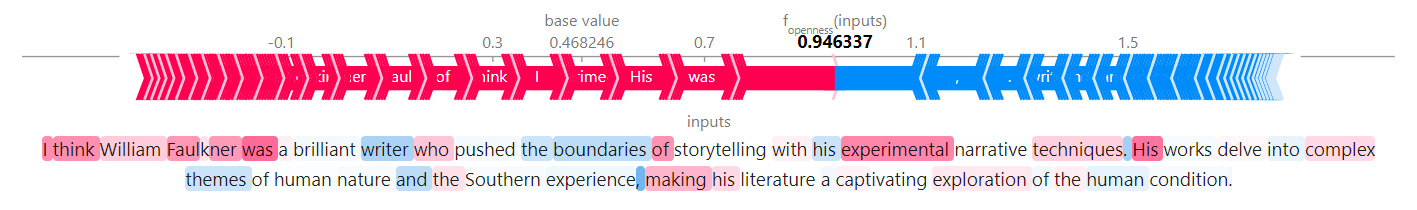}
    \caption{SHAP visualisation for Openness (2/5)}
    \label{fig:open_shap2}
\end{figure*}

\begin{figure*}[htbp]
    \centering
    \includegraphics[width=\textwidth]{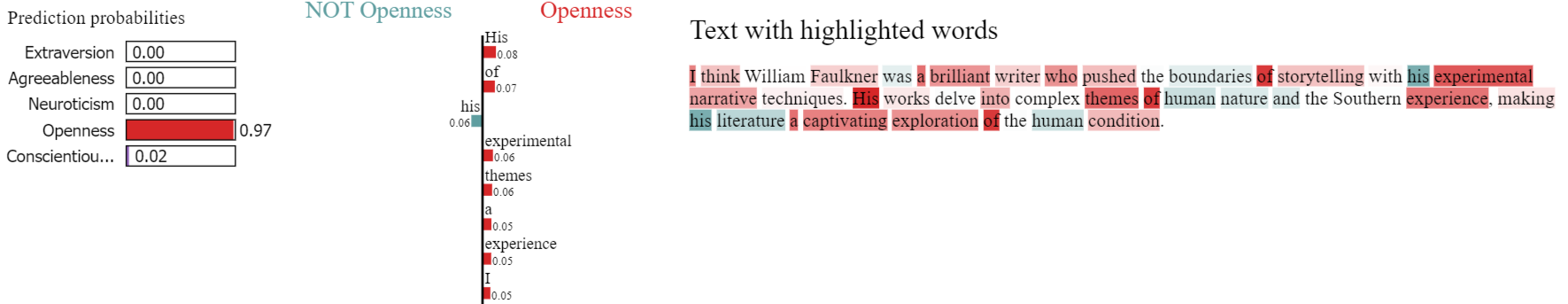}
    \caption{LIME visualisation for Openness (2/5)}
    \label{fig:open_lime2}
\end{figure*}

\begin{figure*}[htbp]
    \centering
    \includegraphics[width=\textwidth]{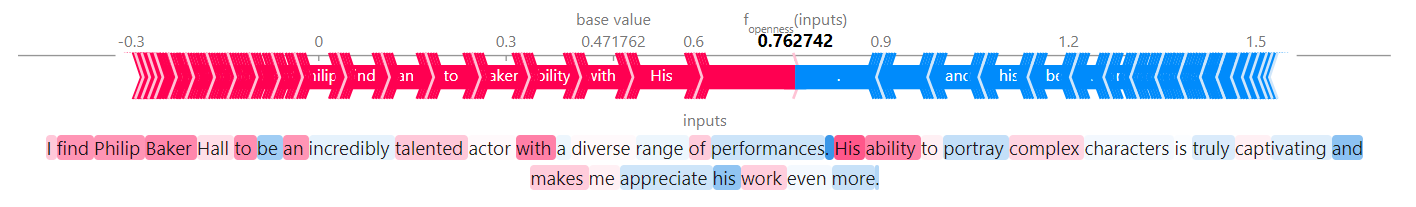}
    \caption{SHAP visualisation for Openness (3/5)}
    \label{fig:open_shap3}
\end{figure*}

\begin{figure*}[htbp]
    \centering
    \includegraphics[width=\textwidth]{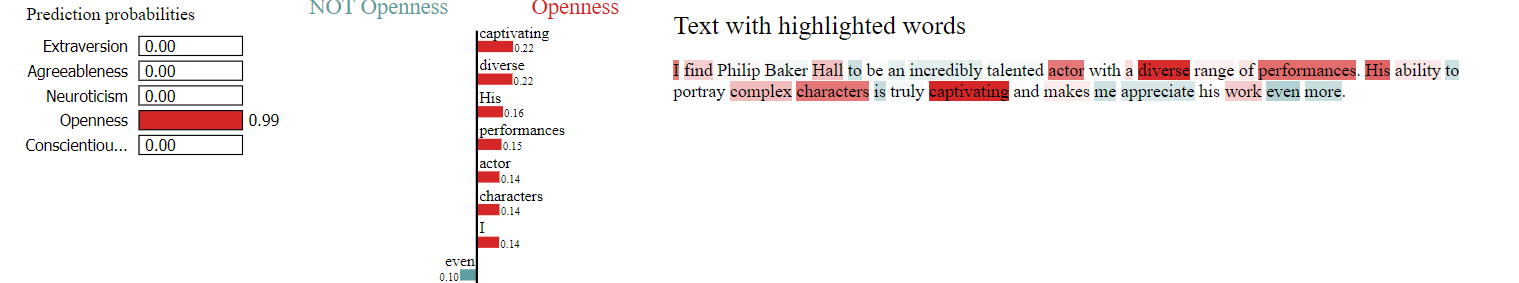}
    \caption{LIME visualisation for Openness (3/5)}
    \label{fig:open_lime3}
\end{figure*}

\begin{figure*}[htbp]
    \centering
    \includegraphics[width=\textwidth]{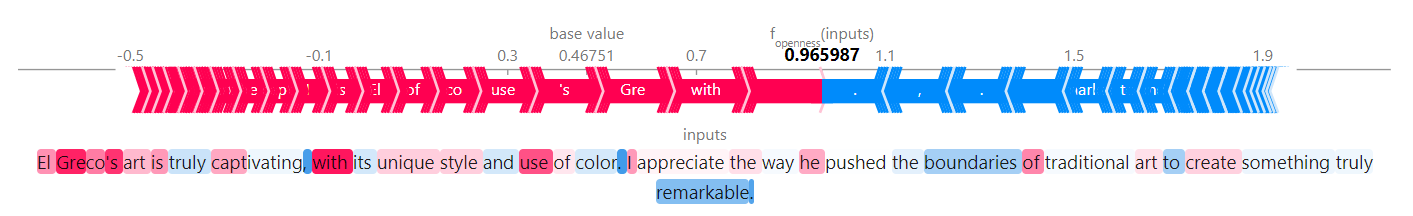}
    \caption{SHAP visualisation for Openness (4/5)}
    \label{fig:open_shap4}
\end{figure*}

\begin{figure*}[htbp]
    \centering
    \includegraphics[width=\textwidth]{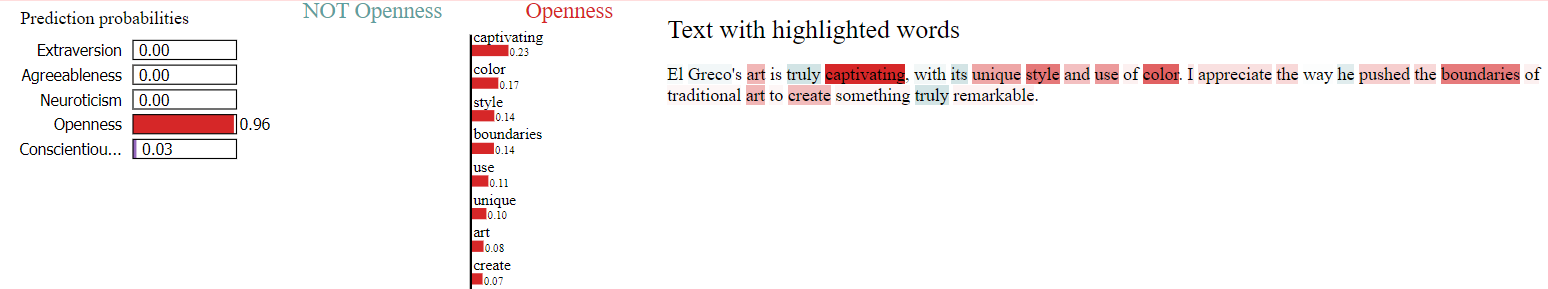}
    \caption{LIME visualisation for Openness (4/5)}
    \label{fig:open_lime4}
\end{figure*}

\clearpage
\begin{figure*}[htbp]
    \centering
    \includegraphics[width=\textwidth]{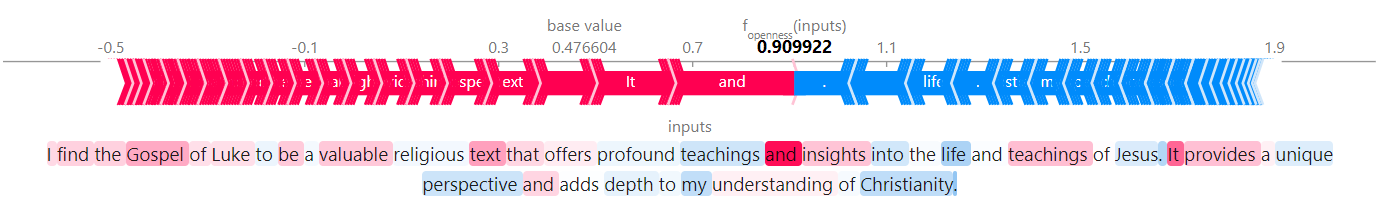}
    \caption{SHAP visualisation for Openness (5/5)}
    \label{fig:open_shap5}
\end{figure*}

\begin{figure*}[htbp]
    \centering
    \includegraphics[width=\textwidth]{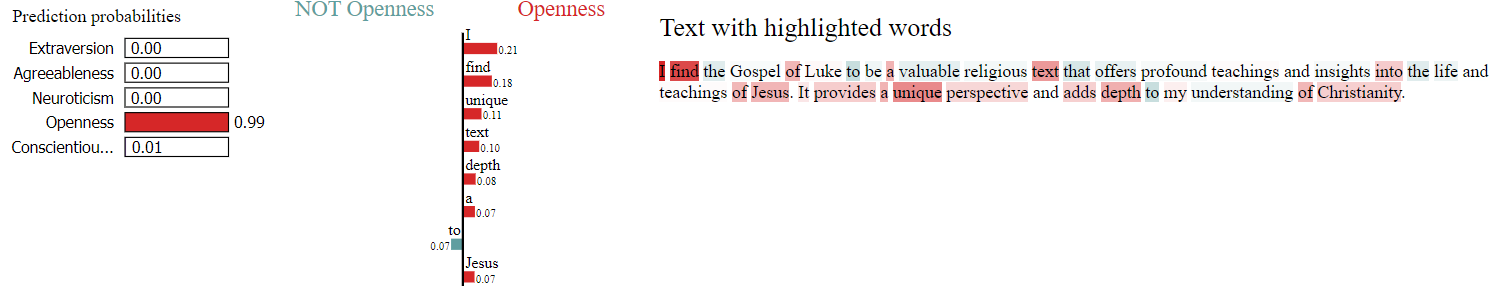}
    \caption{LIME visualisation for Openness (5/5)}
    \label{fig:open_lime5}
\end{figure*}

\end{document}